\documentclass[10pt,twocolumn,letterpaper]{article}

\usepackage{cvpr}              

\usepackage{graphicx}
\usepackage{amsmath}
\usepackage{amssymb}
\usepackage{booktabs}

\usepackage{multirow}
\usepackage{subcaption}
\usepackage[font={small}]{caption}
\usepackage{booktabs, dcolumn}
\usepackage{pifont} 
\usepackage[pagebackref,breaklinks,colorlinks]{hyperref}

\usepackage{color}
\usepackage{colortbl}
\definecolor{myRed}{rgb}{1.0, .0, .0}

\usepackage[capitalize]{cleveref}
\crefname{section}{Sec.}{Secs.}
\Crefname{section}{Section}{Sections}
\Crefname{table}{Table}{Tables}
\crefname{table}{Tab.}{Tabs.}

\def\cvprPaperID{5766} 
\def\confName{CVPR}
\def\confYear{2022}

\begin{document}

\title{NeW CRFs: Neural Window Fully-connected CRFs for Monocular Depth Estimation}

\author{
Weihao Yuan\hspace{0.5cm}
Xiaodong Gu\hspace{0.5cm}
Zuozhuo Dai\hspace{0.5cm}
Siyu Zhu\hspace{0.5cm}
Ping Tan\\
Alibaba Group
\vspace{-20mm}
}

\maketitle


\begin{abstract}

Estimating the accurate depth from a single image is challenging since it is inherently ambiguous and ill-posed. While recent works design increasingly complicated and powerful networks to directly regress the depth map, we take the path of CRFs optimization. 
Due to the expensive computation, CRFs are usually performed between neighborhoods rather than the whole graph.
To leverage the potential of fully-connected CRFs, we split the input into windows and perform the FC-CRFs optimization within each window, which reduces the computation complexity and makes FC-CRFs feasible. 
To better capture the relationships between nodes in the graph, we exploit the multi-head attention mechanism to compute a multi-head potential function, which is fed to the networks to output an optimized depth map.
Then we build a bottom-up-top-down structure, where this neural window FC-CRFs module serves as the decoder, and a vision transformer serves as the encoder.
The experiments demonstrate that our method significantly improves the performance across all metrics on both the KITTI and NYUv2 datasets, compared to previous methods.
Furthermore, the proposed method can be directly applied to panorama images and outperforms all previous panorama methods on the MatterPort3D dataset. \footnote{Project page: \url{https://weihaosky.github.io/newcrfs}}

\end{abstract}



\section{Introduction}
\label{sec:intro}

Depth prediction is a classical task in computer vision and is essential for numerous applications such as 3D reconstruction, autonomous driving, and robotics~\cite{izadi2011kinectfusion, geiger2012we, yuan2019reinforcement, yuan2019end}.
Such a vision task aims to estimate the depth map from a single color image, which is an ill-posed and inherently ambiguous problem since infinitely many 3D scenes can be projected to the same 2D scene.
Therefore, this task is challenging for traditional methods~\cite{michels2005high, nagai2002hmm, saxena2005learning}, which are usually limited to low-dimension and sparse distances~\cite{michels2005high}, or known and fixed objects~\cite{nagai2002hmm}.

Recently, many works have employed the deep networks to directly regress the depth maps and achieved good performances~\cite{eigen2014depth, lee2019big, aich2020bidirectional, lee2021patch, fu2018deep, bhat2021adabins}. 
Nevertheless, since there are no geometric constraints of multi-view~\cite{gu2021dro,yuan2020mfusenet,yuan2021stereo} to exploit, the focus of most works is designing more powerful and more complicated networks.
This renders this task a difficult fitting problem without the help of other guidance.



In traditional monocular depth estimation, some methods build the energy function from Markov Random Fields (MRFs) or Conditional Random Fields (CRFs)~\cite{saxena2005learning, saxena2008make3d, wang2015depth}. 
They exploit the observation cues, such as the texture and position information, along with the last prediction to build the energy function, and then optimize this energy to obtain a depth prediction.
This approach is demonstrated to be effective in guiding the estimation of the depth, and is also introduced in some deep methods~\cite{liu2015deep, hua2016depth, ricci2018monocular, xu2018structured}.
However, they are all limited in neighbor CRFs rather than fully-connected CRFs (FC-CRFs) due to the expensive computation, while the fully-connected CRFs capture the relationship between any node in a graph and are much stronger.

\begin{figure}[t]
\centering
  \includegraphics[width=0.9\columnwidth, trim={0cm 0cm 0cm 0cm}, clip]{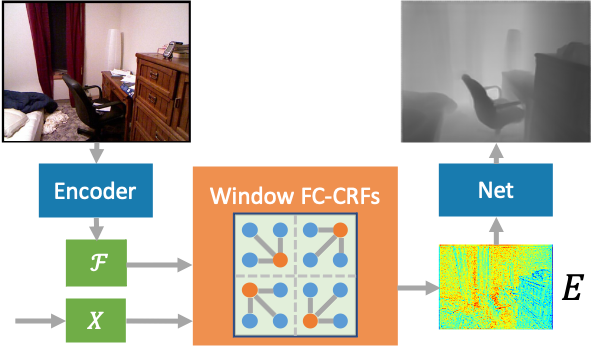}
\caption{The neural window fully-connected CRFs take image feature $\mathcal{F}$ and upper-level prediction $X$ as input, and compute the fully-connected energy $E$ in each window, which is then fed to the networks to output an optimized depth map.
}
\label{fig:intro}
\end{figure}

To address the above challenge, in this work we segment the input to multiple windows, and build the fully-connected CRFs energy within each window, in which way the computation complexity is reduced considerably and the fully-connected CRFs becomes feasible.
To capture more relationships between the nodes in the graph, we exploit the multi-head attention mechanism~\cite{vaswani2017attention} to compute the pairwise potential of the CRFs, and build a new neural CRFs module, as is shown in Figure~\ref{fig:intro}.
By employing this neural window FC-CRFs module as decoder, and a vision transformer as encoder, we build a straightforward bottom-up-top-down network to estimate the depth.
To make up for the isolation of each window, a window shift action~\cite{liu2021swin} is performed,
and the lack of global information in these window FC-CRFs is addressed by aggregating the global features from global average pooling layers~\cite{zhao2017pyramid}.

In the experiments, our method is demonstrated to outperform previous methods by a {significant margin} on both the outdoor dataset, KITTI~\cite{geiger2012we}, and the indoor dataset, NYUv2~\cite{silberman2012indoor}.
Although the state-of-the-art performance on KITTI and NYUv2 has been saturated for a while, our method further reduces the errors considerably on both datasets.
Specifically, the Abs-Rel error and the RMS error of KITTI are decreased by $10.3\%$ and $9.8\%$, and that of NYUv2 are decreased by $7.8\%$ and $8.2\%$.
Our method now {ranks first} among all submissions on the KITTI online benchmark.
In addition, we evaluate our method on the panorama images.
As is well-known, the networks designed for perspective images usually perform poorly on the panorama dataset~\cite{tateno2018distortion,wang2020bifuse,jiang2021unifuse,sun2021hohonet}.
Remarkably, our method also sets a new state-of-the-art performance on the panorama dataset, MatterPort3D~\cite{chang2017matterport3d}. 
This demonstrates that our method can handle the common scenarios in the monocular depth prediction task.


The main contributions of this work are then summarized as follows:

$\bullet$ We split the input image into sub-windows and perform fully-connected CRFs optimization within each window, which reduces the high computation complexity and makes the FC-CRFs feasible.

$\bullet$ We employ the multi-head attention to capture the pairwise relationships in the window FC-CRFs, and embed this neural CRFs module in a network to serve as the decoder.

$\bullet$ We build a new bottom-up-top-down network for monocular depth estimation and show a significant improvement of the monocular depth across all metrics on KITTI, NYUv2, and MatterPort3D datasets.

\section{Related Work}
\label{sec:related}


\subsection{Traditional Monocular Depth Estimation}

Prior to the emergence of deep learning, monocular depth estimation is a challenging task.
Many published works limit themselves in either estimating 1-D distances of obstacles~\cite{michels2005high} or limited in several known and fixed objects~\cite{nagai2002hmm}.
Then Saxena et al.~\cite{saxena2005learning} claim that local features alone are insufficient to predict the depth of a pixel, and the global context of the whole image needs to be considered to infer the depth. Therefore, they use a discriminatively-trained Markov Random Field (MRF) to incorporate multiscale local and global image features, and model both depths at individual pixels as well as the relation between depths at different pixels.
In this way, they infer good depth maps from the monocular cues like colors, pixel positions, occlusion, known object sizes, haze, defocus, etc.
Since then, MRFs~\cite{saxena2008make3d} and CRFs~\cite{wang2015depth} have been well used in monocular depth estimation in traditional methods.
However, the traditional approaches still suffer from estimating accurate high-resolution dense depth maps.

\subsection{Neural Networks Based Monocular Depth}

In monocular depth estimation, neural network based methods have dominated most benchmarks.
There are mainly two kinds of approaches for learning the mapping from images to depth maps.
The first approach directly regresses the continuous depth map from the aggregation of the information in an image~\cite{eigen2014depth, qi2018geonet, lee2019big, yin2019enforcing, aich2020bidirectional, huynh2020guiding, lee2021patch, ranftl2021vision}. 
In this approach, coarse and fine networks are first introduced in \cite{eigen2014depth} and then improved by multi-stage local planar guidance layers in \cite{lee2019big}. 
A bidirectional attention module is proposed in \cite{aich2020bidirectional} to utilize the feed-forward feature maps and incorporate the global context to filter out ambiguity.
Recently, more methods have begun to employ vision transformers to aggregate the information of images~\cite{ranftl2021vision}. 
The second approach tries to discretize the depth space and convert the depth prediction to a classification or ordinal regression problem~\cite{fu2018deep, bhat2021adabins}. A spacing-increasing quantization strategy is used in \cite{fu2018deep} to discretize the depth space more reasonably. Then, an adaptive bins division is computed by the neural networks for better depth quantization. 
Also, other approaches bring in auxiliary information to help the training of the depth network, such as sparse depth~\cite{guizilini2021sparse} or segmentation information~\cite{zhang2019pattern, ochs2019sdnet, klingner2020self, qiao2021vip}.
All these approaches try to directly regress the depth map from the image feature, which falls into a difficult fitting problem. The structures of their networks become increasingly complex.
In contrast to these works, we build an energy with the fully-connected CRFs, and then optimize this energy to obtain a high-quality depth map.

\subsection{Neural CRFs for Monocular Depth}

Since the graph models, like MRFs and CRFs, are effective in traditional depth estimation, some methods try to embed them into neural networks~\cite{li2015depth, liu2015deep, hua2016depth, ricci2018monocular, xu2018structured}. 
These methods regard the patches of pixels as nodes and perform the graph optimization. 
One such approach first uses networks to regress a coarse depth map and then utilizes CRFs to refine it~\cite{li2015depth}, where the post-processing function of CRFs is proven to be effective. 
However, the CRFs are separated from neural networks.
To better combine CRFs and networks, other methods integrate CRFs into the layers of the neural networks and train the whole framework end-to-end~\cite{liu2015deep, hua2016depth, ricci2018monocular, xu2018structured}.
But they are all limited to CRFs rather than fully-connected CRFs due to the high computation complexity. 

In this work, different from previous methods, we split the whole graph into multiple sub-windows, such that the fully-connected CRFs become feasible.
Also, inspired by recent works in vision transformer\cite{vaswani2017attention, dosovitskiy2020image, liu2021swin},
we use the multi-head attention mechanism to capture the pairwise relationship in FC-CRFs and propose a neural window fully-connected CRFs module.
This module is embedded into the network to play the role of the decoder, such that the whole framework can be trained end-to-end.

\section{Neural Window Fully-connected CRFs}
\label{sec:method}

\begin{figure}[t]
\centering
  \includegraphics[width=0.8\columnwidth, trim={0cm 0cm 0cm 0cm}, clip]{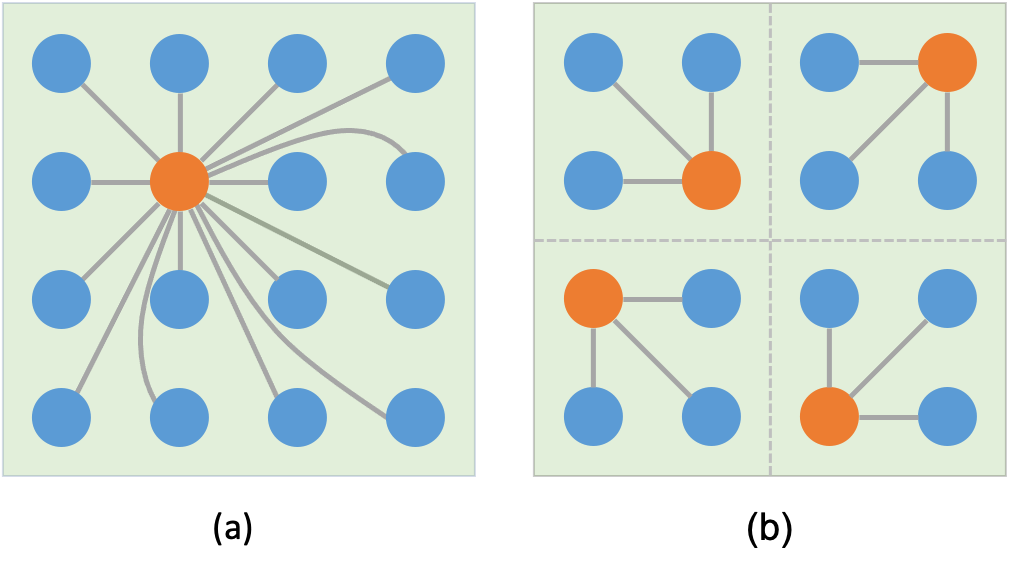}
\caption{Graph model of fully-connected CRFs and window fully-connected CRFs. In a fully-connected CRFs graph (a), taking the orange node as an example, it is connected to all other nodes in the graph. In a window fully-connected CRFs, however, the orange node is only connected to all other nodes within one window.}
\label{fig:graph}
\vspace{-3mm}
\end{figure}

\begin{figure*}[t]
\centering
  \includegraphics[width=1.8\columnwidth, trim={0cm 0cm 0cm 0cm}, clip]{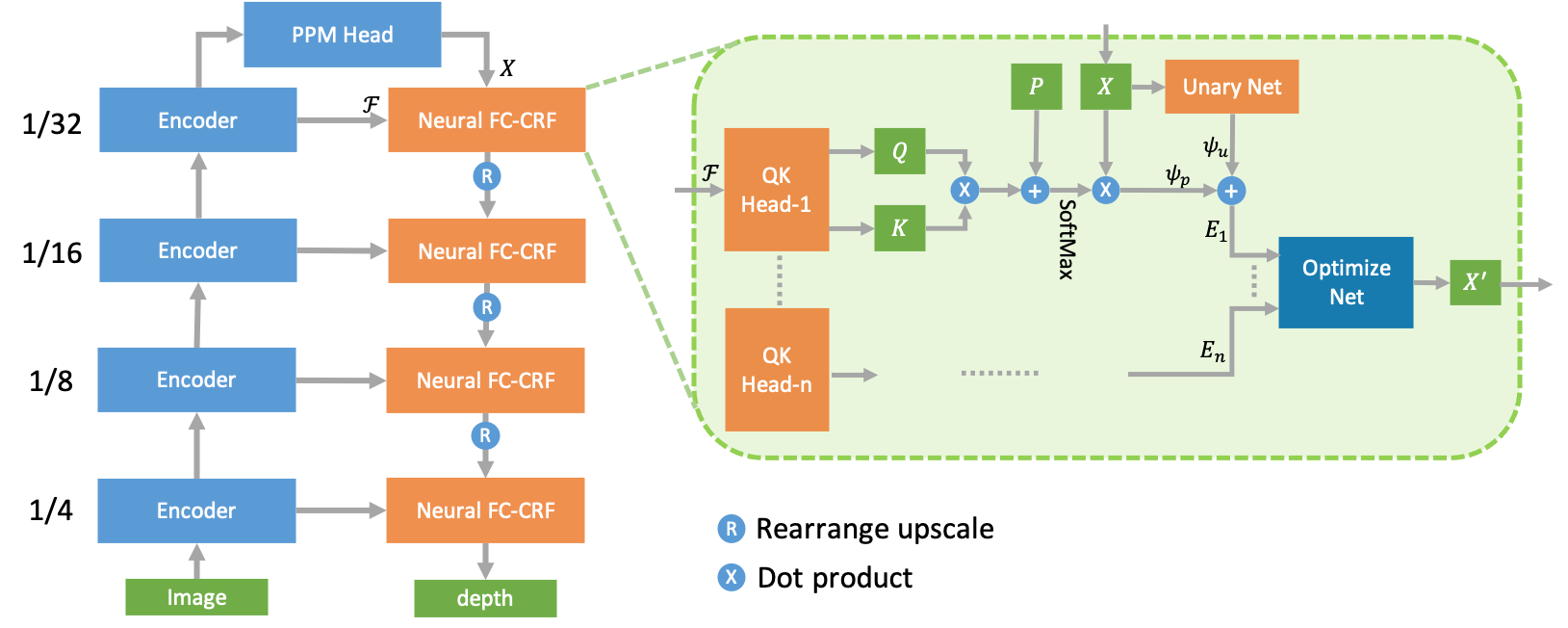}
\caption{Network structure of the proposed framework. The encoder first extracts the features in four levels.
A PPM head aggregates the global and local information and makes the initial prediction $X$ from the top image feature $\mathcal{F}$. Then in each level, the neural window fully-connected CRFs module builds multi-head energy from $X$ and $\mathcal{F}$, and optimizes it to a better prediction $X'$. Between each level a rearrange upscale is performed considering the sharpness and network weight.}
\label{fig:net}
\vspace{-2mm}
\end{figure*}

This section first introduces the window fully-connected CRFs, followed by its integration with neural networks. 
Afterward, the network structure is displayed, where the neural window FC-CRFs module is embedded into a top-down-bottom-up network to serve as the decoder.

\subsection{Fully-connected Conditional Random Fields}

In traditional methods, Markov random fields (MRFs) or conditional random fields (CRFs) are leveraged to handle dense prediction tasks such as monocular depth estimation \cite{saxena2005learning} and semantic segmentation \cite{chen2017deeplab}. They are shown to be effective in correcting the erroneous predictions based on the information of the current and adjacent nodes. Specifically, in a graph model, these methods favor similar label assignments to nodes that are proximal in space and color.

Thus, in this work we employ CRFs to help the depth prediction. Since the depth prediction of the current pixel is determined by long-range pixels in one image, to increase the receptive field, we use fully-connected CRFs \cite{krahenbuhl2011efficient} to build the energy. In a graph model, the energy function of the fully-connected CRFs is usually defined as 
\begin{equation}
    E(\mathbf{x}) = \sum_i \psi_u(x_i) + \sum_{ij} \psi_{p}(x_i, x_j),
\end{equation}
where $x_i$ is the predicted value of node $i$, and $j$ denotes all other nodes in the graph. The unary potential function $\psi_u$ is computed for each node by the predictor according to the image features.

The pairwise potential function $\psi_{p}$ connects pairs of nodes as
\begin{equation}
    \psi_{p} = \mu(x_i, x_j)f(x_i, x_j)g(I_i, I_j)h(p_i, p_j),
\end{equation}
where $\mu(x_i, x_j)=1$ if $i\neq j$ and $\mu(x_i, x_j)=0$ otherwise, $I_i$ is the color of node $i$, $p_i$ is the position of node $i$. The pairwise potential usually considers the color and position information to enforce some heuristic punishments, which make the predicted values $x_i, x_j$ more reasonable and logical.

In regular CRFs, the pairwise potential only computes the edge connection between the current node and neighboring nodes. In fully-connected CRFs, however, the connections between the current node and any other nodes in a graph need to be computed, as shown in Figure~\ref{fig:graph}~(a).

\subsection{Window Fully-connected CRFs}

Although the fully-connected CRFs can bring global-range connections, its disadvantage is also obvious. 
On the one hand, the number of edges connecting all pixels in an image is large, which makes the computation of this kind of pairwise potential quite resource-consuming. 
On the other hand, the depth of a pixel is usually not determined by distant pixels. Only pixels within some distance need to be considered.

Therefore, in this work we propose the window-based fully-connected CRFs. We segment an image into multiple patch-based windows. Each window includes $N\times N$ image patches, of which each patch is composed of $n\times n$ pixels.
In our graph model, each patch rather than each pixel is regarded as one node.
All patches within one window are fully-connected with edges, while the patches of different windows are not connected, as displayed in Figure~\ref{fig:graph}~(b). In this case, the computation of pairwise potential only considers the patches within one window, so that the computation complexity is reduced significantly.

Taking an image with $h\times w$ patches as an example, the computation complexity of FC-CRFs and window FC-CRFs for one iteration are
\begin{equation}
\begin{aligned}
    \Omega(\text{FC-CRFs}) = & \ hw \times \Omega (\psi_u) + hw (hw-1) \times \Omega(\psi_p)\\
    \Omega(\text{Window FC}) = & \ hw \times \Omega (\psi_u) +  hw (N^2 -1) \times \Omega(\psi_p) ,
\end{aligned}
\end{equation}
where $N$ is the window size, $\Omega (\mu_u)$ and $\Omega(\mu_p)$ are the computation complexity of one unary potential and one pairwise potential, respectively.

In the window fully-connected CRFs, all windows are non-overlapped, which means there is no information connection between any windows. The adjacent windows, however, are physically connected. To resolve the isolation of windows, we shift the windows by $(\frac{N}{2}, \frac{N}{2})$ patches in the image and calculate the energy function of shifted windows after computing that of the original windows, similar to swin-transformer~\cite{liu2021swin}. 
In this way, the isolated neighboring pixels are connected in the shifted windows.
Hence, each time we calculate the energy function, we calculate two energy functions successively, one for the original windows and the other one for the shifted windows.

\begin{figure*}[t]
\centering
\begin{subfigure}{0.49\columnwidth}
  \centering
  \includegraphics[width=1\columnwidth, trim={0cm 0cm 0cm 0cm}, clip]{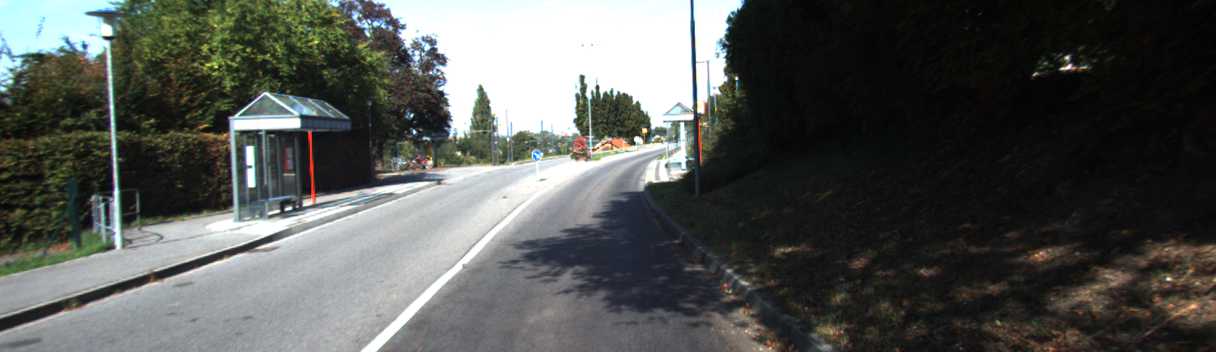}
\end{subfigure}
\begin{subfigure}{0.49\columnwidth}
  \centering
  \includegraphics[width=1\columnwidth, trim={0cm 0cm 0cm 0cm}, clip]{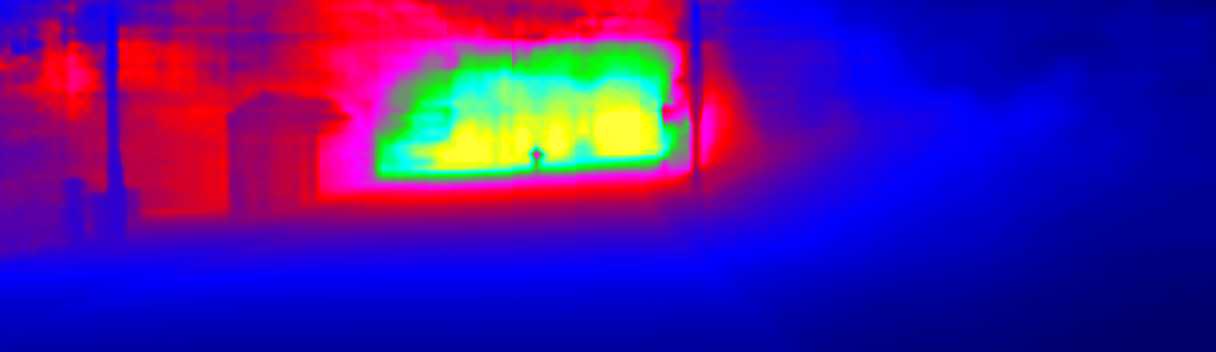}
\end{subfigure}
\begin{subfigure}{0.49\columnwidth}
  \centering
  \includegraphics[width=1\columnwidth, trim={0cm 0cm 0cm 0cm}, clip]{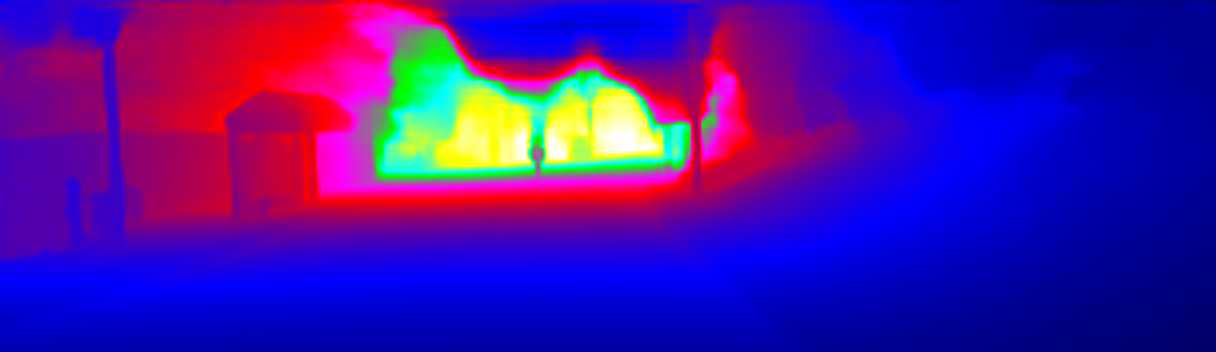}
\end{subfigure}
\begin{subfigure}{0.49\columnwidth}
  \centering
  \includegraphics[width=1\columnwidth, trim={0cm 0cm 0cm 0cm}, clip]{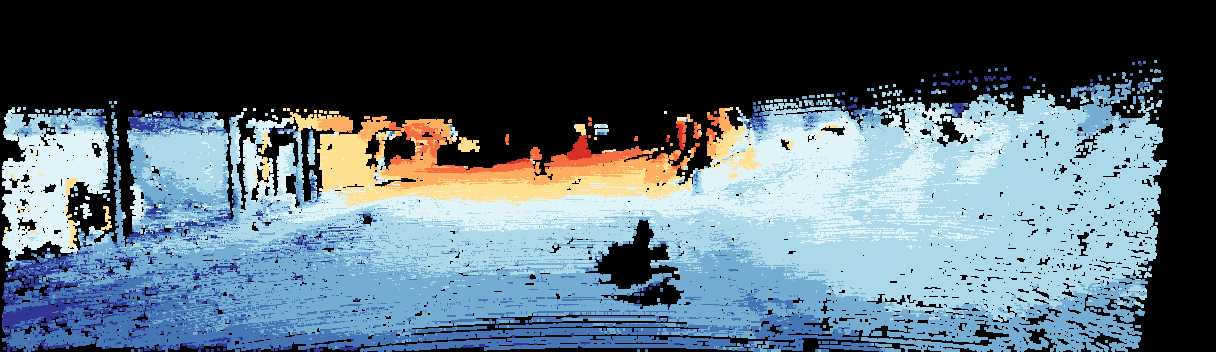}
\end{subfigure}

\begin{subfigure}{0.49\columnwidth}
  \centering
  \includegraphics[width=1\columnwidth, trim={0cm 0cm 0cm 0cm}, clip]{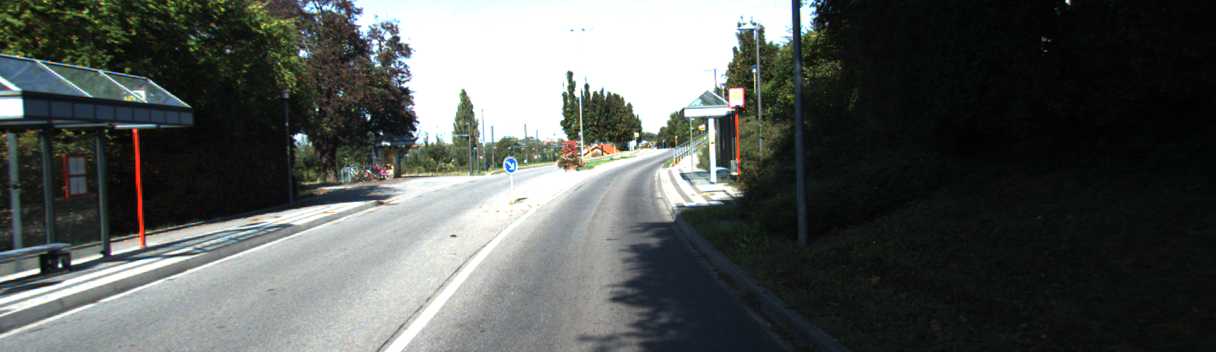}
\end{subfigure}
\begin{subfigure}{0.49\columnwidth}
  \centering
  \includegraphics[width=1\columnwidth, trim={0cm 0cm 0cm 0cm}, clip]{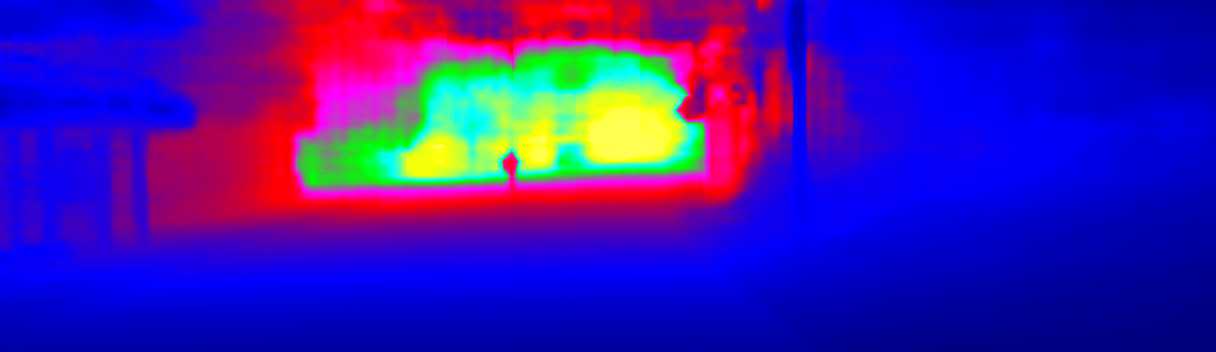}
\end{subfigure}
\begin{subfigure}{0.49\columnwidth}
  \centering
  \includegraphics[width=1\columnwidth, trim={0cm 0cm 0cm 0cm}, clip]{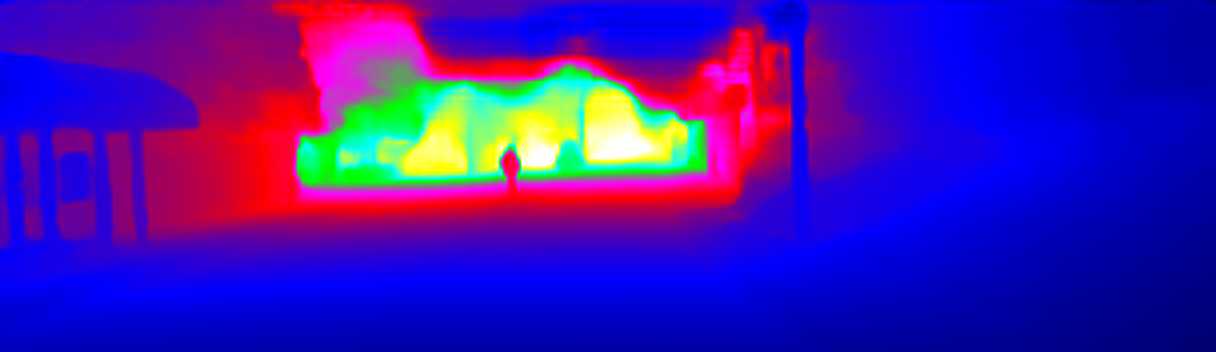}
\end{subfigure}
\begin{subfigure}{0.49\columnwidth}
  \centering
  \includegraphics[width=1\columnwidth, trim={0cm 0cm 0cm 0cm}, clip]{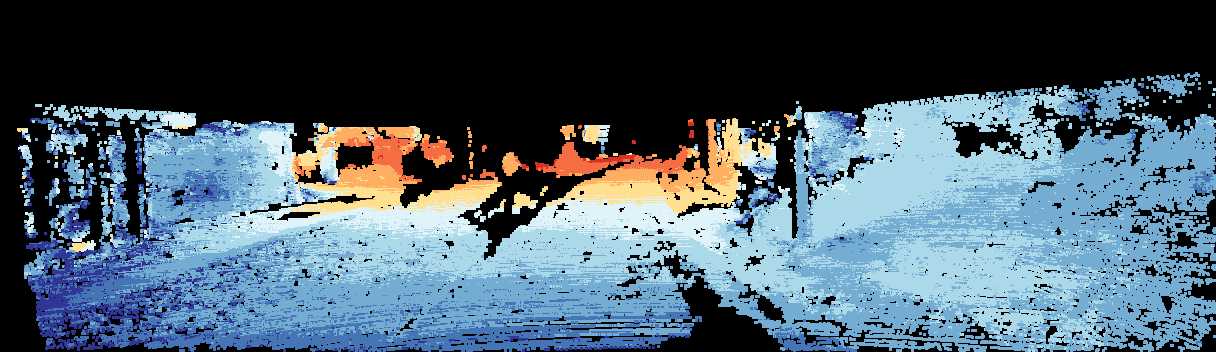}
\end{subfigure}

\begin{subfigure}{0.49\columnwidth}
  \centering
  \includegraphics[width=1\columnwidth, trim={0cm 0cm 0cm 0cm}, clip]{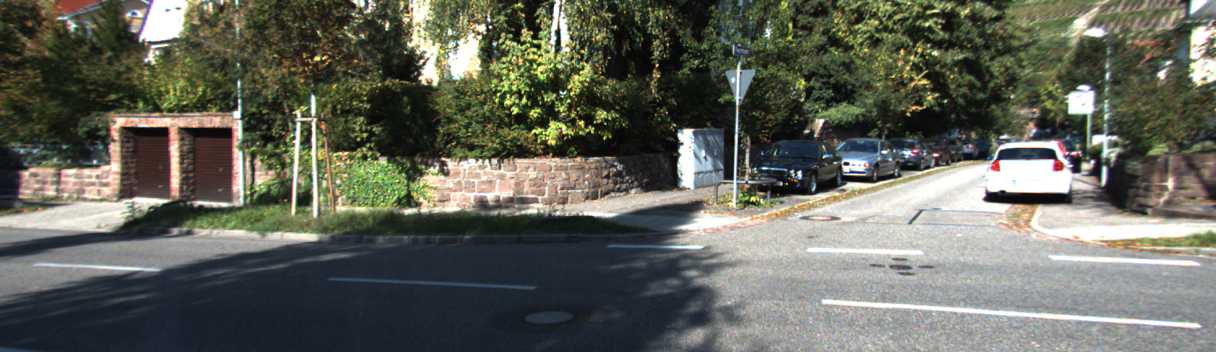}
\end{subfigure}
\begin{subfigure}{0.49\columnwidth}
  \centering
  \includegraphics[width=1\columnwidth, trim={0cm 0cm 0cm 0cm}, clip]{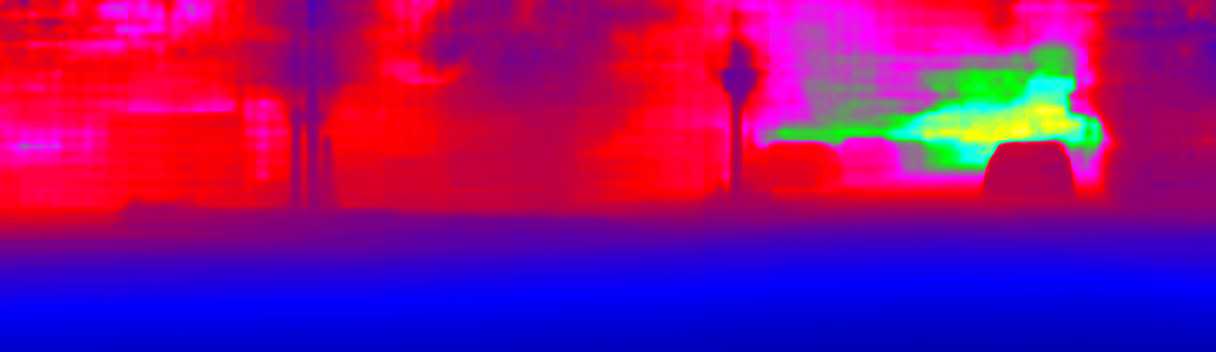}
\end{subfigure}
\begin{subfigure}{0.49\columnwidth}
  \centering
  \includegraphics[width=1\columnwidth, trim={0cm 0cm 0cm 0cm}, clip]{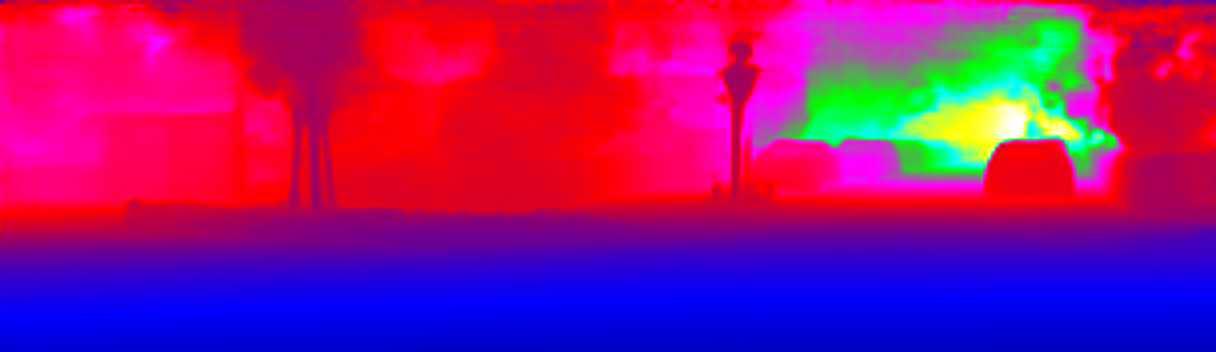}
\end{subfigure}
\begin{subfigure}{0.49\columnwidth}
  \centering
  \includegraphics[width=1\columnwidth, trim={0cm 0cm 0cm 0cm}, clip]{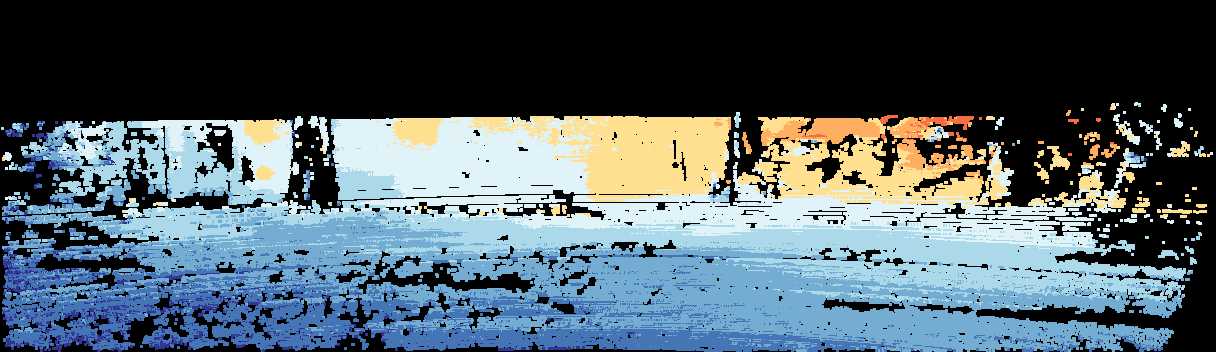}
\end{subfigure}

\begin{subfigure}{0.49\columnwidth}
  \centering
  \includegraphics[width=1\columnwidth, trim={0cm 0cm 0cm 0cm}, clip]{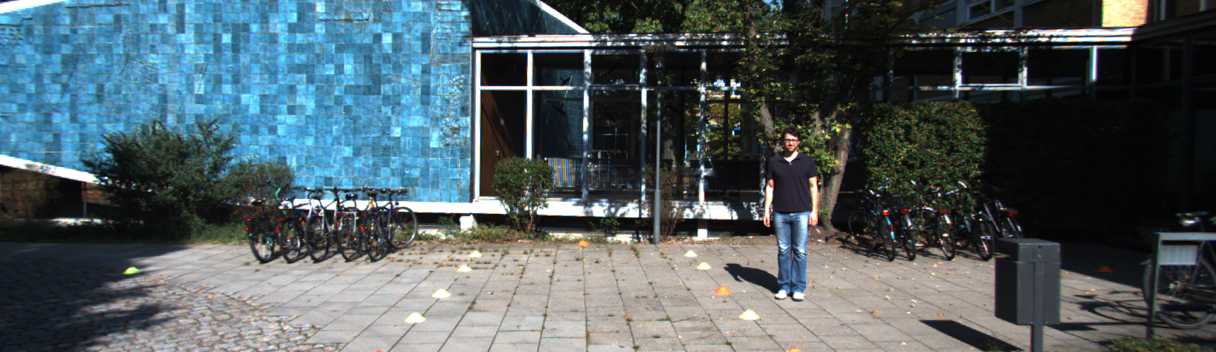}
\end{subfigure}
\begin{subfigure}{0.49\columnwidth}
  \centering
  \includegraphics[width=1\columnwidth, trim={0cm 0cm 0cm 0cm}, clip]{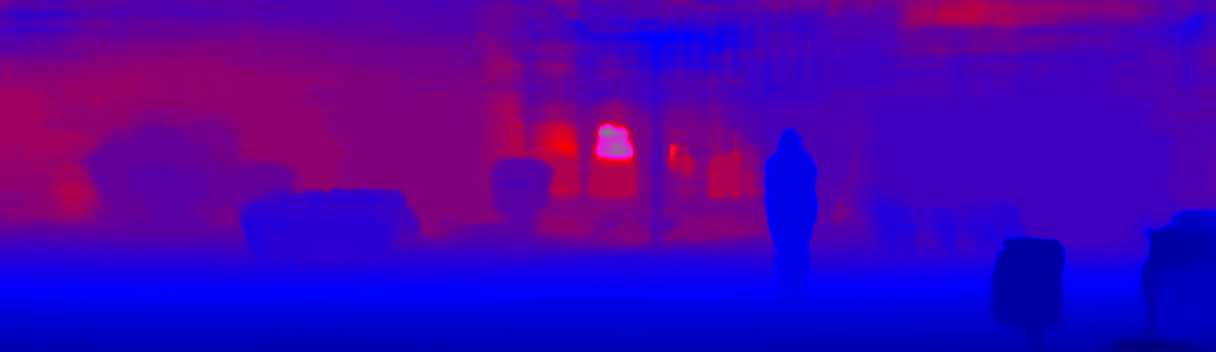}
\end{subfigure}
\begin{subfigure}{0.49\columnwidth}
  \centering
  \includegraphics[width=1\columnwidth, trim={0cm 0cm 0cm 0cm}, clip]{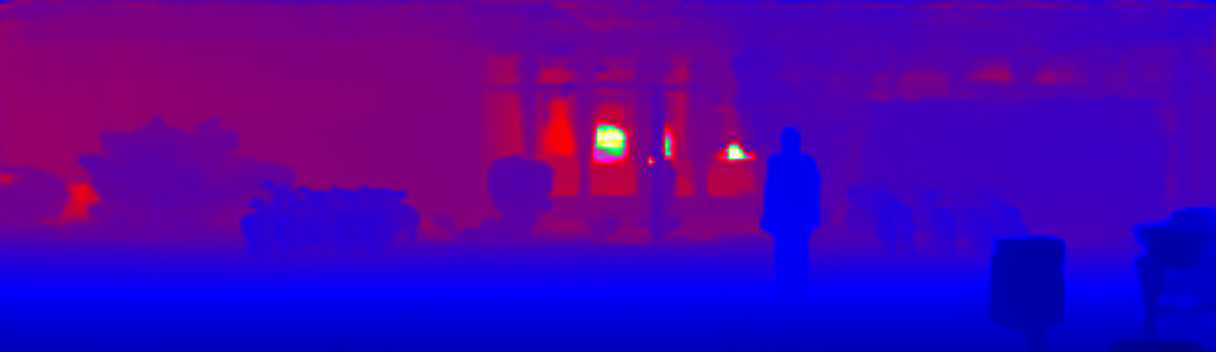}
\end{subfigure}
\begin{subfigure}{0.49\columnwidth}
  \centering
  \includegraphics[width=1\columnwidth, trim={0cm 0cm 0cm 0cm}, clip]{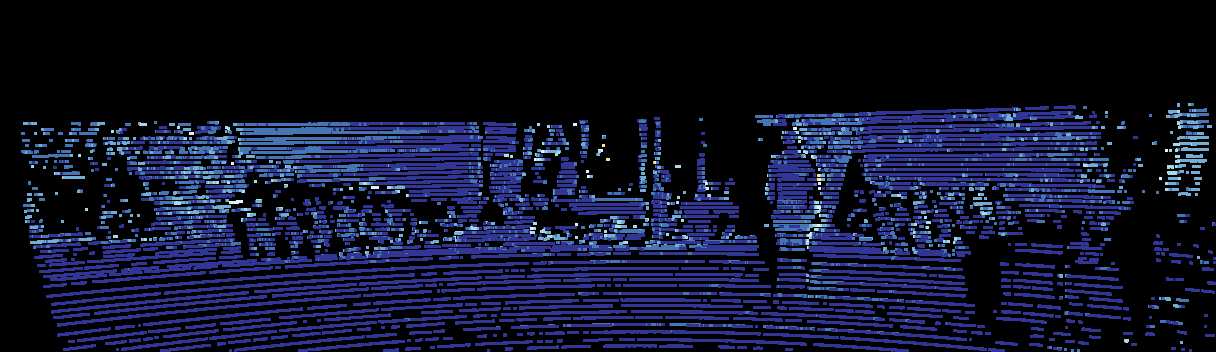}
\end{subfigure}

\begin{subfigure}{0.49\columnwidth}
  \centering
  \includegraphics[width=1\columnwidth, trim={0cm 0cm 0cm 0cm}, clip]{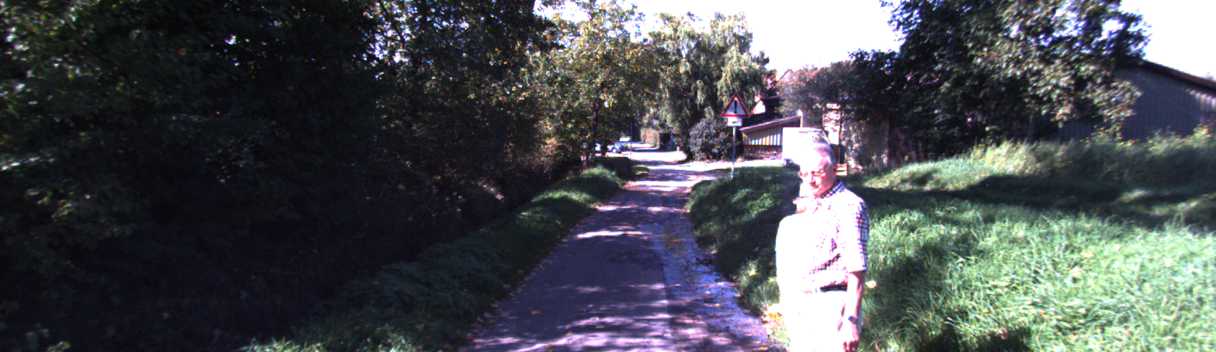}
  \caption*{Input image}
\end{subfigure}
\begin{subfigure}{0.49\columnwidth}
  \centering
  \includegraphics[width=1\columnwidth, trim={0cm 0cm 0cm 0cm}, clip]{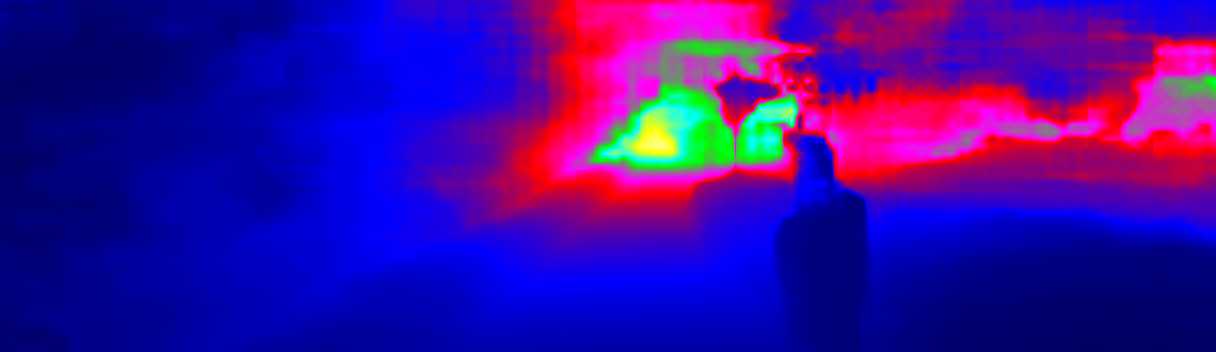}
  \caption*{DORN}
\end{subfigure}
\begin{subfigure}{0.49\columnwidth}
  \centering
  \includegraphics[width=1\columnwidth, trim={0cm 0cm 0cm 0cm}, clip]{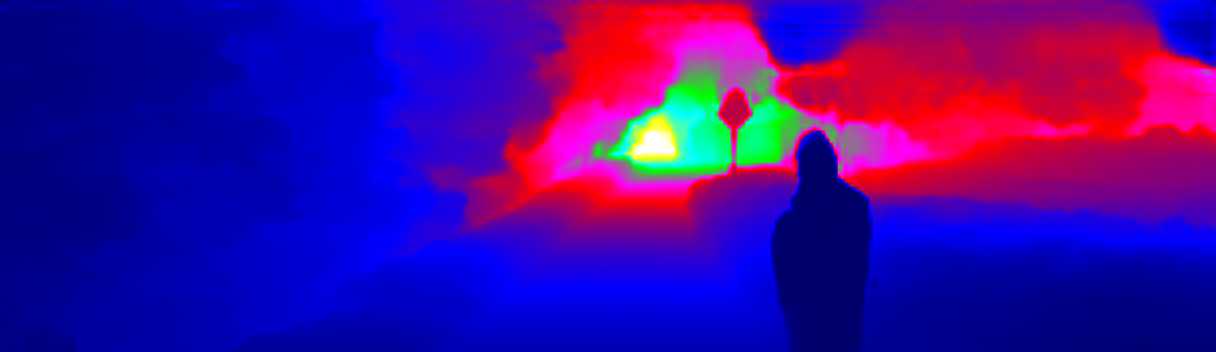}
  \caption*{Ours}
\end{subfigure}
\begin{subfigure}{0.49\columnwidth}
  \centering
  \includegraphics[width=1\columnwidth, trim={0cm 0cm 0cm 0cm}, clip]{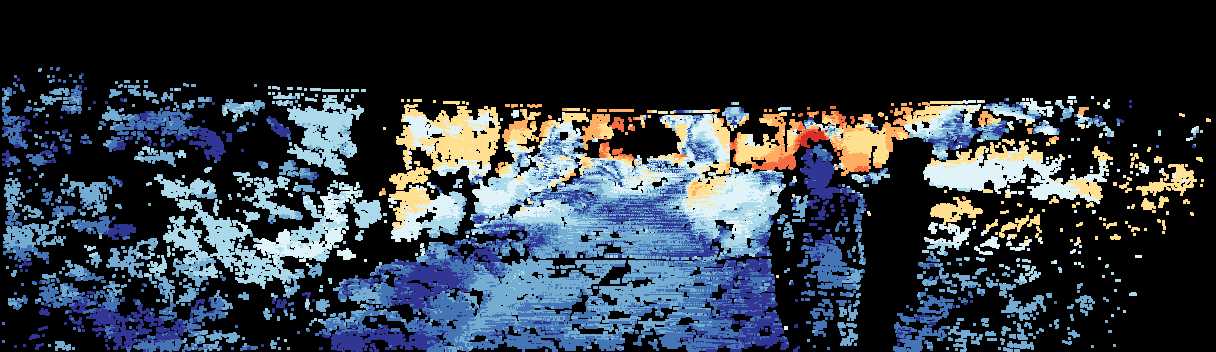}
  \caption*{Error map}
\end{subfigure}

\vspace{-2mm}
\caption{Qualitative results on the KITTI online benchmark, which are generated by the online server.}
\label{fig:res-kitti}
\vspace{-2mm}
\end{figure*}

\subsection{Neural Window FC-CRFs}

In traditional CRFs, the unary potential is usually acted by a distribution over the predicted values, e.g., 
\begin{equation}
    \psi_u(x_i) = -\log P (x_i|I),
\end{equation}
where $I$ is the input color image and $P$ is the probability distribution of the value prediction.
The pairwise potential is usually computed according to the colors and positions of pixel pairs, e.g.,
\begin{equation}
    \psi_{p}(x_i, x_j) = \mu(x_i, x_j)||x_i-x_j||
                e^{-\frac{||I_i-I_j||}{2\sigma^2}}
                e^{-\frac{||p_i-p_j||}{2\sigma^2}}.
\end{equation}
This potential encourages distinct-color and distant pixels to have various value predictions while punishing the value discrepancies in similar-color and adjacent pixels. 

These potential functions are designed by hands and cannot be too complicated. Thus they are hard to represent high-dimensional information and describe complex connections. So in this work, we propose to use neural networks to perform the potential functions.
 
For the unary potential, it is computed from the image features such that it can be directly obtained by the network as
\begin{equation}
    \psi_u(x_i) = \theta_u (I, x_i),
\end{equation}
where $\theta$ is the parameters of a unary network.

For the pairwise potential, we realize that it is composed of values of the current node and other nodes, and a weight computed based on the color and position information of the node pairs. So we reformulate it as
\begin{equation}
    \psi_{p}(x_i, x_j) = w(\mathcal{F}_i, \mathcal{F}_j, p_i, p_j) ||x_i - x_j||,
\label{eq:pairwise1}
\end{equation}
where $\mathcal{F}$ is the feature map and $w$ is the weighting function. 
We calculate the pairwise potential node by node. For each node $i$, we sum all its pairwise potentials and obtain
\begin{equation}
    \psi_{p_i} = \alpha (\mathcal{F}_i, \mathcal{F}_j, p_i, p_j) x_i + \sum_{j\neq i} \beta(\mathcal{F}_i, \mathcal{F}_j, p_i, p_j) x_j,
\label{eq:pairwise}
\end{equation}
where $\alpha, \beta$ are the weighting functions and will be computed by the networks.

Inspired by recent works in transformer \cite{vaswani2017attention, dosovitskiy2020image}, we calculate a query vector $\mathbf{q}$ and a key vector $\mathbf{k}$ from the feature map of each patch in a window and combine vectors of all patches to matrices $Q$ and $K$.
Then we calculate the dot product of matrices $Q$ and $K$ to get the potential weight between any pair, after which the predicted values $X$ are multiplied by the weights to get the final pairwise potential. 
To introduce the position information, we also add a relative position embedding $P$. Therefore, the equation~\ref{eq:pairwise} can be calculated as

\begin{equation}
\begin{aligned}
    \psi_{p_i} = & \ \text{SoftMax}(q \cdot K^T + P) \cdot X   \\
    \sum_i \psi_{p_i} = & \ \text{SoftMax}(Q \cdot K^T + P) \cdot X  ,
\label{eq:neural}
\end{aligned}
\end{equation}
where $\cdot$ denotes dot production. Thus, the output of the SoftMax gets the weights $\alpha$ and $\beta$ of Equation~\ref{eq:pairwise}. Therefore, the dot product between $Q$ and $K$ calculates the scores between each node with any other node, which determines the message passing weights with $P$, while the dot product between previous prediction $X$ and the output of the SoftMax performs the message passing.

\subsection{Network Structure}

\paragraph{Overview.} 
To embed the neural window fully-connected CRFs into a depth prediction network, we build a bottom-up-top-down structure, where four levels of CRFs optimizations are performed, as is shown in Figure~\ref{fig:net}. We embed this neural window FC-CRFs module into the network to act as a decoder, which predicts the next-level depth according to the coarse depth and image features.
For the encoder, we employ the swin-transformer\cite{liu2021swin} to extract the features.

For an image with the size of $H \times W$, there are four levels of image patches for the feature extraction encoder and the CRFs optimization decoder, from $4\times 4$ pixels to $32\times 32$ pixels.
At each level, $N \times N$ patches make up a window. The window size $N$ is fixed at all levels, so there will be $\frac{H}{4N}\times \frac{W}{4N}$ windows at the bottom level and $\frac{H}{32N}\times \frac{W}{32N}$ windows at the top level. 

\vspace{-3mm}
\paragraph{Global Information Aggregation.}
At the top level, to make up for the lack of global information of the window FC-CRFs, we use the pyramid pooling module (PPM) \cite{zhao2017pyramid} to aggregate the information of the whole image. Similar to \cite{zhao2017pyramid}, we use global averaging pooling of scales $1, 2, 3, 6$ to extract the global information, which is then concatenated with the input feature to map to the top-level prediction $X$ by a convolutional layer.

\vspace{-3mm}
\paragraph{Neural Window FC-CRFs Module.}
In each neural window FC-CRFs block, there are two successive CRFs optimizations, one for regular windows and the other one for shifted windows. 
To cooperate with the transformer encoder, the window size $N$ is set to $7$, which means each window contains $7 \times 7$ patches.
The unary potential is computed by a convolutional network and the pairwise potential is computed according to equation~\ref{eq:neural}.
In each CRFs optimization, multiple-head $Q$ and $K$ are calculated to obtain multi-head potentials, which can enhance the relationship capturing ability of the energy function. 
From the top level to the bottom level, a structure of $32, 16, 8, 4$ heads is adopted.
Then the energy function is fed into an optimization network composed of two fully-connected layers to output the optimized depth map $X'$. 

\vspace{-3mm}
\paragraph{Upscale Module.}
After the neural window FC-CRFs decoders at the top three levels, a shuffle operation is performed to rearrange the pixels, by which the image is upscaled from $\frac{h}{2} \times \frac{w}{2} \times d$ to $h \times w \times \frac{d}{4}$. On the one hand, this operation increases the flow to the next level with a larger scale without losing the sharpness like upsampling. On the other hand, this reduces the feature dimension to lighten the subsequent networks. 


\vspace{-3mm}
\paragraph{Training Loss.}

Following previous works~\cite{lee2019big, bhat2021adabins, lee2021patch}, we use a Scale-Invariant Logarithmic (SILog) loss proposed by \cite{eigen2014depth} to supervise the training. Given the ground-truth depth map, we first calculate the logarithm difference between the predicted depth map and the real depth:
\begin{equation}
    \Delta d_i = \log \hat{d_i} - \log d^*_i,
\end{equation}
where $d^*_i$ is the ground-truth depth value and $\hat{d_i}$ is the predicted depth at pixel $i$.

Then for $K$ pixels with valid depth values in an image, the scale-invariant loss is computed as
\begin{equation}
    \mathcal{L} = \alpha \sqrt{\frac{1}{K}\sum_i \Delta d_i^2 - \frac{\lambda}{K^2}(\sum_i \Delta d_i)^2},
\end{equation}
where $\lambda$ is a variance minimizing factor, and $\alpha$ is a scale constant.
In our experiments, $\lambda$ is set to $0.85$ and $\alpha$ is set to $10$ following previous works~\cite{lee2019big}.

\setlength{\tabcolsep}{3pt}
\begin{table*}[]
\small
\centering
\begin{tabular}{r c c c c c c c c c c}
\toprule
Method & cap & Abs Rel $\downarrow$ & Sq Rel $\downarrow$ & RMSE $\downarrow$ & RMSE$_{log}$ $\downarrow$ & $\delta<1.25$ $\uparrow$ & $\delta<1.25^2$ $\uparrow$ & $\delta<1.25^3$ $\uparrow$  \\
\midrule
Eigen et al.~\cite{eigen2014depth} & \ \  0-80m \ \  & $0.190$ & $1.515$ & $7.156$ & $0.270$ & $0.692$ & $0.899$ & $0.967$\\
Liu et al.~\cite{liu2015deep} & 0-80m & $0.217$ & $-$ & $7.046$ & $-$ & $0.656$ & $0.881$ & $0.958$\\
Xu et al.~\cite{xu2018structured} & 0-80m & $0.122$ & $0.897$ & $4.677$ & $-$ & $0.818$ & $0.954$ & $0.985$\\
DORN~\cite{fu2018deep} & 0-80m & $0.072$ & $0.307$ & $2.727$ & $0.120$ & $0.932$ & $0.984$ & $0.995$\\
Yin et al.~\cite{yin2019enforcing} & 0-80m & $0.072$ & $-$ & $3.258$ & $0.117$ & $0.938$ & $0.990$ & $0.998$\\
BTS~\cite{lee2019big} & 0-80m & $0.059$ & $0.241$ & $2.756$ & $0.096$ & $0.956$ & $0.993$ & $0.998$\\
PackNet-SAN~\cite{guizilini2021sparse} & 0-80m & $0.062$ & $-$ & $2.888$ & $-$ & $0.955$ & $-$ & $-$\\
Adabin~\cite{bhat2021adabins} & 0-80m & $0.058$ & $0.190$ & $2.360$ & $0.088$ & $0.964$ & $0.995$ & $\mathbf{0.999}$\\
DPT*~\cite{ranftl2021vision} & 0-80m & $0.062$ & $-$ & $2.573$ & $0.092$ & $0.959$ & $0.995$ & $\mathbf{0.999}$\\
PWA~\cite{lee2021patch} & 0-80m & $0.060$ & $0.221$ & $2.604$ & $0.093$ & $0.958$ & $0.994$ & $\mathbf{0.999}$\\
Ours & 0-80m & $\mathbf{0.052}$ & $\mathbf{0.155}$ & $\mathbf{2.129}$ & $\mathbf{0.079}$ & $\mathbf{0.974}$ & $\mathbf{0.997}$ & $\mathbf{0.999}$\\

\bottomrule
\end{tabular}
\vspace{-2mm}
\caption{Quantitative results on the Eigen split of KITTI dataset. Seven widely used metrics are reported. ``Abs Rel" error is the main ranking metric. Note that the ``Sq Rel" error is calculated in a different way here. ``*" means using additional data for training.}
\label{tab:kitti_eigen}
\vspace{-2mm}
\end{table*}
\setlength{\tabcolsep}{3pt}

\setlength{\tabcolsep}{3pt}
\begin{table*}[]
\small
\centering
\begin{tabular}{r c c c c c c c c c c}
\toprule
Method & dataset & SILog $\downarrow$ & Abs Rel $\downarrow$ & Sq Rel $\downarrow$ & iRMSE $\downarrow$ & RMSE $\downarrow$ & $\delta<1.25$ $\uparrow$ & $\delta<1.25^2$ $\uparrow$ & $\delta<1.25^3$ $\uparrow$ \\
\midrule
DORN~\cite{fu2018deep} & val  & $12.22$ & $11.78$ & $3.03$ & $11.68$ & $3.80$ & $0.913$ & $0.985$ & $0.995$\\
BTS~\cite{lee2019big} & val & $10.67$ & $7.51$ & $1.59$ & $8.10$ & $3.37$ & $0.938$ & $0.987$ & $0.996$\\
BA-Full~\cite{aich2020bidirectional} & val & $10.64$ & $8.25$ & $1.81$ & $8.47$ & $3.30$ & $0.938$ & $0.988$ & $0.997$\\
Ours & val & $\mathbf{8.31}$ & $\mathbf{5.54}$ & $\mathbf{0.89}$ & $\mathbf{6.34}$ & $\mathbf{2.55}$ & $\mathbf{0.968}$ & $\mathbf{0.995}$ & $\mathbf{0.998}$\\
\midrule
DORN~\cite{fu2018deep} & online test & $11.77$ & $8.78$ & $2.23$ & $12.98$ & $-$ & $-$ & $-$ & $-$\\
BTS~\cite{lee2019big} & online test  & $11.67$ & $9.04$ & $2.21$ & $12.23$ & $-$ & $-$ & $-$ & $-$\\
BA-Full~\cite{aich2020bidirectional} & online test  & $11.61$ & $9.38$ & $2.29$ & $12.23$ & $-$ & $-$ & $-$ & $-$\\
PackNet-SAN~\cite{guizilini2021sparse} & online test & $11.54$ & $9.12$ & $2.35$ & $12.38$ & $-$ & $-$ & $-$ & $-$\\
PWA~\cite{lee2021patch} & online test & $11.45$ & $9.05$ & $2.30$ & $12.32$ & $-$ & $-$ & $-$ & $-$\\
Ours & \ \ online test \ \  & $\mathbf{10.39}$ & $\mathbf{8.37}$ & $\mathbf{1.83}$ & $\mathbf{11.03}$ & $-$ & $-$ & $-$ & $-$\\

\bottomrule
\end{tabular}
\vspace{-2mm}
\caption{Quantitative results on the official split of KITTI dataset. Eight widely used metrics are reported for the validation set while only four metrics are available from the online evaluation server for the test set. ``SILog" error is the main ranking metric. Our method \textbf{ranks 1st} among all submissions on the KITTI depth prediction online benchmark at the submission time of this paper.}
\label{tab:kitti_official}
\vspace{-5mm}
\end{table*}
\setlength{\tabcolsep}{3pt}

\begin{figure*}[t]
\centering
\begin{subfigure}{0.3\columnwidth}
  \centering
  \includegraphics[width=1\columnwidth, trim={0cm 0cm 0cm 0cm}, clip]{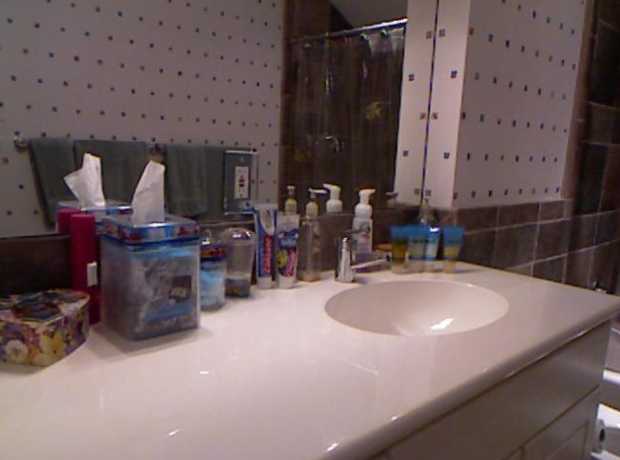}
\end{subfigure}
\begin{subfigure}{0.3\columnwidth}
  \centering
  \includegraphics[width=1\columnwidth, trim={0cm 0cm 0cm 0cm}, clip]{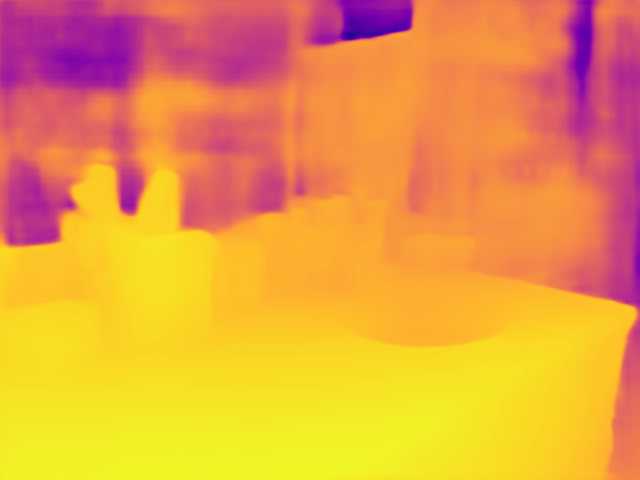}
\end{subfigure}
\begin{subfigure}{0.3\columnwidth}
  \centering
  \includegraphics[width=1\columnwidth, trim={0cm 0cm 0cm 0cm}, clip]{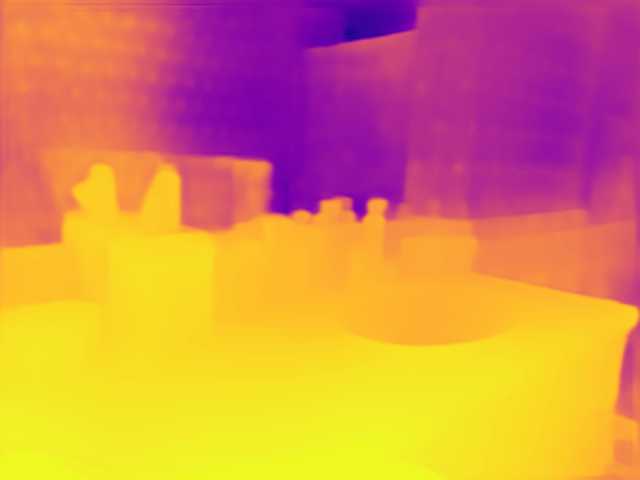}
\end{subfigure}
\begin{subfigure}{0.3\columnwidth}
  \centering
  \includegraphics[width=1\columnwidth, trim={0cm 0cm 0cm 0cm}, clip]{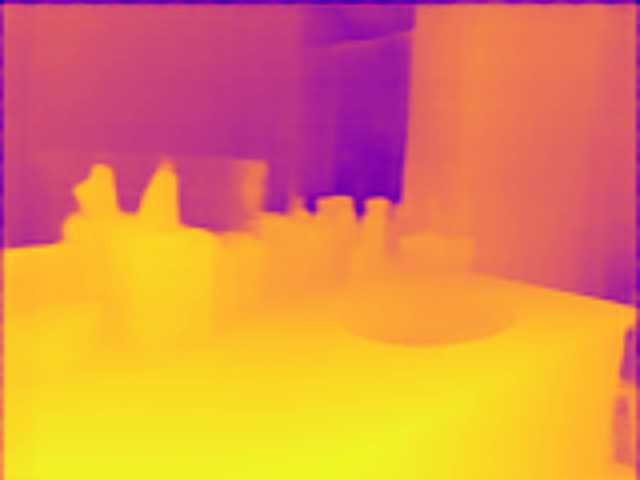}
\end{subfigure}
\begin{subfigure}{0.3\columnwidth}
  \centering
  \includegraphics[width=1\columnwidth, trim={0cm 0cm 0cm 0cm}, clip]{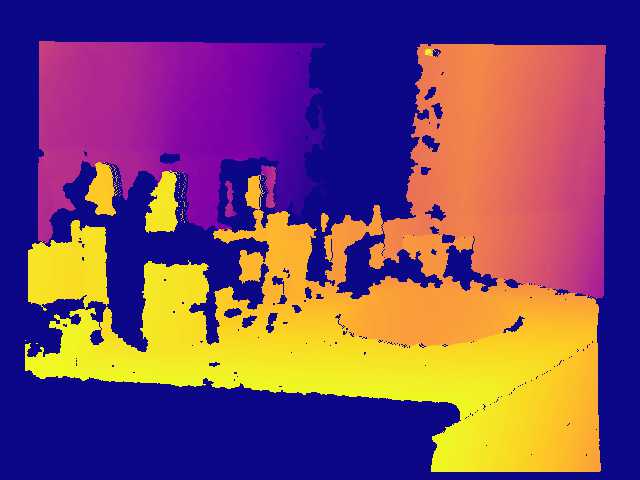}
\end{subfigure}

\begin{subfigure}{0.3\columnwidth}
  \centering
  \includegraphics[width=1\columnwidth, trim={0cm 0cm 0cm 0cm}, clip]{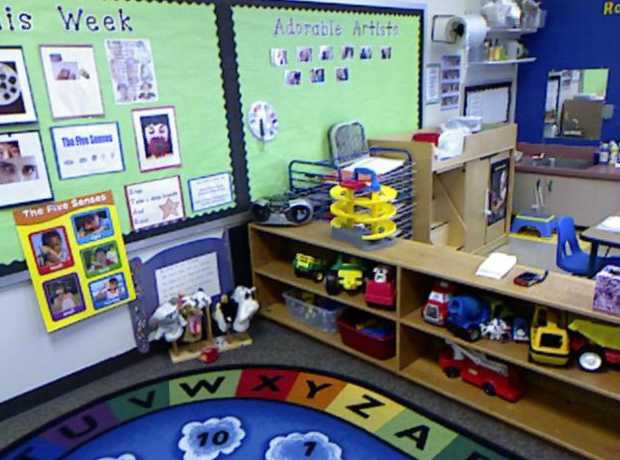}
\end{subfigure}
\begin{subfigure}{0.3\columnwidth}
  \centering
  \includegraphics[width=1\columnwidth, trim={0cm 0cm 0cm 0cm}, clip]{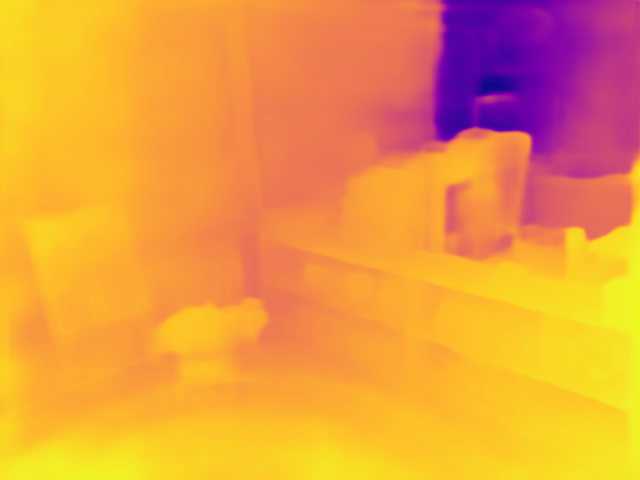}
\end{subfigure}
\begin{subfigure}{0.3\columnwidth}
  \centering
  \includegraphics[width=1\columnwidth, trim={0cm 0cm 0cm 0cm}, clip]{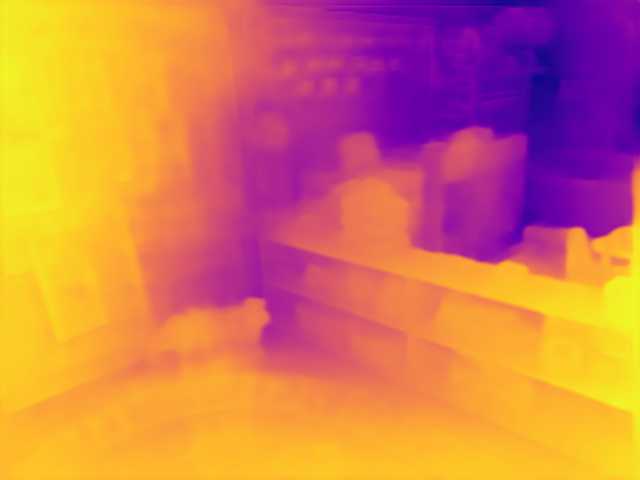}
\end{subfigure}
\begin{subfigure}{0.3\columnwidth}
  \centering
  \includegraphics[width=1\columnwidth, trim={0cm 0cm 0cm 0cm}, clip]{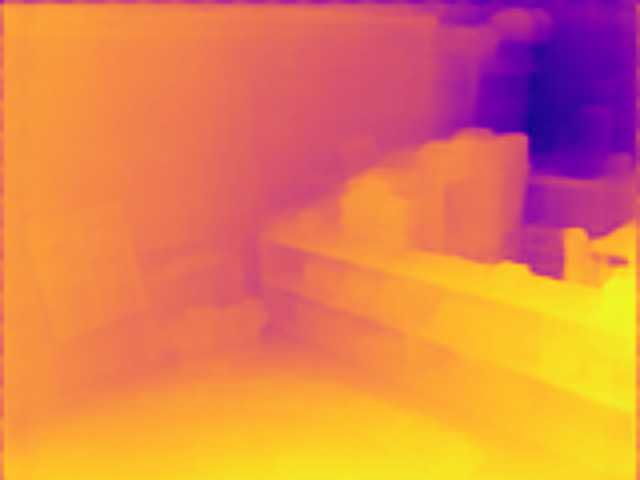}
\end{subfigure}
\begin{subfigure}{0.3\columnwidth}
  \centering
  \includegraphics[width=1\columnwidth, trim={0cm 0cm 0cm 0cm}, clip]{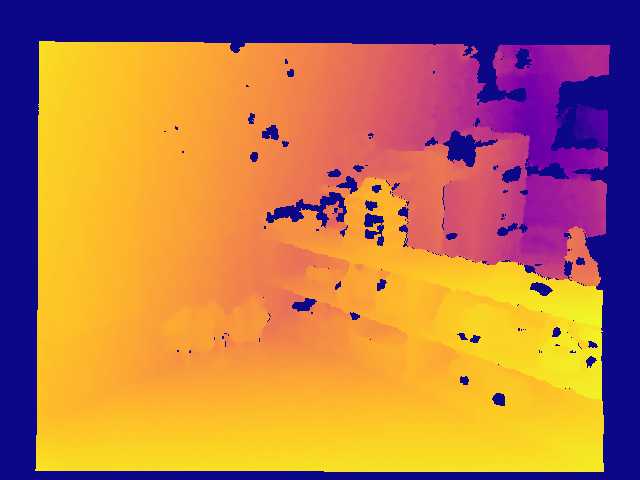}
\end{subfigure}


\begin{subfigure}{0.3\columnwidth}
  \centering
  \includegraphics[width=1\columnwidth, trim={0cm 0cm 0cm 0cm}, clip]{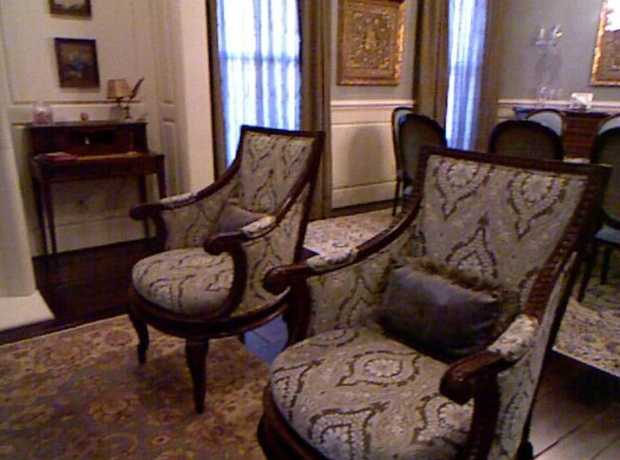}
\end{subfigure}
\begin{subfigure}{0.3\columnwidth}
  \centering
  \includegraphics[width=1\columnwidth, trim={0cm 0cm 0cm 0cm}, clip]{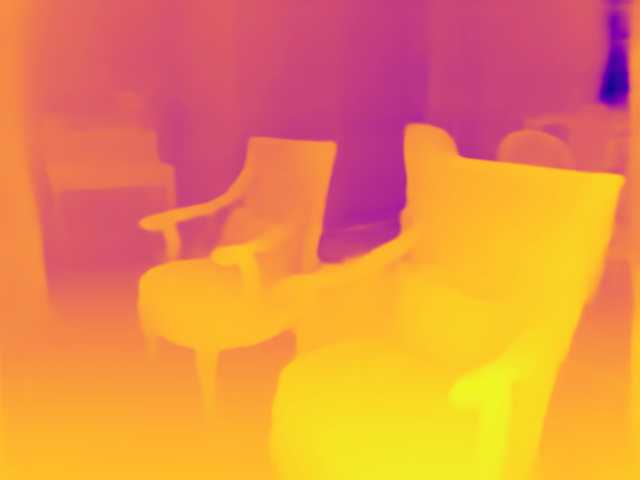}
\end{subfigure}
\begin{subfigure}{0.3\columnwidth}
  \centering
  \includegraphics[width=1\columnwidth, trim={0cm 0cm 0cm 0cm}, clip]{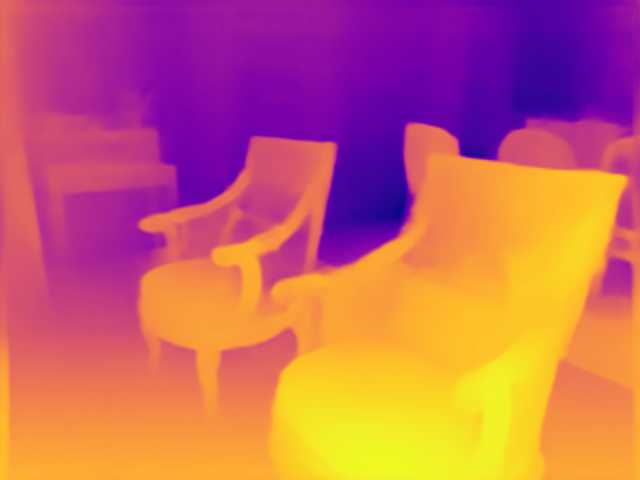}
\end{subfigure}
\begin{subfigure}{0.3\columnwidth}
  \centering
  \includegraphics[width=1\columnwidth, trim={0cm 0cm 0cm 0cm}, clip]{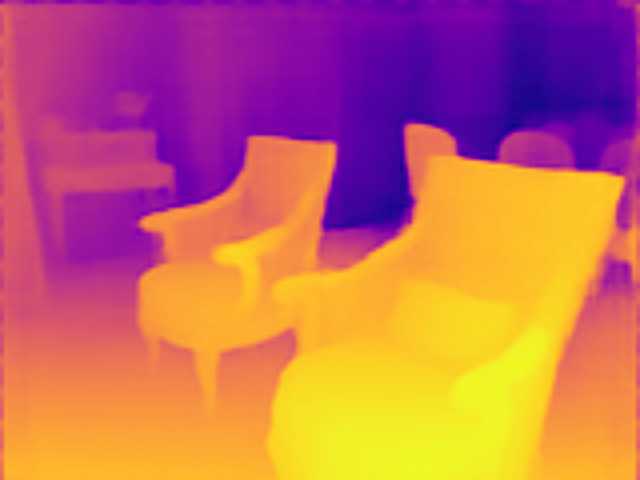}
\end{subfigure}
\begin{subfigure}{0.3\columnwidth}
  \centering
  \includegraphics[width=1\columnwidth, trim={0cm 0cm 0cm 0cm}, clip]{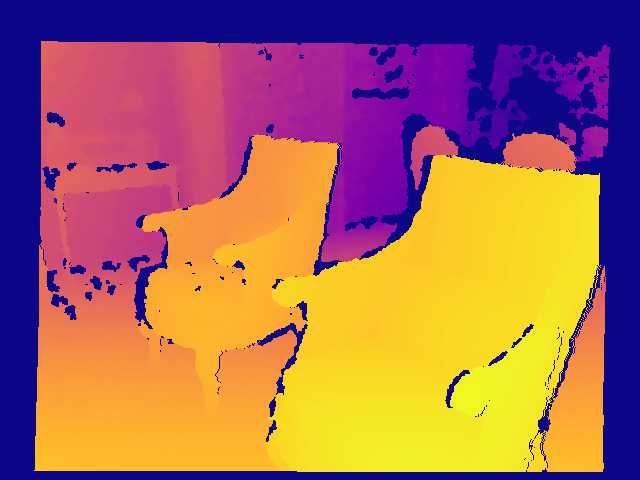}
\end{subfigure}

\begin{subfigure}{0.3\columnwidth}
  \centering
  \includegraphics[width=1\columnwidth, trim={0cm 0cm 0cm 0cm}, clip]{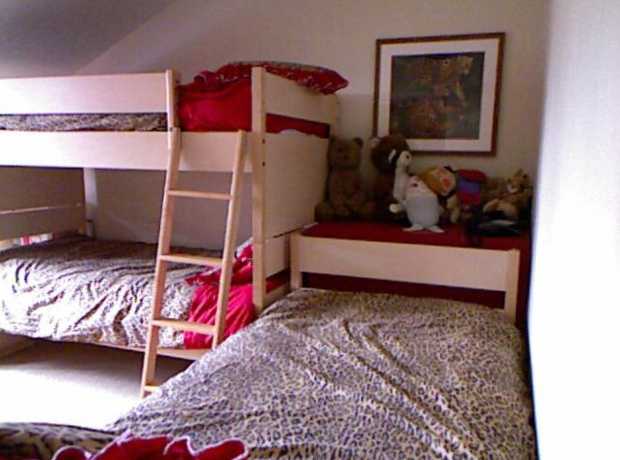}
  \caption*{Input image}
\end{subfigure}
\begin{subfigure}{0.3\columnwidth}
  \centering
  \includegraphics[width=1\columnwidth, trim={0cm 0cm 0cm 0cm}, clip]{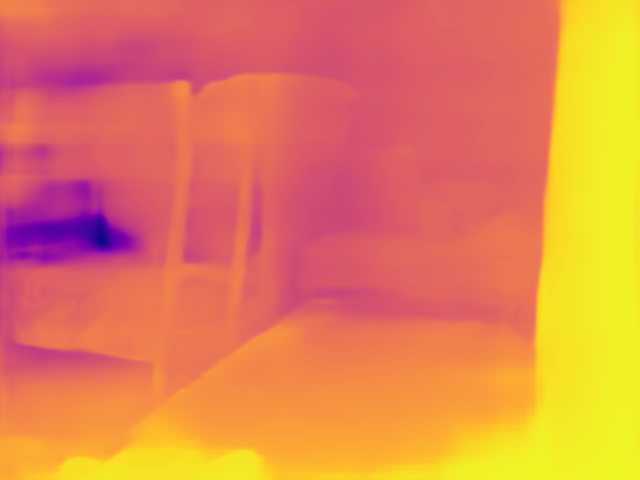}
  \caption*{BTS}
\end{subfigure}
\begin{subfigure}{0.3\columnwidth}
  \centering
  \includegraphics[width=1\columnwidth, trim={0cm 0cm 0cm 0cm}, clip]{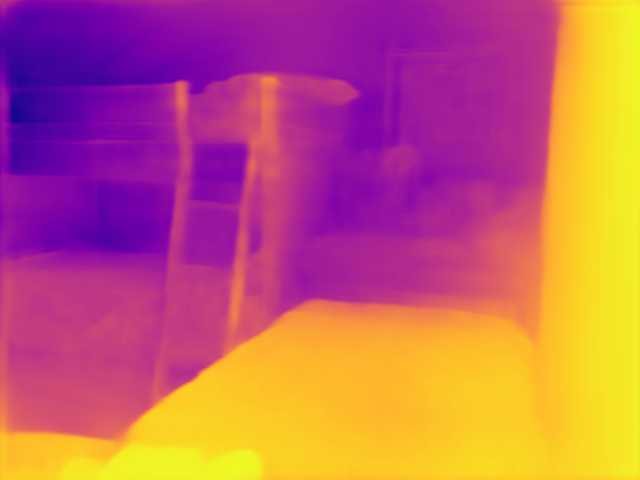}
  \caption*{Adabins}
\end{subfigure}
\begin{subfigure}{0.3\columnwidth}
  \centering
  \includegraphics[width=1\columnwidth, trim={0cm 0cm 0cm 0cm}, clip]{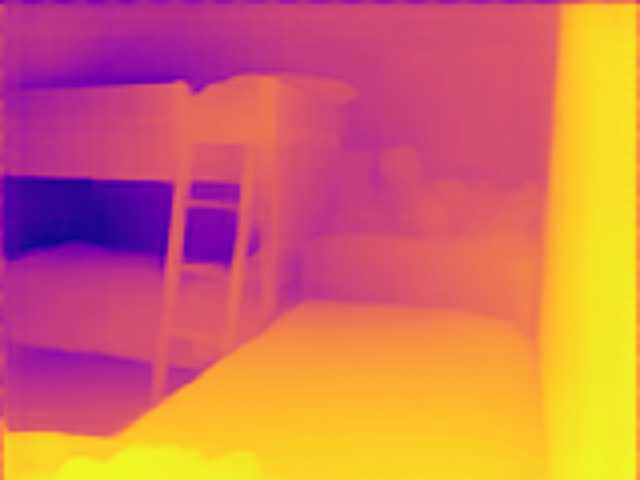}
  \caption*{Ours}
\end{subfigure}
\begin{subfigure}{0.3\columnwidth}
  \centering
  \includegraphics[width=1\columnwidth, trim={0cm 0cm 0cm 0cm}, clip]{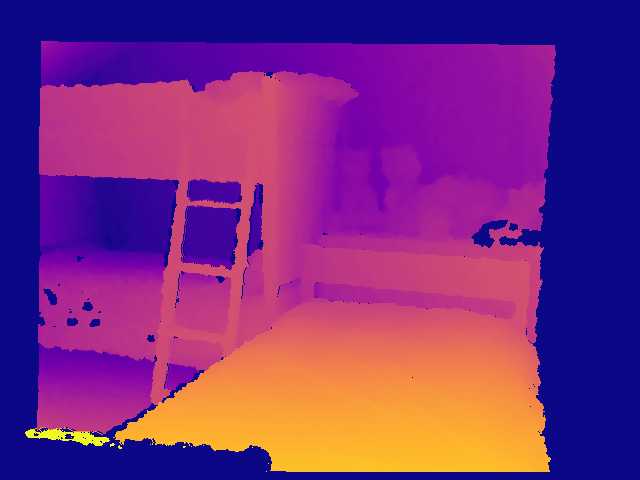}
  \caption*{Ground truth}
\end{subfigure}

\vspace{-3mm}
\caption{Qualitative results on the NYUv2 dataset.}
\label{fig:res-nyu}
\vspace{-4mm}
\end{figure*}

\setlength{\tabcolsep}{3pt}
\begin{table*}[]
\small
\centering
\begin{tabular}{r c c c c c c c c c c}
\toprule
Method & Abs Rel $\downarrow$ & Sq Rel $\downarrow$ & RMSE $\downarrow$ & RMSE$_{log}$ $\downarrow$ & log10 $\downarrow$ & $\delta<1.25$ $\uparrow$ & $\delta<1.25^2$ $\uparrow$ & $\delta<1.25^3$ $\uparrow$  \\
\midrule
Liu et al.~\cite{liu2015deep} & $0.230$ & $-$ & $0.824$ & $-$ & $0.095$ & $0.614$ & $0.883$ & $0.971$\\
Xu et al.~\cite{xu2018structured} & $0.125$ & $-$ & $0.593$ & $-$ & $0.057$ & $0.806$ & $0.952$ & $0.986$\\
DORN~\cite{fu2018deep} & $0.115$ & $-$ & $0.509$ & $-$ & $0.051$ & $0.828$ & $0.965$ & $0.992$\\
Yin et al.~\cite{yin2019enforcing} & $0.108$ & $-$ & $0.416$ & $-$ & $0.048$ & $0.875$ & $0.976$ & $0.994$\\
BTS~\cite{lee2019big} & $0.110$ & $0.066$ & $0.392$ & $0.142$ & $0.047$ & $0.885$ & $0.978$ & $0.994$\\
DAV~\cite{huynh2020guiding} & $0.108$ & $-$ & $0.412$ & $-$ & $-$ & $0.882$ & $0.980$ & $0.996$\\
PackNet-SAN*~\cite{guizilini2021sparse} & $0.106$ & $-$ & $0.393$ & $-$ & $-$ & $0.892$ & $0.979$ & $0.995$\\
Adabin~\cite{bhat2021adabins} & $0.103$ & $-$ & $0.364$ & $-$ & $0.044$ & $0.903$ & $0.984$ & $0.997$\\
DPT*~\cite{ranftl2021vision} & $0.110$ & $-$ & $0.357$ & $-$ & $0.045$ & $0.904$ & $0.988$ & $\mathbf{0.998}$\\
PWA~\cite{lee2021patch} & $0.105$ & $-$ & $0.374$ & $-$ & $0.045$ & $0.892$ & $0.985$ & $0.997$\\
Ours & $\mathbf{0.095}$ & $\mathbf{0.045}$ & $\mathbf{0.334}$ & $\mathbf{0.119}$ & $\mathbf{0.041}$ & $\mathbf{0.922}$ & $\mathbf{0.992}$ & $\mathbf{0.998}$\\
\bottomrule
\end{tabular}
\vspace{-2mm}
\caption{Quantitative results on NYUv2. ``Abs Rel" and ``RMSE" are the main ranking metrics. ``*" means using additional data.}
\label{tab:nyu}
\vspace{-4mm}
\end{table*}
\setlength{\tabcolsep}{3pt}

\section{Experiments}
\label{sec:experiments}


\subsection{Implementation Details}

Our work is implemented in Pytorch and experimented on Nvidia GTX 2080 Ti GPUs. The network is optimized end-to-end with the Adam optimizer ($\beta_1=0.9$, $\beta_1=0.999$). The training runs for $20$ epochs with the batch size of $8$ and the learning rate decreasing from $1\times 10^{-4}$ to $1\times 10^{-5}$. 
The output depth map of our network is of $\frac{1}{4} \times \frac{1}{4}$ size of the original image, which is then resized to the full resolution.

\subsection{Datasets}

\textbf{KITTI dataset.}
KITTI dataset~\cite{geiger2012we} is the most used benchmark with outdoor scenes captured from a moving vehicle. 
There are two mainly used splits for monocular depth estimation.
One is the training/testing split proposed by Eigen et al.~\cite{eigen2014depth} with $23488$ training image pairs and $697$ testing images. The other one is the official split proposed by Geiger et al.~\cite{geiger2012we} with $42949$ training image pairs, $1000$ validation images, and $500$ testing images. For the official split, the ground-truth depth maps for the testing images are withheld by the online evaluation benchmark. 


\textbf{NYUv2 dataset.}
NYUv2~\cite{silberman2012indoor} is an indoor datasets with $120K$ RGB-D videos captured from $464$ indoor scenes. We follow the official training/testing split to evaluate our method, where $249$ scenes are used for training and $654$ images from $215$ scenes are used for testing. 

\textbf{MatterPort3D dataset.}
To verify the effectiveness of our method on more domains, we also evaluate our method on the panorama images. MatterPort3D~\cite{chang2017matterport3d} is the biggest real-world dataset among all widely used datasets in panorama depth estimation. Following the official split, we use $7829$ images from $61$ houses to train our network and then evaluate the model on the merged set of $957$ validation images and $2014$ testing images. 
All images are resized to $1024 \times 512$ in both training and evaluation.

\setlength{\tabcolsep}{4pt}
\begin{table*}[]
\small
\centering
\begin{tabular}{r c c c c c c c c c c}
\toprule
Method & Abs Rel $\downarrow$ & Abs $\downarrow$ & RMSE $\downarrow$ & RMSE$_{log}$ $\downarrow$ & $\delta<1.25$ $\uparrow$ & $\delta<1.25^2$ $\uparrow$ & $\delta<1.25^3$ $\uparrow$  \\
\midrule
OmniDepth~\cite{zioulis2018omnidepth} & $0.2901$ & $0.4838$ & $0.7643$ & $0.1450$ & $0.6830$ & $0.8794$ & $0.9429$\\
BiFuse~\cite{wang2020bifuse} & $0.2048$ & $0.3470$ & $0.6259$ & $0.1134$ & $0.8452$ & $0.9319$ & $0.9632$\\
SliceNet~\cite{pintore2021slicenet} & $0.1764$ & $0.3296$ & $0.6133$ & $0.1045$ & $0.8716$ & $0.9483$ & $0.9716$\\
HoHoNet~\cite{sun2021hohonet} & $0.1488$ & $0.2862$ & $0.5138$ & $0.0871$ & $0.8786$ & $0.9519$ & $0.9771$\\
UniFuse~\cite{jiang2021unifuse} & $0.1063$ & $0.2814$ & $0.4941$ & $0.0701$ & $0.8897$ & $0.9623$ & $0.9831$\\
Ours & $\mathbf{0.0906}$ & $\mathbf{0.2252}$ & $\mathbf{0.4778}$ & $\mathbf{0.0638}$ & $\mathbf{0.9197}$ & $\mathbf{0.9761}$ & $\mathbf{0.9909}$\\
\midrule
Ours* & $\mathbf{0.0793}$ & $\mathbf{0.1970}$ & $\mathbf{0.4279}$ & $\mathbf{0.0575}$ & $\mathbf{0.9376}$ & $\mathbf{0.9812}$ & $\mathbf{0.9933}$\\
\bottomrule
\end{tabular}
\vspace{-2mm}
\caption{Quantitative results on the Matterport3D dataset. ``*" means using additional data for training.}
\label{tab:matterport}
\vspace{-5mm}
\end{table*}
\setlength{\tabcolsep}{4pt}

\subsection{Evaluations}

\textbf{Evaluation on KITTI. }
For outdoor scenes, we evaluate our method on the KITTI dataset. 
We first perform the training and testing on the Eigen split, of which the testing images are available so that the network can be better tuned.
The results are reported in Table~\ref{tab:kitti_eigen}, where we can see that our method outperforms previous methods by a significant margin. 
Almost all errors are reduced by about $10\%$. 
Specifically, the ``Abs-Rel", ``Sq Rel", ``RMSE" and ``RMSE$_{log}$" errors are decreased by $10.3\%$, $18.4\%$, $9.8\%$, and $10.2\%$, respectively.
Although our method is trained without additional data, it can outperform previous methods trained with additional training data.

We then evaluate our method on the KITTI official split, where the testing images are hidden. The results on the validation set and the testing set are all presented in Table~\ref{tab:kitti_official}. The results of the testing set are cited from the online benchmark and the results of the validation set are cited from BANet~\cite{aich2020bidirectional}.
Here we can see that our method reduces the main ranking metric, the SILog error, markedly. 
Our method now ranks 1st among all submissions on the KITTI depth prediction online server.
The colorful visualizations of the predicted depth maps and the error maps generated by the online server are shown in Figure~\ref{fig:res-kitti}. Our method predicts cleaner and smoother depth while maintaining sharper edges of objects, e.g., the edges of the humans.

\textbf{Evaluation on NYUv2. }
For indoor scenes, we evaluate our method on the NYUv2 dataset. 
Since the state-of-the-art performance on NYUv2 dataset has been saturated for a while, some methods have begun to use additional data to pretrain the model and then finetune it on NYUv2 training set~\cite{guizilini2021sparse, ranftl2021vision}. 
Differently, without any additional data, our method can significantly improve the performance in all metrics, as is shown in Table~\ref{tab:nyu}. 
Specifically, the ``Abs Rel" error is reduced to within $0.1$ and the ``$\delta<1.25^2$" accuracy reaches $99\%$.
This emphasizes the contribution of our method in improving the results.
The qualitative results in Figure~\ref{fig:res-nyu} illustrate that our method estimates better depth especially in difficult regions, such as repeated texture, messy environment, and bad light.

\textbf{Evaluation on MatterPort3D. }
As is studied in previous works, directly applying a deep network for perspective images to the standard representation of spherical panoramas, i.e., the equirectangular projection, is suboptimal, as it becomes distorted towards the poles~\cite{wang2020bifuse, jiang2021unifuse, sun2021hohonet, tateno2018distortion}. 
As such, methods in this task try all kinds of ways to convert the panorama images to distortion-free shape, e.g., the cubemap projection~\cite{wang2020bifuse, jiang2021unifuse}, the horizontal feature representation~\cite{sun2021hohonet}, and spherical convolutional filters~\cite{tateno2018distortion}.
In comparison to the above-mentioned methods, we directly apply our network designed for perspective images to the panorama images, and outperforms all previous methods, as is presented in Table~\ref{tab:matterport}. Specifically, the ``Abs Rel" and ``Abs" errors are decreased by $14.8\%$ and $20.0\%$.

In addition, we realize that the number of the training set of MatterPort3D is small, so we collect more data in the real world. We use $50K$ images to pretrain the network and then finetune it on the MatterPort3D training set, which results in a better performance, as shown in Table~\ref{tab:matterport}. The model pretrained with more data is denoted by ``Ours*". This demonstrates the pretraining with more images can clearly boost the performance in panorama depth estimation.

\setlength{\tabcolsep}{3pt}
\begin{table}[]
\small
\centering
\begin{tabular}{r c c c c c c c}
\toprule
Setting & Abs Rel & Sq Rel & RMSE & R$_{log}$ & $1.25$ & $1.25^2$  \\
\midrule
Baseline & $0.069$ & $0.256$ & $2.610$ & $0.103$ & $0.947$ & $0.993$  \\
Neural CRFs & $0.055$ & $0.185$ & $2.322$ & $0.086$ & $0.965$ & $0.995$  \\
+ S & $0.054$ & $0.174$ & $2.297$ & $0.084$ & $0.968$ & $0.996$  \\
+ S + R & $0.054$ & $0.168$ & $2.271$ & $0.083$ & $0.970$ & $0.996$  \\
+ S + R + P & $0.052$ & $0.155$ & $2.129$ & $0.079$ & $0.974$ & $0.997$  \\
\midrule
8, 4, 2, 1 & $0.055$ & $0.165$ & $2.203$ & $0.083$ & $0.970$ & $0.996$   \\
16, 8, 4, 2 & $0.054$ & $0.162$ & $2.172$ & $0.081$ & $0.972$ & $0.997$   \\
32, 16, 8, 4 & $0.052$ & $0.155$ & $2.129$ & $0.079$ & $0.974$ & $0.997$  \\
\bottomrule
\end{tabular}
\caption{Ablation study on the Eigen split of KITTI dataset. The first six metrics of those used in Table~\ref{tab:kitti_eigen} are reported here. ``S" refers to window shift, ``R" refers to rearrange upscale, and ``P" refers to PPM head. The last three rows display the results of using different numbers of heads.}
\label{tab:ablation}
\vspace{-5mm}
\end{table}
\setlength{\tabcolsep}{3pt}


\subsection{Abalation Study}

To better inspect the effect of each module in our method, we evaluate each component by an ablation study and present the results in Table~\ref{tab:ablation}.

\textbf{Baseline vs. Neural CRFs. } To verify the effectiveness of the proposed neural window fully-connected FC-CRFs, we build a baseline model. This model is a well-used UNet structure with the same encoder as ours. In other words, compared to our full method, the PPM head and the rearrange upscale are removed, and the decoder is replaced by the well-used convolutional decoder.
Then based on this baseline, we only replace the decoder with our neural window FC-CRFs module, and obtain a noticeable performance improvement as shown in Table~\ref{tab:ablation}.
The ``Abs Rel" error is reduced from $0.069$ to $0.055$, and then to $0.054$ by adding the shift action.
This demonstrates the effectiveness of the neural window FC-CRFs in estimating accurate depths.

\textbf{Rearrange upscale. } On top of the basic neural FC-CRFs structure, we add the rearrange upscale module. The performance increment gained from this module is not large, but visually the output depth maps have sharper edges, and the parameters of the network are reduced.

\textbf{PPM head. } The PPM head aggregates the global information, which is lacking in window FC-CRFs. This module can help in some regions that are difficult for estimating with only local information, e.g., the complex texture and the white walls. From the results in Table~\ref{tab:ablation}, we see this module contributes to the performance of our framework. 


\textbf{Multi-head energy. } 
The CRFs energy is calculated in a multi-head manner. With more heads, the ability of capturing the pairwise relationship would be stronger but the weight of the network would be heavier. In previous experiments, the numbers of the heads in four levels are set to 32, 16, 8, 4. Here we use fewer heads to see how a lightweight structure performs. From the results in Table~\ref{tab:ablation}, fewer heads lead to a small performance decrease.

\section{Conclusion}

We propose a neural window fully-connected CRFs module to address the monocular depth estimation problem. To solve the expensive computation of FC-CRFs, we split the input into sub-windows and calculate the pairwise potential within each window. To capture the relationships between nodes of the graph, we exploit the multi-head attention to compute a neural potential function. 
This neural window FC-CRFs module can be directly embedded into a bottom-up-top-down structure and serves as a decoder, which cooperates with a transformer encoder and predicts accurate depth maps.
The experiments show that our method significantly outperforms previous methods and sets a new state-of-the-art performance on KITTI, NYUv2, and MatterPort3D datasets.

{\small
\bibliographystyle{ieee_fullname}
\bibliography{egbib}
}

\end{document}


\title{NeW CRFs: Neural Window Fully-connected CRFs for Monocular Depth Estimation}

\author{
Weihao Yuan\hspace{0.5cm}
Xiaodong Gu\hspace{0.5cm}
Zuozhuo Dai\hspace{0.5cm}
Siyu Zhu\hspace{0.5cm}
Ping Tan\\
Alibaba Group
\vspace{-20mm}
}
\maketitle

\appendix

    

\begin{figure*}[bp]
\centering
  \includegraphics[width=1.9\columnwidth, trim={0cm 0cm 0cm 0cm}, clip]{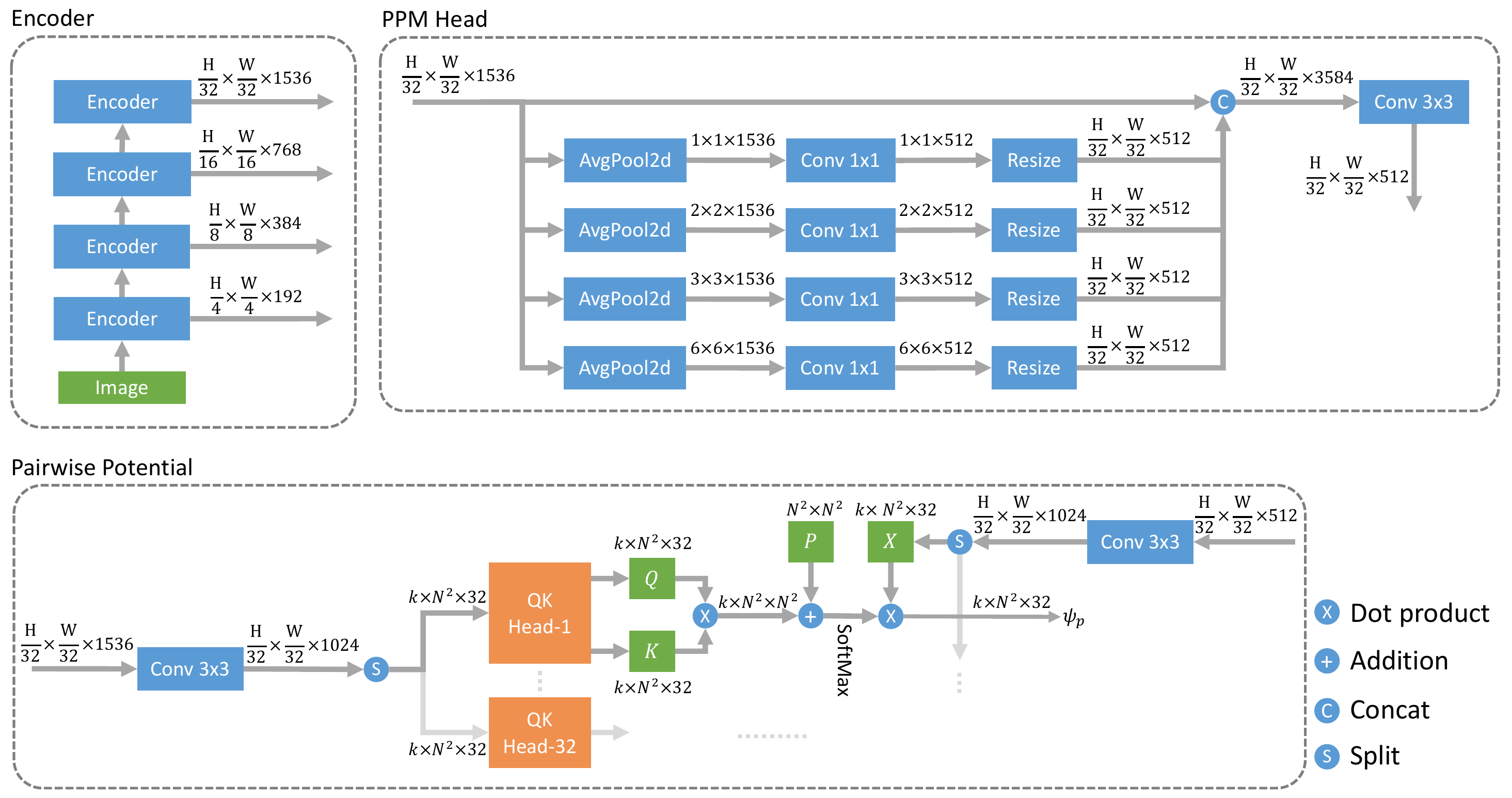}
\caption{Network details of encoder, PPM head, and pairwise potential. The pairwise potential in the top level is computed with $32$ heads, each of which is of $32$ channels. The graph of the image is split into $k$ windows of size $N \times N$.
}
\label{fig:details}
\end{figure*}

\section{Network Structure}

More details of our network are shown in Figure~\ref{fig:details}. The top-level output of the encoder is fed into the PPM head~\cite{zhao2017pyramid} for aggregating the local and global information.
Then in each level, a graph with patches as nodes is built, which is split into $k$ windows with size of $N \times N$.
From the top level to the bottom level, a combination of $32, 16, 8, 4$ is adopted for the head numbers.

\section{Efficiency Experiments}

To inspect the efficiency of the neural window FC-CRFs, we perform the experiments of different window size and report the efficiency and accuracy in Table~\ref{tab:effciency}.
The window FC-CRFs of window size 2 costs $121-95=26$ ms and $2.7-2.6=0.1$ G without considering the feature extraction. 
With the increase of window size $N$, the improving of the accuracy gradually tapers and is almost saturated at $0.055$ when $N=7$, but the increase of the cost grows up exponentially, costing $250$ ms and $5.3$ G when $N=48$.
Our window FC-CRFs of window size $N=2$ is equivalent to {traditional CRFs}, and that of window size $352\times1216$ is equivalent to the {FC-CRFs}. FC-CRFs consumes much more memory and computation than our window FC-CRF, even of large N (e.g. 48).
In traditional graph-based methods, the CRFs for depth estimation costs $59.1$ s in \cite{wang2015depth}, and the FC-CRFs for segmentation costs $0.5$ s in \cite{chen2017deeplab, krahenbuhl2011efficient}, while our window FC-CRFs of $7\times7$ costs only $29$ ms (without considering the feature extraction).

\setlength{\tabcolsep}{3pt}
\begin{table}[h]
\vspace{-3mm}
\small
\centering
\begin{tabular}{r c c c c c c c c c}
\toprule
Window size & Image size & Time & Memory & Abs Rel & RMSE  \\
\hline
Feature extract & $352\times 1216$ & $95$ms & 2.6G   \\
2 & $352\times 1216$ & $121$ms & $2.7$ G & $0.061$ & $2.524$   \\
4 & $352\times 1216$ & $122$ms & $2.7$ G & $0.058$ & $2.358$  \\
7/8 & $352\times 1216$ & $124$ms & $2.7$ G & $0.055$ & $2.234$    \\
16 & $352\times 1216$ & $136$ms & $2.9$ G  & $0.055$ & $2.188$    \\
24 & $352\times 1216$ & $149$ms & $3.5$ G & $0.055$ & $2.201$    \\
32 & $352\times 1216$ & $192$ms & $5.1$ G & $0.054$ & $2.232$    \\ 
48 & $352\times 1216$ & $345$ms & $7.3$ G & \multicolumn{2}{c}{OOM in training}   \\
\hline
~\cite{saxena2005learning} & $86\times 107$ & $7.5$s & -     \\
~\cite{wang2015depth} & $55\times 305$ & $59.1$s & -     \\
~\cite{chen2017deeplab, krahenbuhl2011efficient} & $480\times 640$ & $0.5$s & -     \\
\bottomrule
\end{tabular}
\caption{Efficiency experiments on the Eigen split of KITTI dataset. The first row is the time of other modules except the CRFs decoder, and then we add that of different window size. Models are retrained with a random crop of $352\times352$ of original images due to the high memory consumption of large window FC-CRFs. ``OOM" denotes out of memory. BTS, DPT, and our models are all tested on RTX 2080 Ti. The time of [4] is only the CRFs post-processing.}
\label{tab:effciency}
\end{table}
\setlength{\tabcolsep}{3pt}

\section{More Qualitative Results}

\begin{figure*}[t]
\centering
\begin{subfigure}{0.315\columnwidth}
  \centering
  \includegraphics[width=1\columnwidth, trim={0cm 0cm 0cm 0cm}, clip]{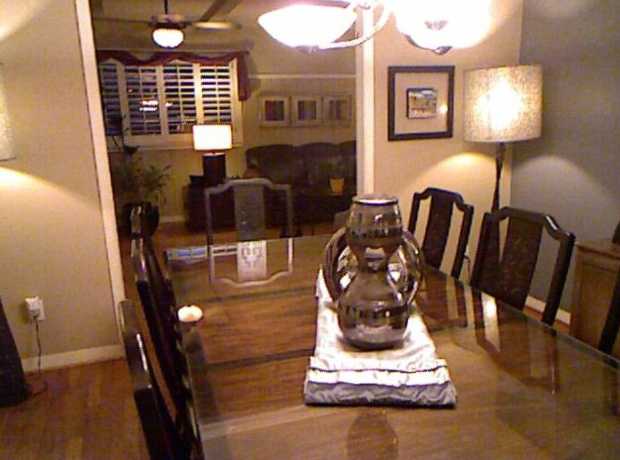}
\end{subfigure}
\begin{subfigure}{0.315\columnwidth}
  \centering
  \includegraphics[width=1\columnwidth, trim={0cm 0cm 0cm 0cm}, clip]{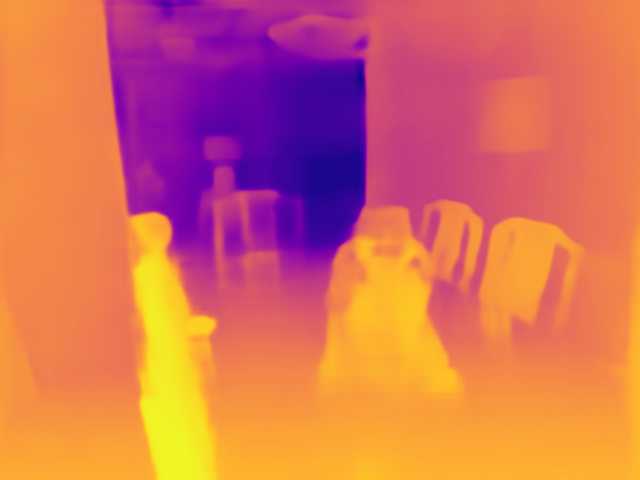}
\end{subfigure}
\begin{subfigure}{0.315\columnwidth}
  \centering
  \includegraphics[width=1\columnwidth, trim={0cm 0cm 0cm 0cm}, clip]{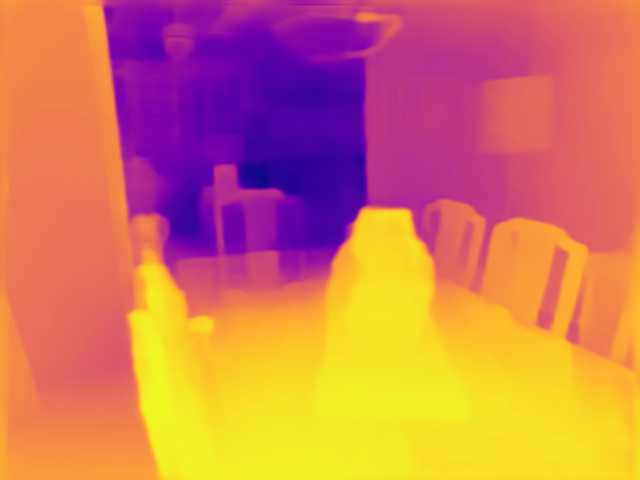}
\end{subfigure}
\begin{subfigure}{0.315\columnwidth}
  \centering
  \includegraphics[width=1\columnwidth, trim={0cm 0cm 0cm 0cm}, clip]{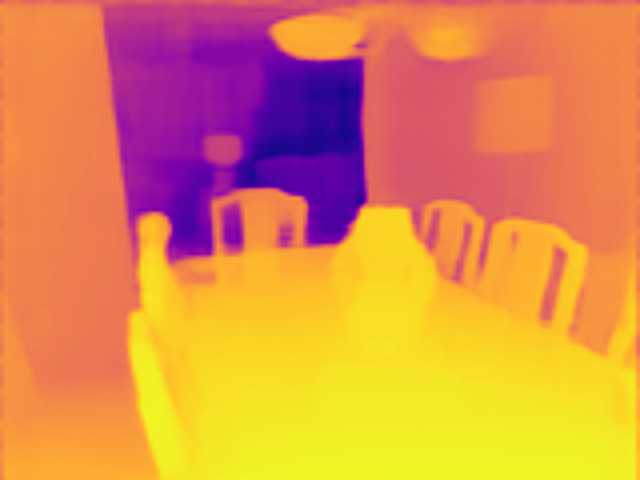}
\end{subfigure}
\begin{subfigure}{0.315\columnwidth}
  \centering
  \includegraphics[width=1\columnwidth, trim={0cm 0cm 0cm 0cm}, clip]{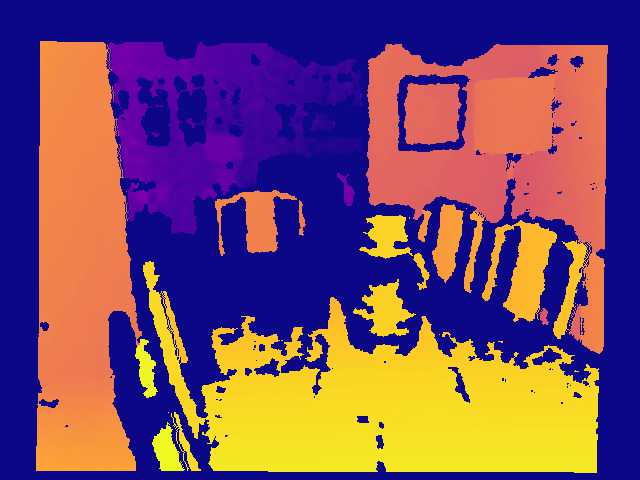}
\end{subfigure}

\begin{subfigure}{0.315\columnwidth}
  \centering
  \includegraphics[width=1\columnwidth, trim={0cm 0cm 0cm 0cm}, clip]{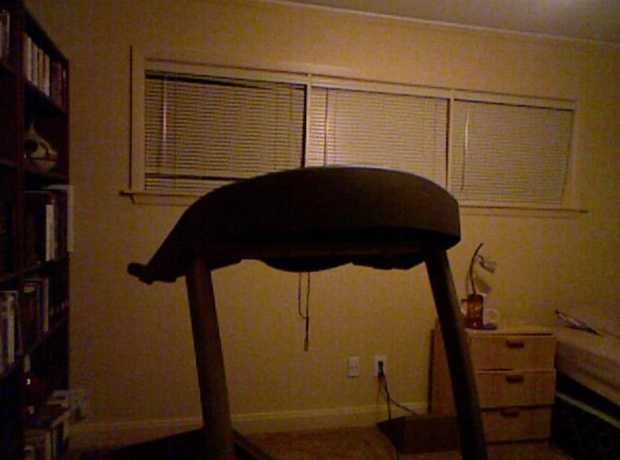}
\end{subfigure}
\begin{subfigure}{0.315\columnwidth}
  \centering
  \includegraphics[width=1\columnwidth, trim={0cm 0cm 0cm 0cm}, clip]{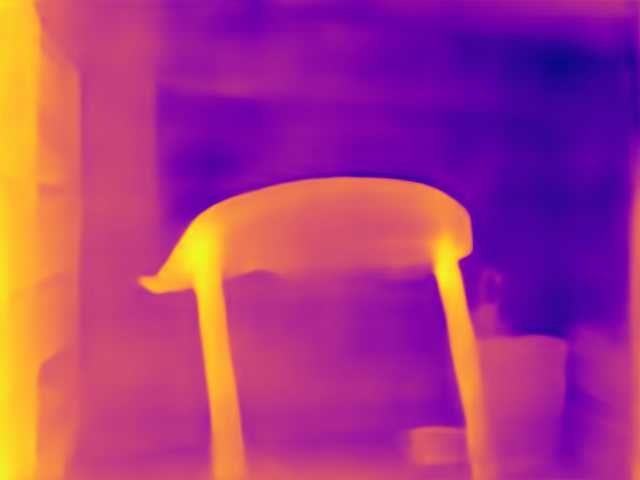}
\end{subfigure}
\begin{subfigure}{0.315\columnwidth}
  \centering
  \includegraphics[width=1\columnwidth, trim={0cm 0cm 0cm 0cm}, clip]{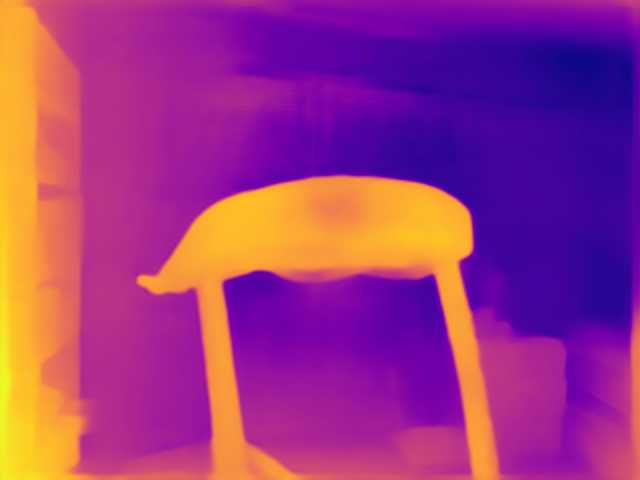}
\end{subfigure}
\begin{subfigure}{0.315\columnwidth}
  \centering
  \includegraphics[width=1\columnwidth, trim={0cm 0cm 0cm 0cm}, clip]{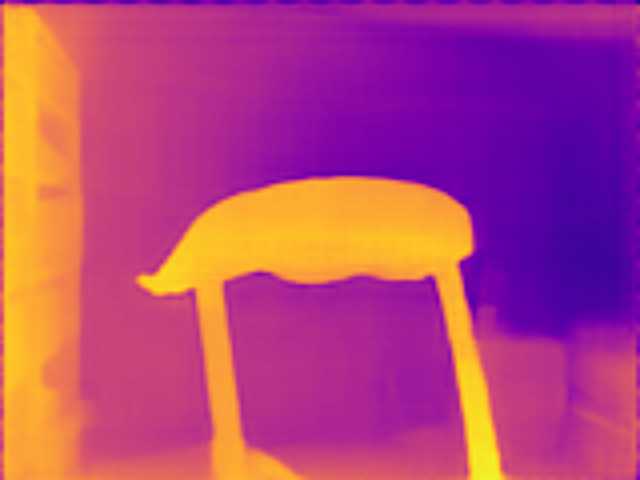}
\end{subfigure}
\begin{subfigure}{0.315\columnwidth}
  \centering
  \includegraphics[width=1\columnwidth, trim={0cm 0cm 0cm 0cm}, clip]{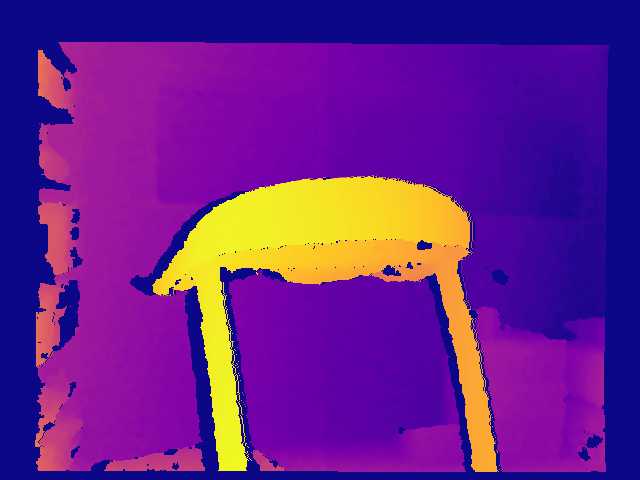}
\end{subfigure}

\begin{subfigure}{0.315\columnwidth}
  \centering
  \includegraphics[width=1\columnwidth, trim={0cm 0cm 0cm 0cm}, clip]{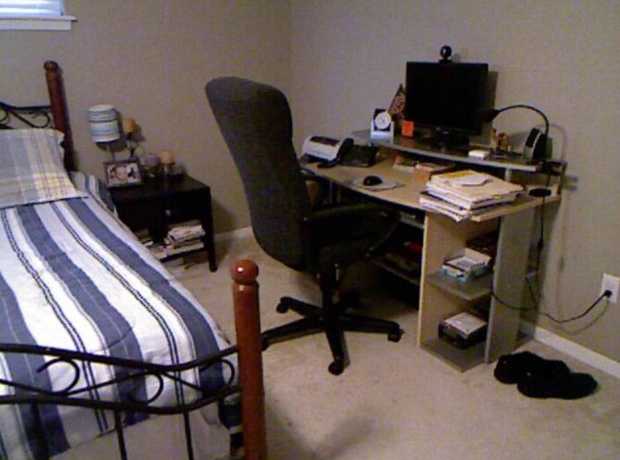}
\end{subfigure}
\begin{subfigure}{0.315\columnwidth}
  \centering
  \includegraphics[width=1\columnwidth, trim={0cm 0cm 0cm 0cm}, clip]{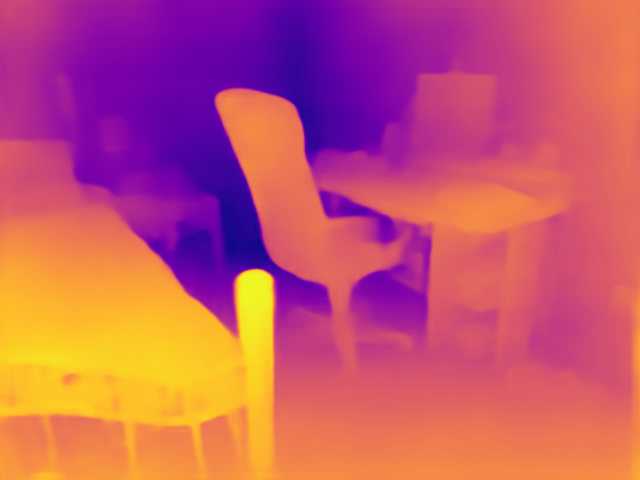}
\end{subfigure}
\begin{subfigure}{0.315\columnwidth}
  \centering
  \includegraphics[width=1\columnwidth, trim={0cm 0cm 0cm 0cm}, clip]{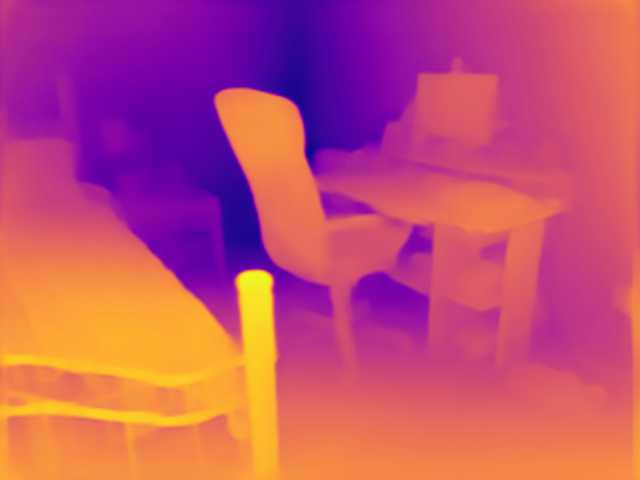}
\end{subfigure}
\begin{subfigure}{0.315\columnwidth}
  \centering
  \includegraphics[width=1\columnwidth, trim={0cm 0cm 0cm 0cm}, clip]{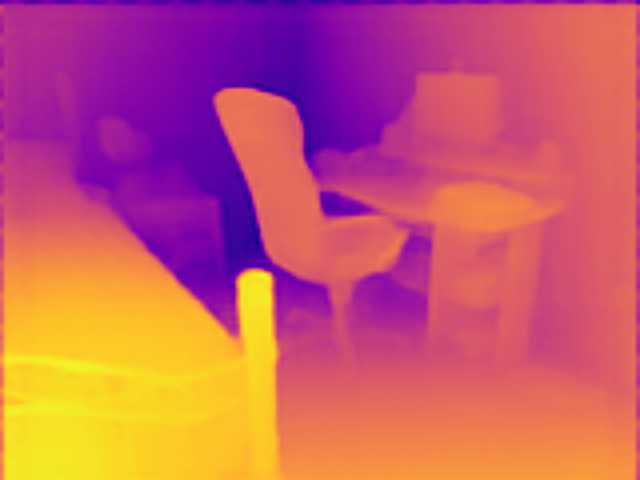}
\end{subfigure}
\begin{subfigure}{0.315\columnwidth}
  \centering
  \includegraphics[width=1\columnwidth, trim={0cm 0cm 0cm 0cm}, clip]{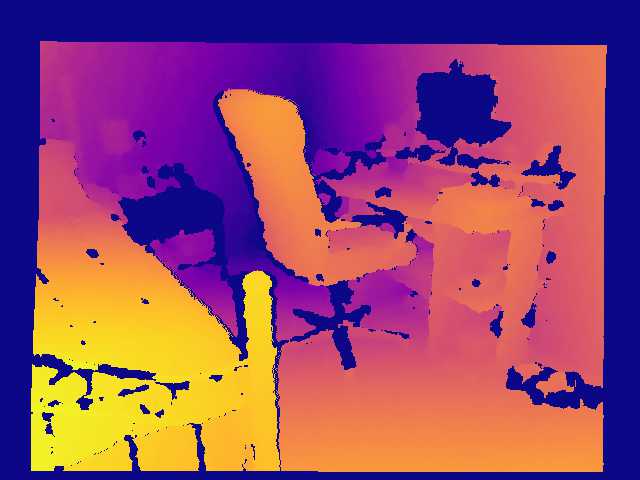}
\end{subfigure}

\begin{subfigure}{0.315\columnwidth}
  \centering
  \includegraphics[width=1\columnwidth, trim={0cm 0cm 0cm 0cm}, clip]{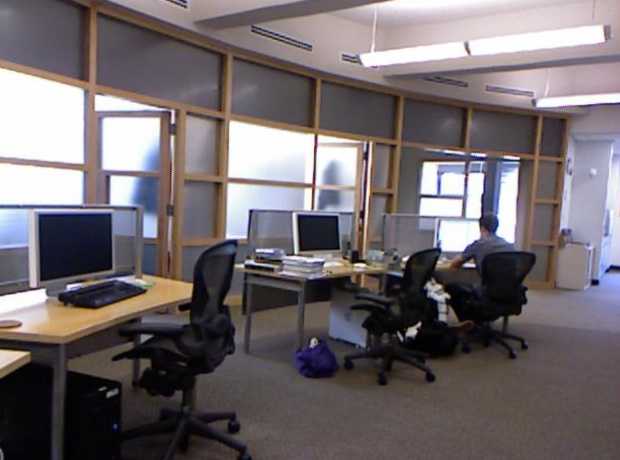}
\end{subfigure}
\begin{subfigure}{0.315\columnwidth}
  \centering
  \includegraphics[width=1\columnwidth, trim={0cm 0cm 0cm 0cm}, clip]{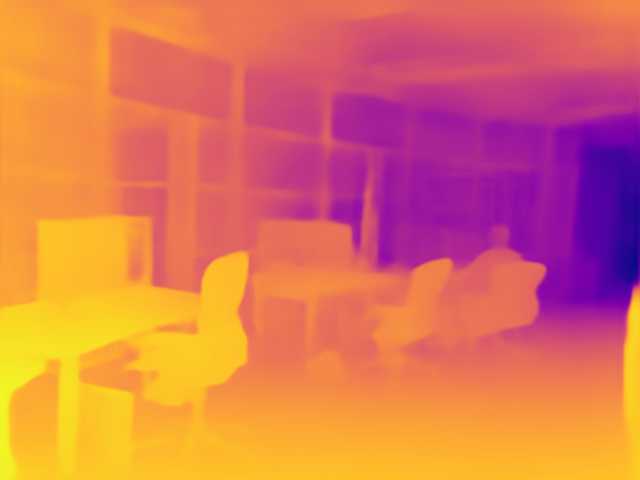}
\end{subfigure}
\begin{subfigure}{0.315\columnwidth}
  \centering
  \includegraphics[width=1\columnwidth, trim={0cm 0cm 0cm 0cm}, clip]{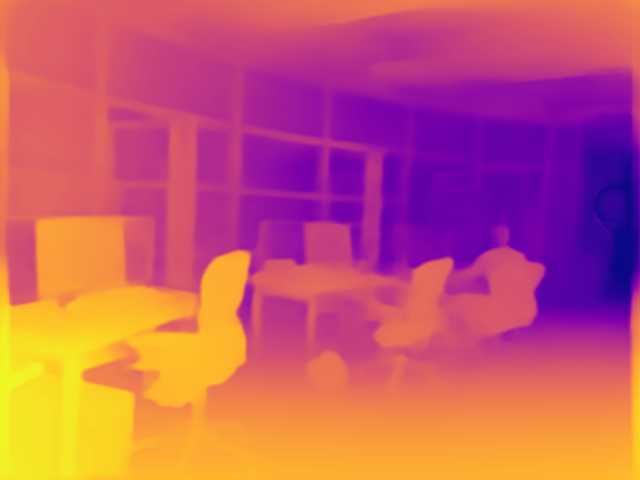}
\end{subfigure}
\begin{subfigure}{0.315\columnwidth}
  \centering
  \includegraphics[width=1\columnwidth, trim={0cm 0cm 0cm 0cm}, clip]{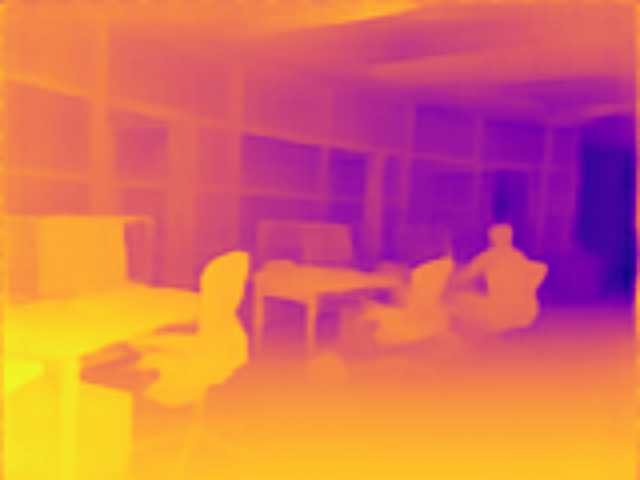}
\end{subfigure}
\begin{subfigure}{0.315\columnwidth}
  \centering
  \includegraphics[width=1\columnwidth, trim={0cm 0cm 0cm 0cm}, clip]{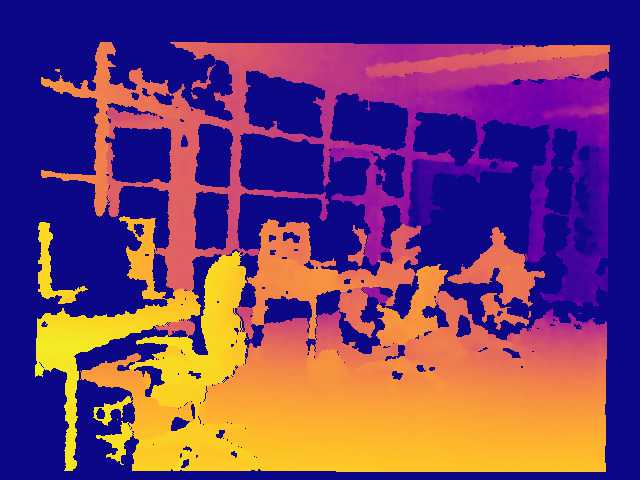}
\end{subfigure}

\begin{subfigure}{0.315\columnwidth}
  \centering
  \includegraphics[width=1\columnwidth, trim={0cm 0cm 0cm 0cm}, clip]{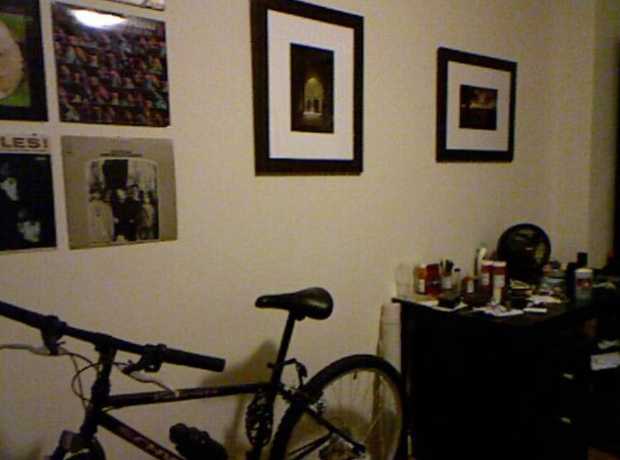}
\end{subfigure}
\begin{subfigure}{0.315\columnwidth}
  \centering
  \includegraphics[width=1\columnwidth, trim={0cm 0cm 0cm 0cm}, clip]{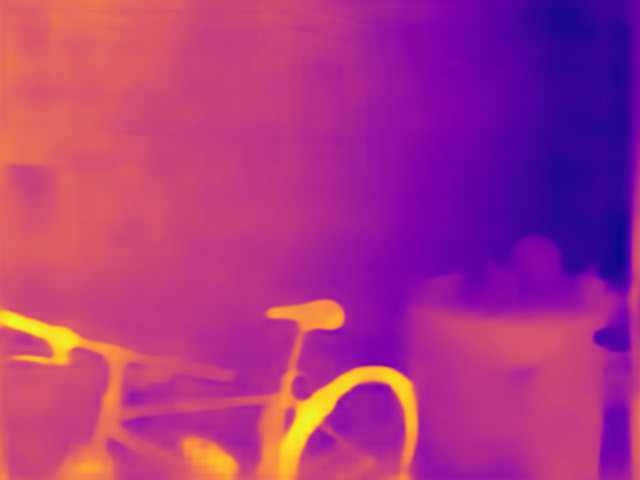}
\end{subfigure}
\begin{subfigure}{0.315\columnwidth}
  \centering
  \includegraphics[width=1\columnwidth, trim={0cm 0cm 0cm 0cm}, clip]{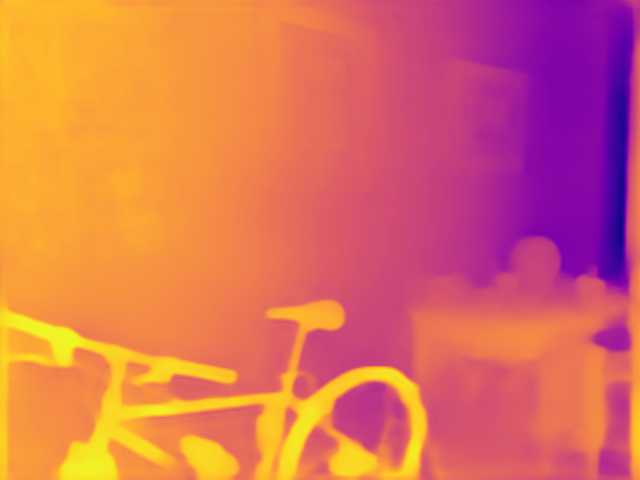}
\end{subfigure}
\begin{subfigure}{0.315\columnwidth}
  \centering
  \includegraphics[width=1\columnwidth, trim={0cm 0cm 0cm 0cm}, clip]{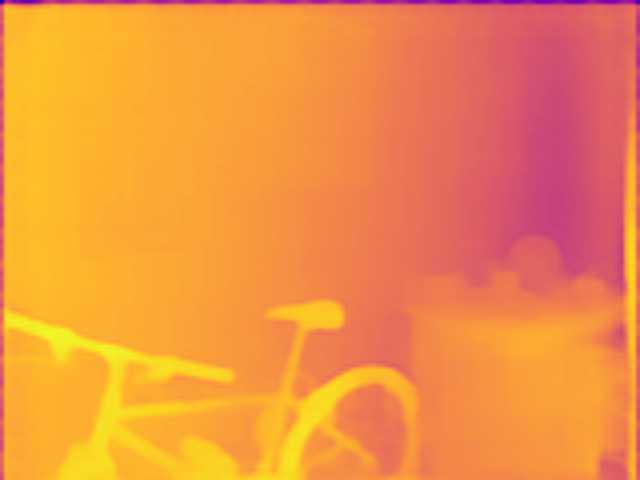}
\end{subfigure}
\begin{subfigure}{0.315\columnwidth}
  \centering
  \includegraphics[width=1\columnwidth, trim={0cm 0cm 0cm 0cm}, clip]{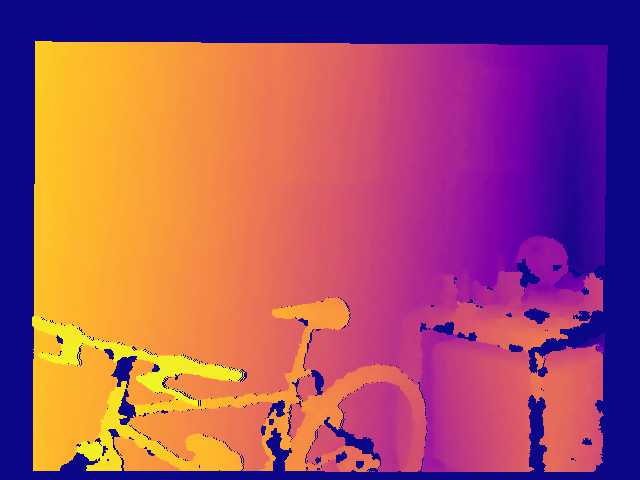}
\end{subfigure}

\begin{subfigure}{0.315\columnwidth}
  \centering
  \includegraphics[width=1\columnwidth, trim={0cm 0cm 0cm 0cm}, clip]{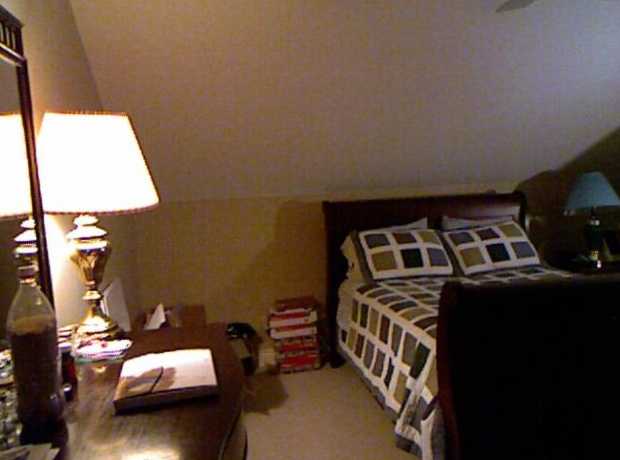}
\end{subfigure}
\begin{subfigure}{0.315\columnwidth}
  \centering
  \includegraphics[width=1\columnwidth, trim={0cm 0cm 0cm 0cm}, clip]{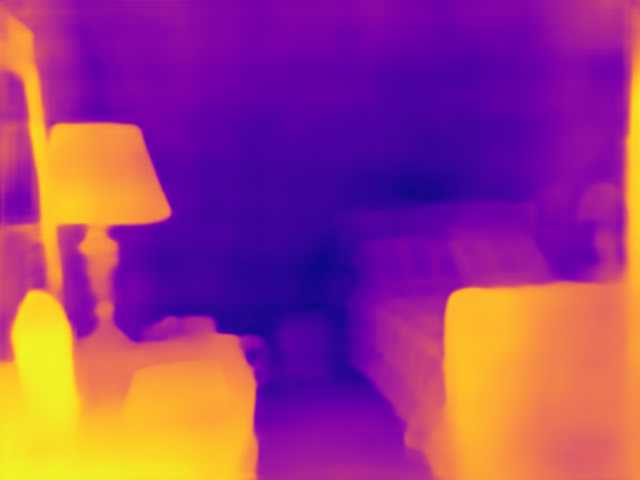}
\end{subfigure}
\begin{subfigure}{0.315\columnwidth}
  \centering
  \includegraphics[width=1\columnwidth, trim={0cm 0cm 0cm 0cm}, clip]{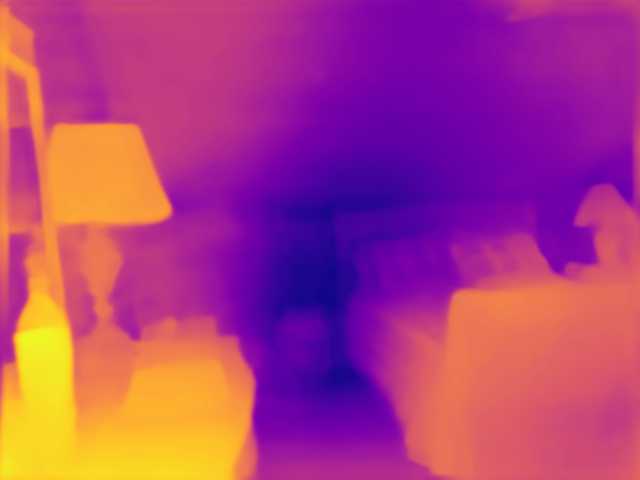}
\end{subfigure}
\begin{subfigure}{0.315\columnwidth}
  \centering
  \includegraphics[width=1\columnwidth, trim={0cm 0cm 0cm 0cm}, clip]{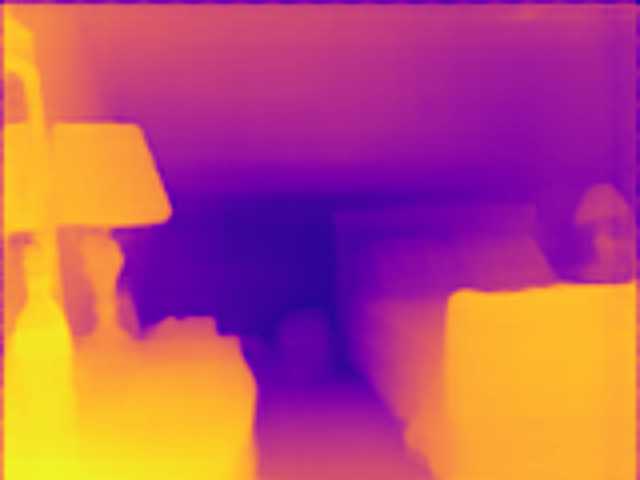}
\end{subfigure}
\begin{subfigure}{0.315\columnwidth}
  \centering
  \includegraphics[width=1\columnwidth, trim={0cm 0cm 0cm 0cm}, clip]{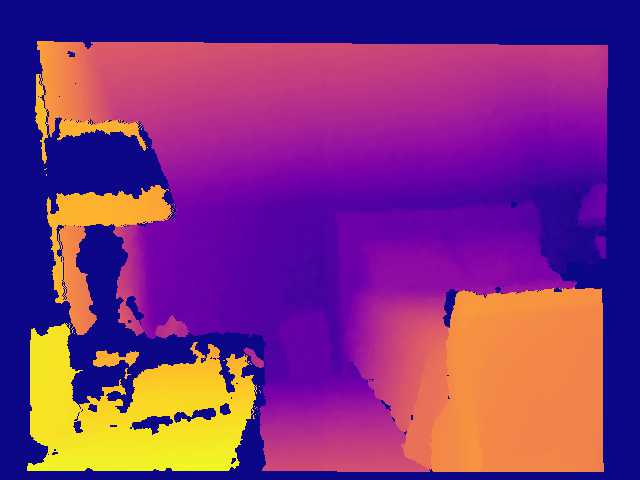}
\end{subfigure}

\begin{subfigure}{0.315\columnwidth}
  \centering
  \includegraphics[width=1\columnwidth, trim={0cm 0cm 0cm 0cm}, clip]{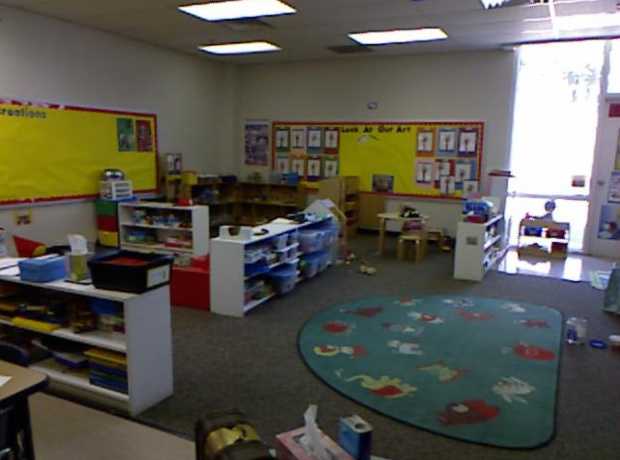}
\end{subfigure}
\begin{subfigure}{0.315\columnwidth}
  \centering
  \includegraphics[width=1\columnwidth, trim={0cm 0cm 0cm 0cm}, clip]{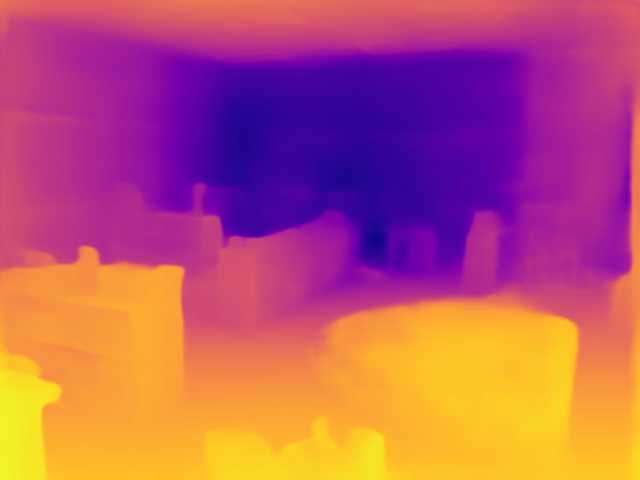}
\end{subfigure}
\begin{subfigure}{0.315\columnwidth}
  \centering
  \includegraphics[width=1\columnwidth, trim={0cm 0cm 0cm 0cm}, clip]{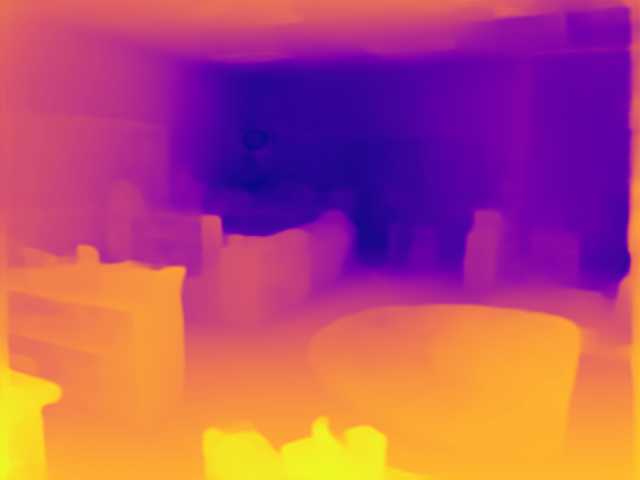}
\end{subfigure}
\begin{subfigure}{0.315\columnwidth}
  \centering
  \includegraphics[width=1\columnwidth, trim={0cm 0cm 0cm 0cm}, clip]{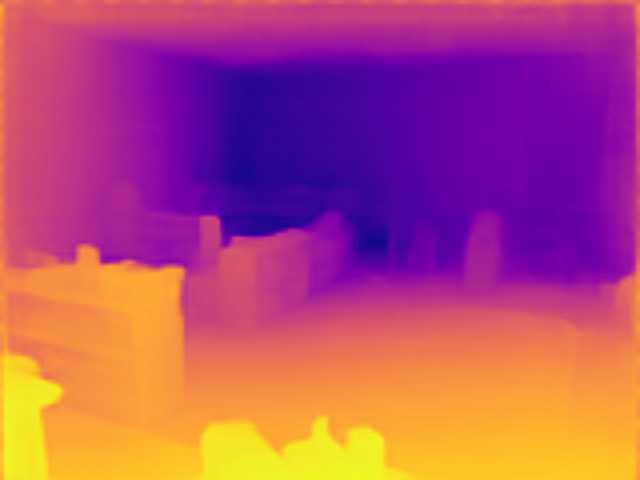}
\end{subfigure}
\begin{subfigure}{0.315\columnwidth}
  \centering
  \includegraphics[width=1\columnwidth, trim={0cm 0cm 0cm 0cm}, clip]{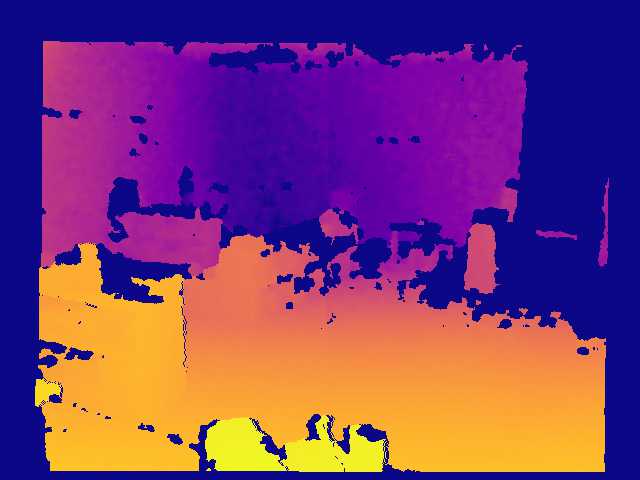}
\end{subfigure}

\begin{subfigure}{0.315\columnwidth}
  \centering
  \includegraphics[width=1\columnwidth, trim={0cm 0cm 0cm 0cm}, clip]{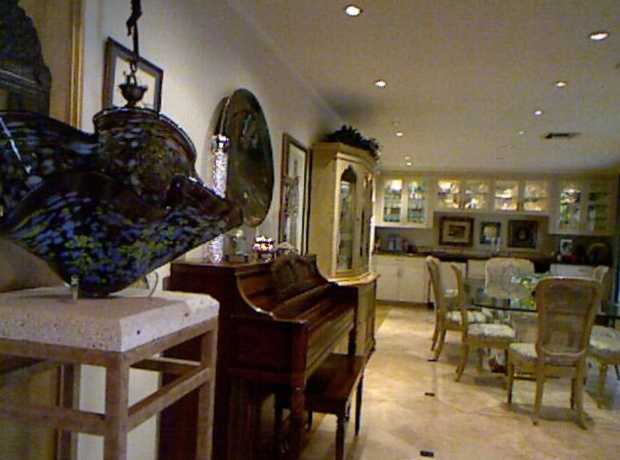}
\end{subfigure}
\begin{subfigure}{0.315\columnwidth}
  \centering
  \includegraphics[width=1\columnwidth, trim={0cm 0cm 0cm 0cm}, clip]{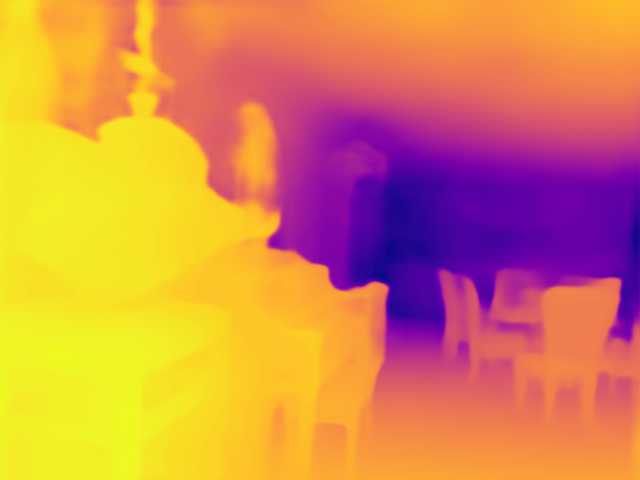}
\end{subfigure}
\begin{subfigure}{0.315\columnwidth}
  \centering
  \includegraphics[width=1\columnwidth, trim={0cm 0cm 0cm 0cm}, clip]{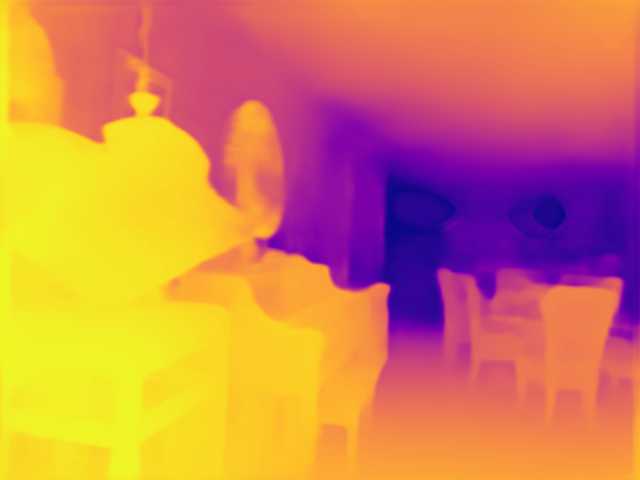}
\end{subfigure}
\begin{subfigure}{0.315\columnwidth}
  \centering
  \includegraphics[width=1\columnwidth, trim={0cm 0cm 0cm 0cm}, clip]{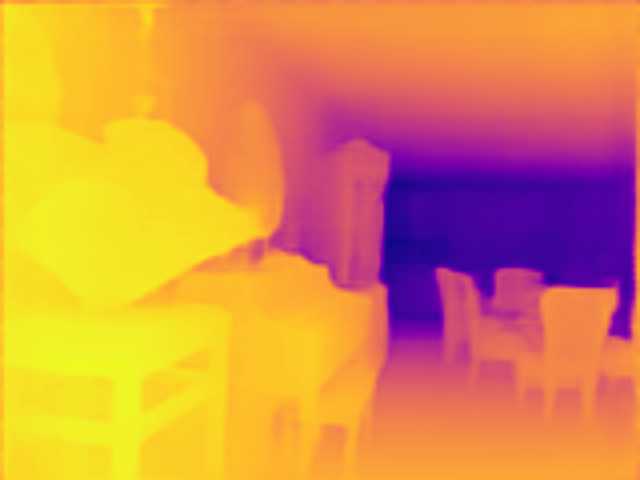}
\end{subfigure}
\begin{subfigure}{0.315\columnwidth}
  \centering
  \includegraphics[width=1\columnwidth, trim={0cm 0cm 0cm 0cm}, clip]{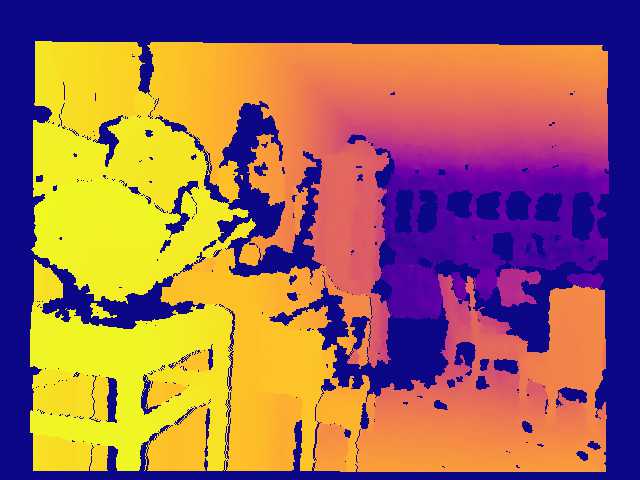}
\end{subfigure}

\begin{subfigure}{0.315\columnwidth}
  \centering
  \includegraphics[width=1\columnwidth, trim={0cm 0cm 0cm 0cm}, clip]{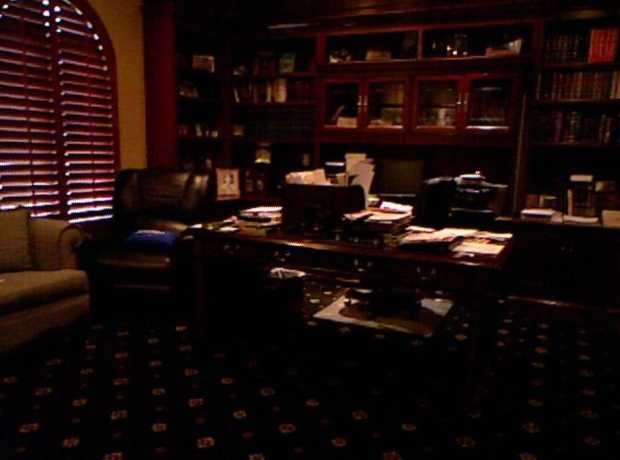}
\end{subfigure}
\begin{subfigure}{0.315\columnwidth}
  \centering
  \includegraphics[width=1\columnwidth, trim={0cm 0cm 0cm 0cm}, clip]{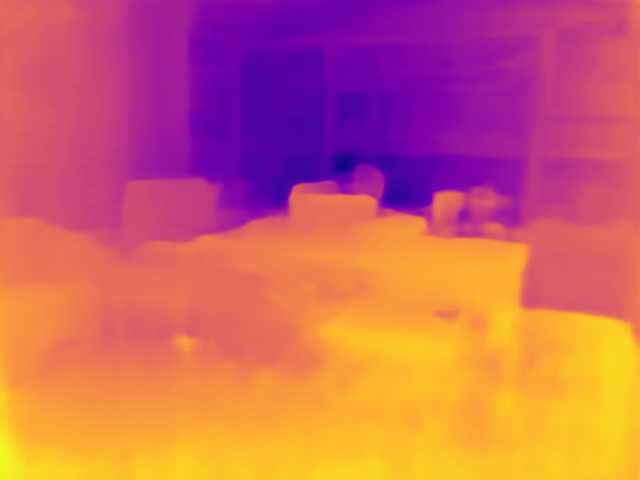}
\end{subfigure}
\begin{subfigure}{0.315\columnwidth}
  \centering
  \includegraphics[width=1\columnwidth, trim={0cm 0cm 0cm 0cm}, clip]{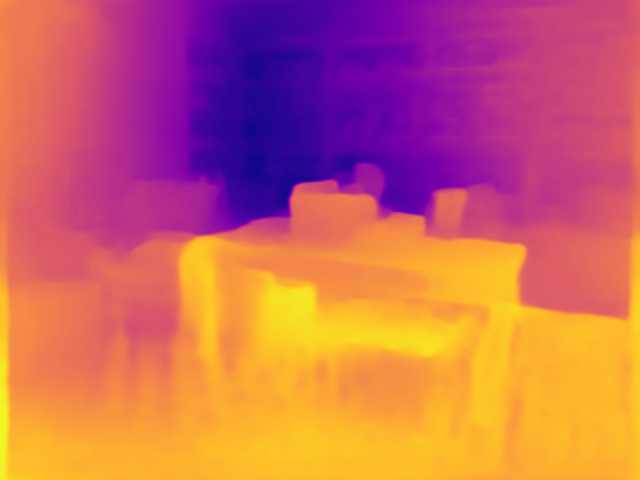}
\end{subfigure}
\begin{subfigure}{0.315\columnwidth}
  \centering
  \includegraphics[width=1\columnwidth, trim={0cm 0cm 0cm 0cm}, clip]{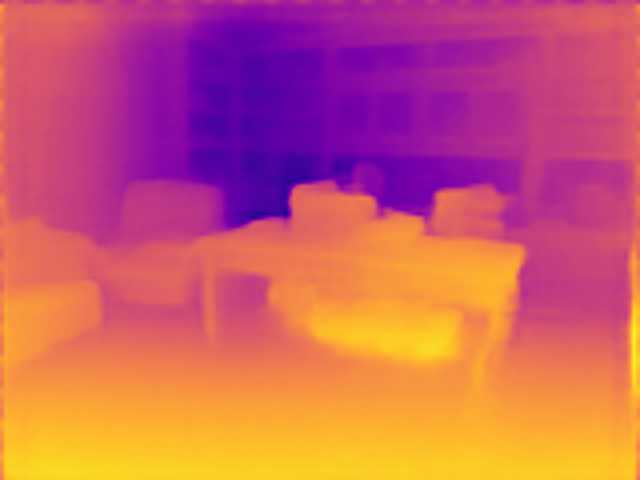}
\end{subfigure}
\begin{subfigure}{0.315\columnwidth}
  \centering
  \includegraphics[width=1\columnwidth, trim={0cm 0cm 0cm 0cm}, clip]{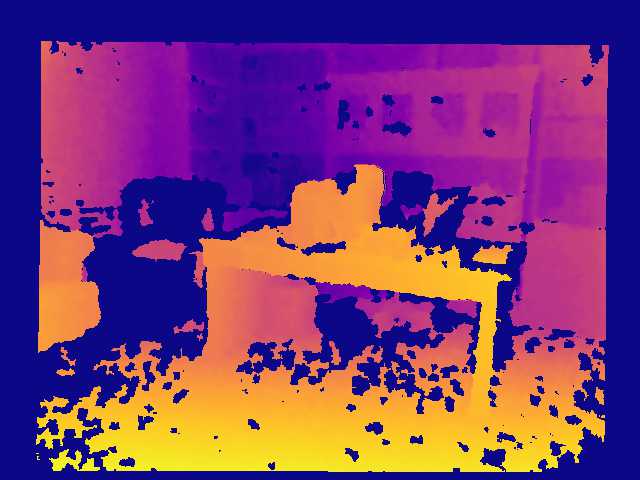}
\end{subfigure}

\begin{subfigure}{0.315\columnwidth}
  \centering
  \includegraphics[width=1\columnwidth, trim={0cm 0cm 0cm 0cm}, clip]{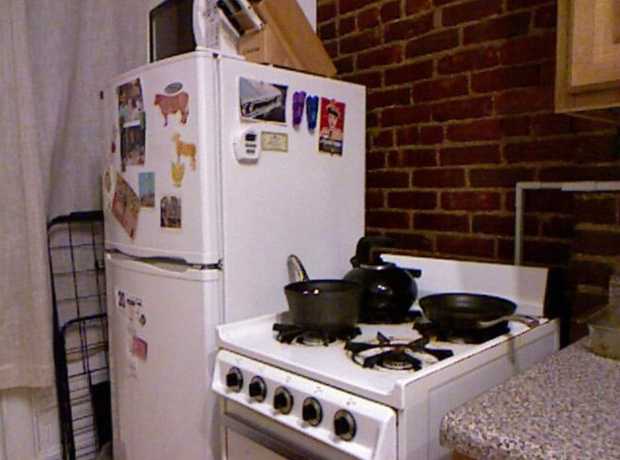}
\end{subfigure}
\begin{subfigure}{0.315\columnwidth}
  \centering
  \includegraphics[width=1\columnwidth, trim={0cm 0cm 0cm 0cm}, clip]{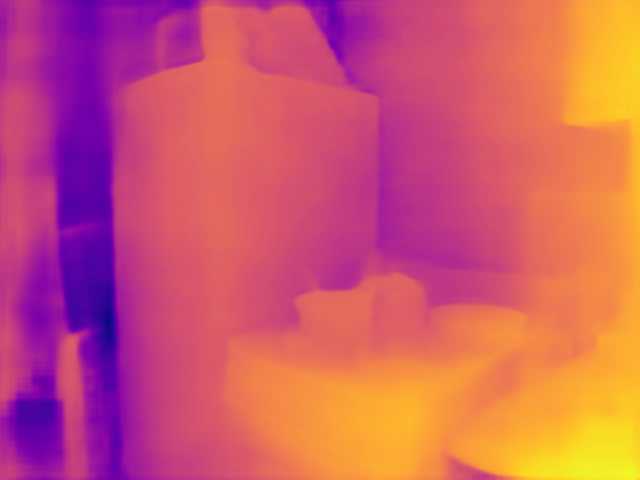}
\end{subfigure}
\begin{subfigure}{0.315\columnwidth}
  \centering
  \includegraphics[width=1\columnwidth, trim={0cm 0cm 0cm 0cm}, clip]{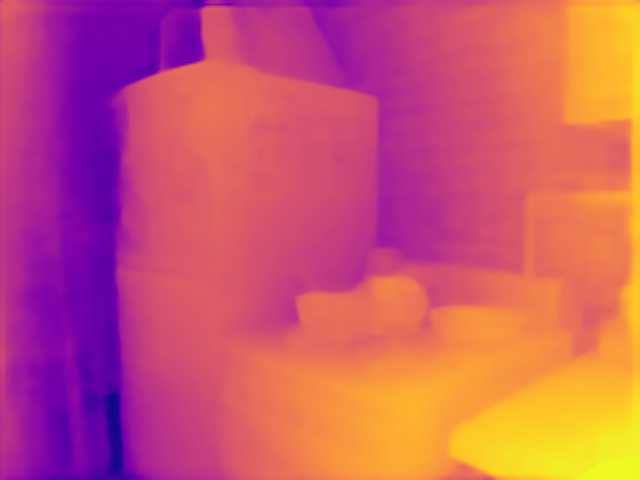}
\end{subfigure}
\begin{subfigure}{0.315\columnwidth}
  \centering
  \includegraphics[width=1\columnwidth, trim={0cm 0cm 0cm 0cm}, clip]{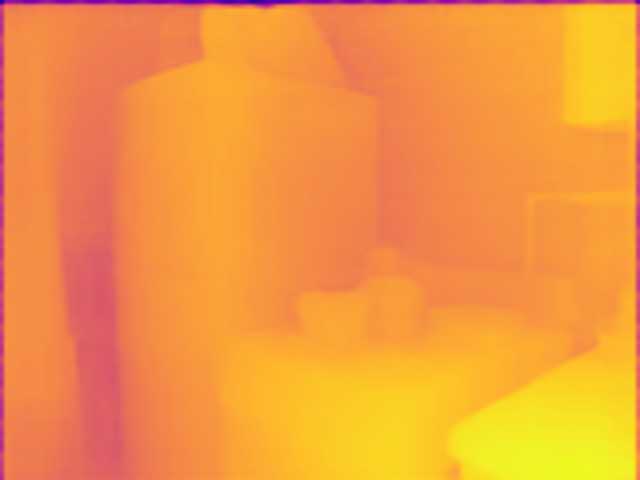}
\end{subfigure}
\begin{subfigure}{0.315\columnwidth}
  \centering
  \includegraphics[width=1\columnwidth, trim={0cm 0cm 0cm 0cm}, clip]{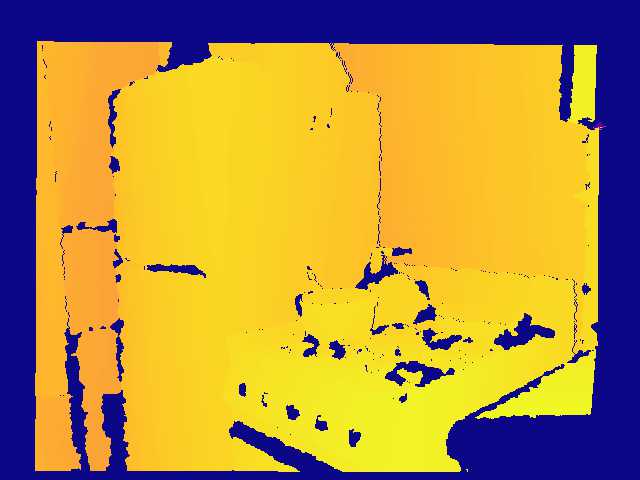}
\end{subfigure}

\begin{subfigure}{0.315\columnwidth}
  \centering
  \includegraphics[width=1\columnwidth, trim={0cm 0cm 0cm 0cm}, clip]{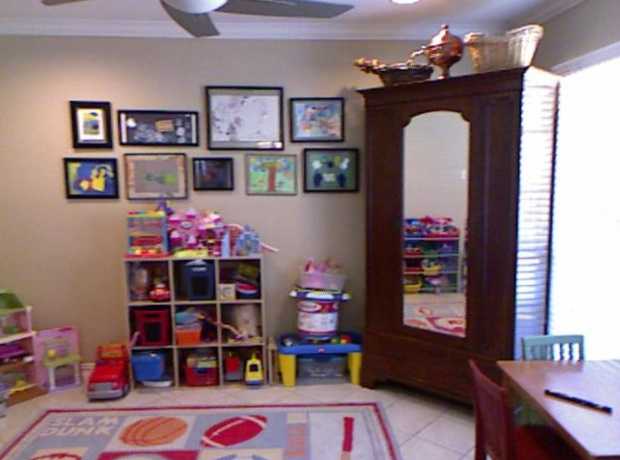}
  \caption*{Input image}
\end{subfigure}
\begin{subfigure}{0.315\columnwidth}
  \centering
  \includegraphics[width=1\columnwidth, trim={0cm 0cm 0cm 0cm}, clip]{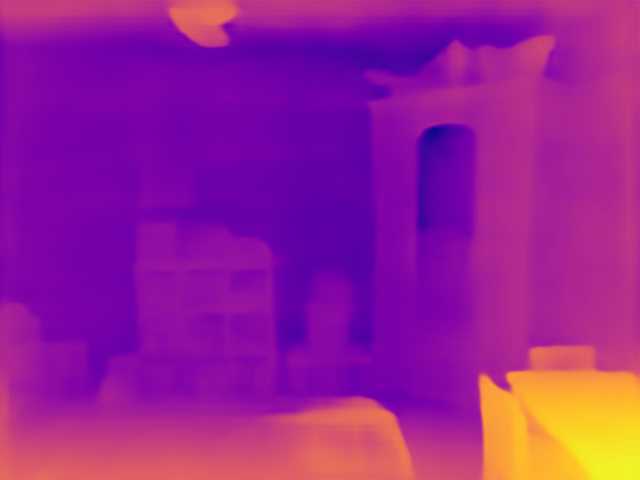}
  \caption*{BTS~\cite{lee2019big}}
\end{subfigure}
\begin{subfigure}{0.315\columnwidth}
  \centering
  \includegraphics[width=1\columnwidth, trim={0cm 0cm 0cm 0cm}, clip]{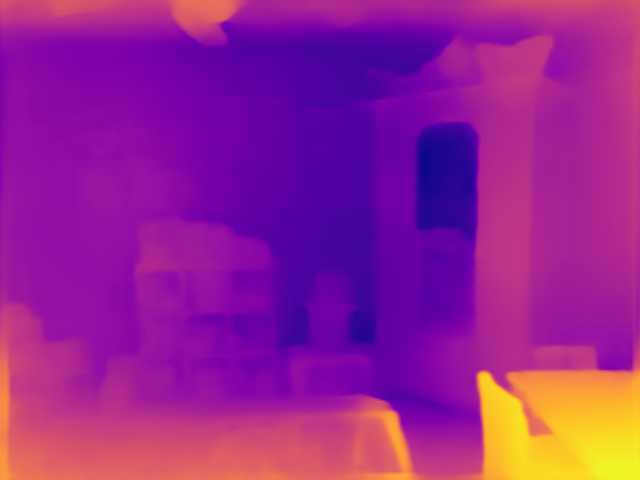}
  \caption*{Adabins~\cite{bhat2021adabins}}
\end{subfigure}
\begin{subfigure}{0.315\columnwidth}
  \centering
  \includegraphics[width=1\columnwidth, trim={0cm 0cm 0cm 0cm}, clip]{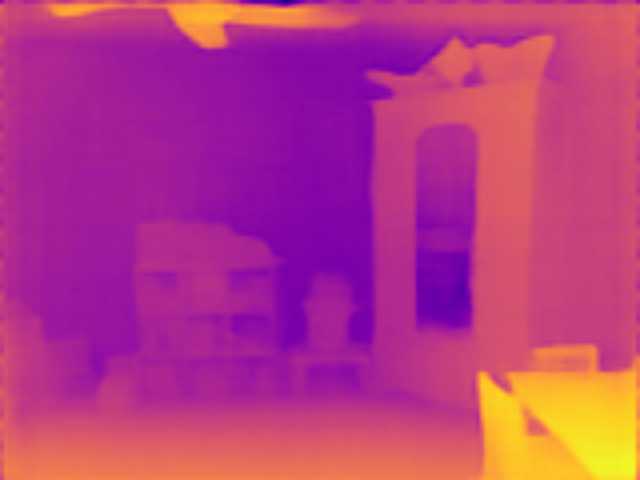}
  \caption*{Ours}
\end{subfigure}
\begin{subfigure}{0.315\columnwidth}
  \centering
  \includegraphics[width=1\columnwidth, trim={0cm 0cm 0cm 0cm}, clip]{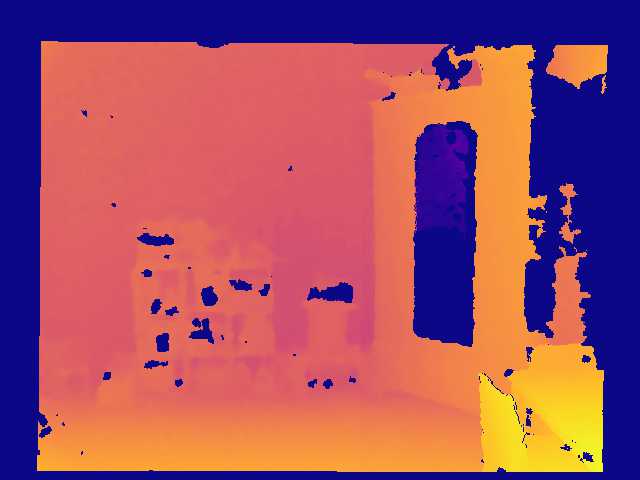}
  \caption*{Ground truth}
\end{subfigure}

\caption{Qualitative results on the test set of NYUv2 dataset.}
\label{fig:res-nyuapp}
\end{figure*}

To make more comparisons to previous state-of-the-art methods, we display more qualitative results of BTS~\cite{lee2019big}, Adabins~\cite{bhat2021adabins}, and our method on the test set of NYUv2~\cite{silberman2012indoor} dataset, as is shown in Figure~\ref{fig:res-nyuapp}. 
From the results, our method estimates better depth and recover more details, especially in difficult regions, such as repeated texture, messy environment, and bad light.

\section{Point Cloud Visualization}

\begin{figure*}[t]
\centering
\begin{subfigure}{0.5\columnwidth}
  \centering
  \includegraphics[width=1\columnwidth, trim={0cm 0cm 0cm 0cm}, clip]{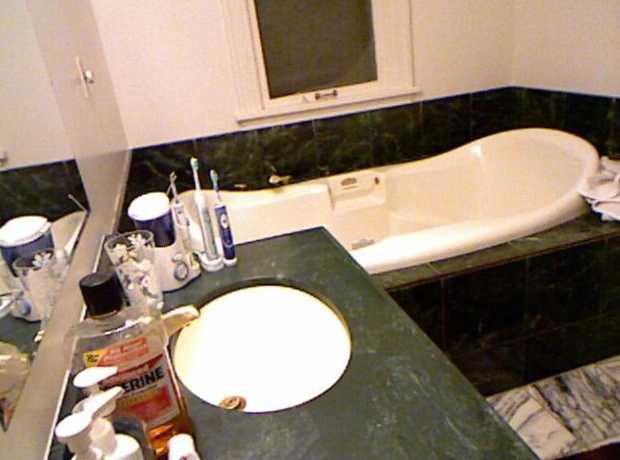}
\end{subfigure}
\begin{subfigure}{0.5\columnwidth}
  \centering
  \includegraphics[width=1\columnwidth, trim={0cm 0cm 0cm 0cm}, clip]{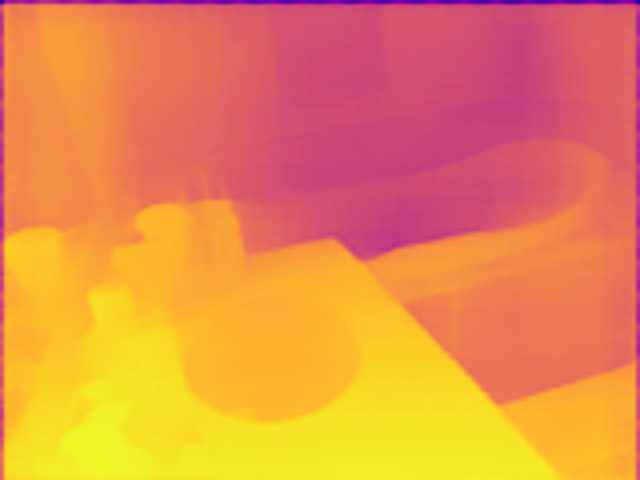}
\end{subfigure}
\begin{subfigure}{1.0\columnwidth}
  \centering
  \includegraphics[width=1\columnwidth, trim={0cm 0cm 0cm 0cm}, clip]{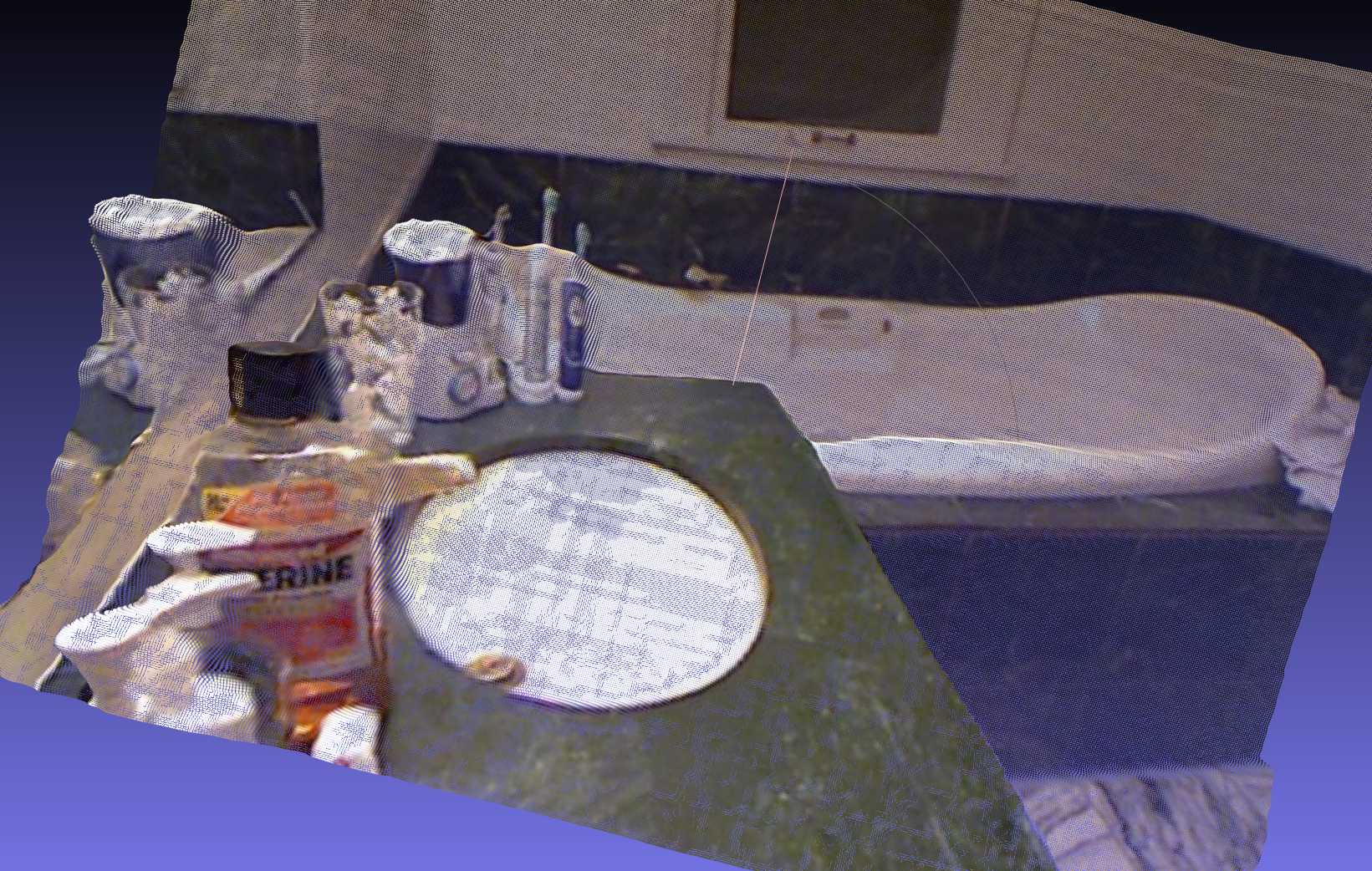}
\end{subfigure}

\begin{subfigure}{0.5\columnwidth}
  \centering
  \includegraphics[width=1\columnwidth, trim={0cm 0cm 0cm 0cm}, clip]{figures/nyu/bedroom_01077}
\end{subfigure}
\begin{subfigure}{0.5\columnwidth}
  \centering
  \includegraphics[width=1\columnwidth, trim={0cm 0cm 0cm 0cm}, clip]{figures/nyu/bedroom_01077_depth}
\end{subfigure}
\begin{subfigure}{1.0\columnwidth}
  \centering
  \includegraphics[width=1\columnwidth, trim={0cm 3cm 0cm 3cm}, clip]{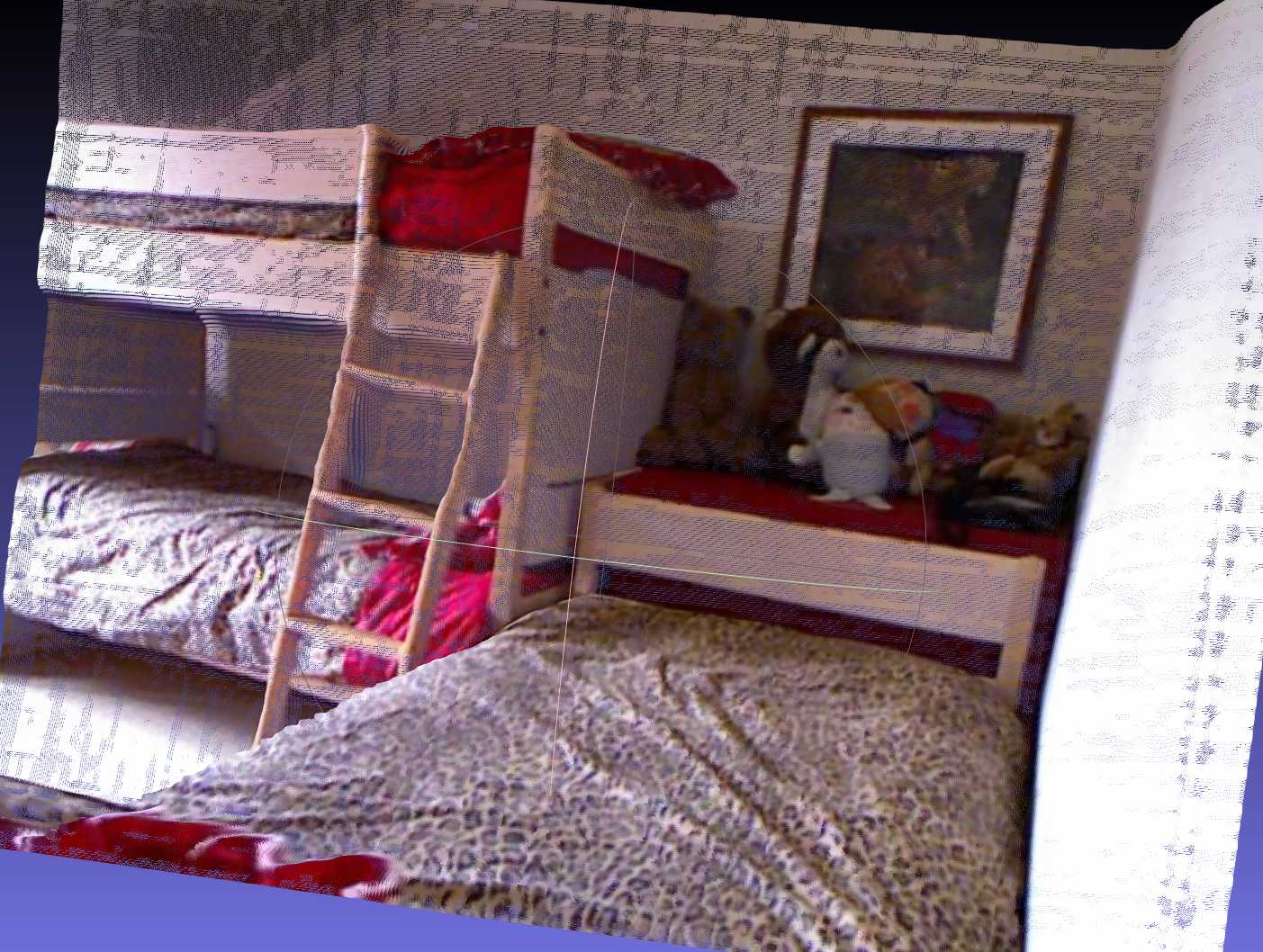}
\end{subfigure}

\begin{subfigure}{0.5\columnwidth}
  \centering
  \includegraphics[width=1\columnwidth, trim={0cm 0cm 0cm 0cm}, clip]{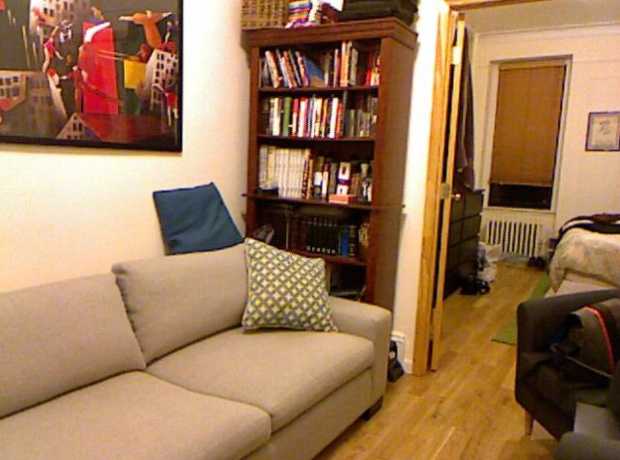}
\end{subfigure}
\begin{subfigure}{0.5\columnwidth}
  \centering
  \includegraphics[width=1\columnwidth, trim={0cm 0cm 0cm 0cm}, clip]{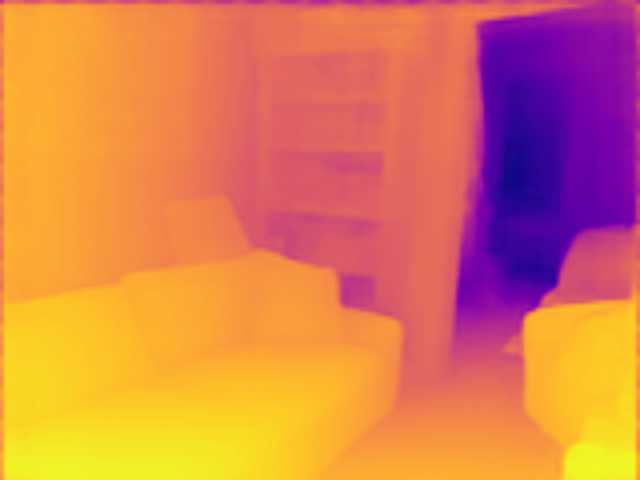}
\end{subfigure}
\begin{subfigure}{1.0\columnwidth}
  \centering
  \includegraphics[width=1\columnwidth, trim={0cm 3cm 0cm 3cm}, clip]{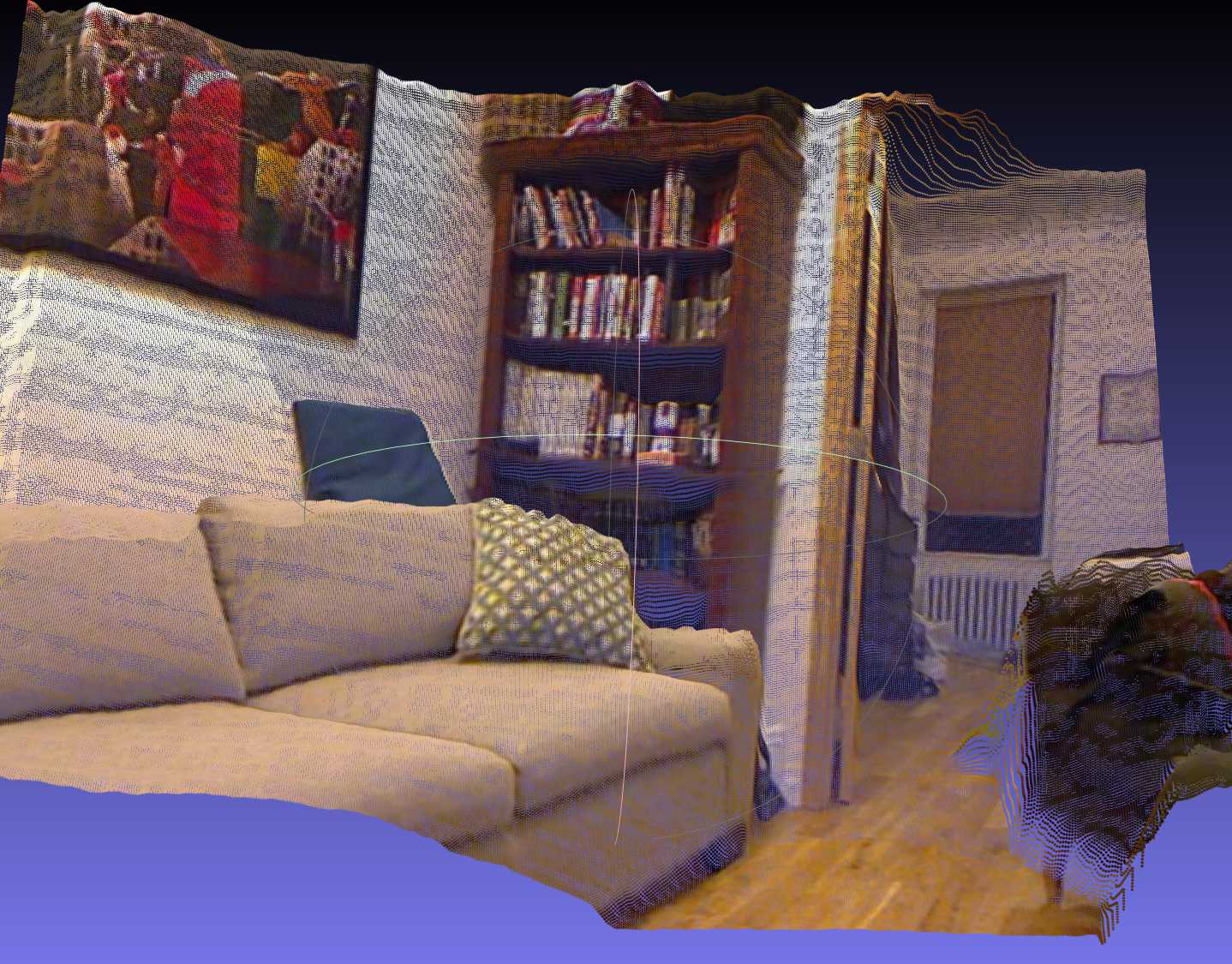}
\end{subfigure}

\begin{subfigure}{0.5\columnwidth}
  \centering
  \includegraphics[width=1\columnwidth, trim={0cm 0cm 0cm 0cm}, clip]{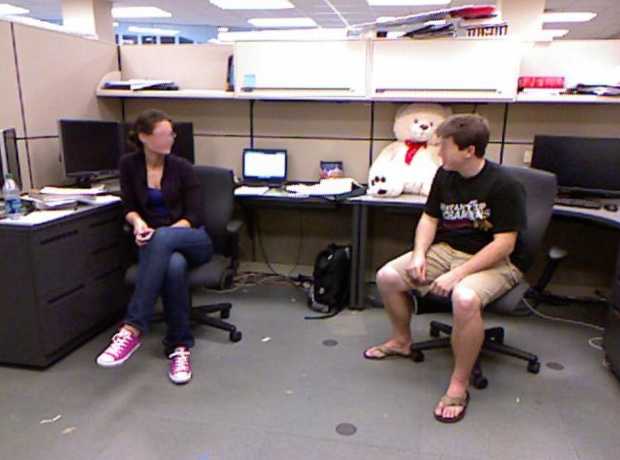}
  \caption*{Input image}
\end{subfigure}
\begin{subfigure}{0.5\columnwidth}
  \centering
  \includegraphics[width=1\columnwidth, trim={0cm 0cm 0cm 0cm}, clip]{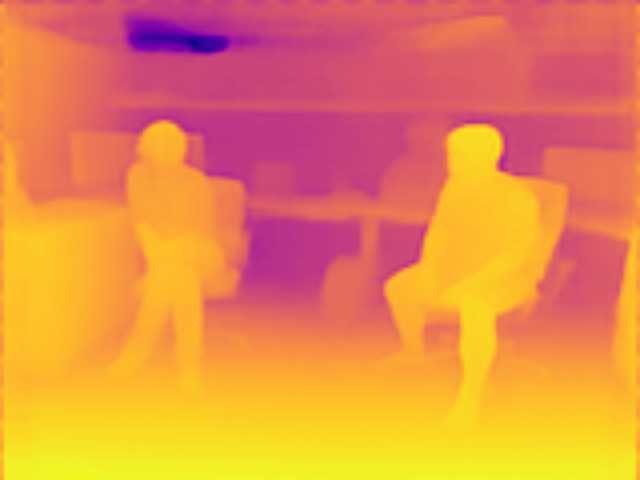}
  \caption*{Estimated depth}
\end{subfigure}
\begin{subfigure}{1.0\columnwidth}
  \centering
  \includegraphics[width=1\columnwidth, trim={0cm 0cm 0cm 0cm}, clip]{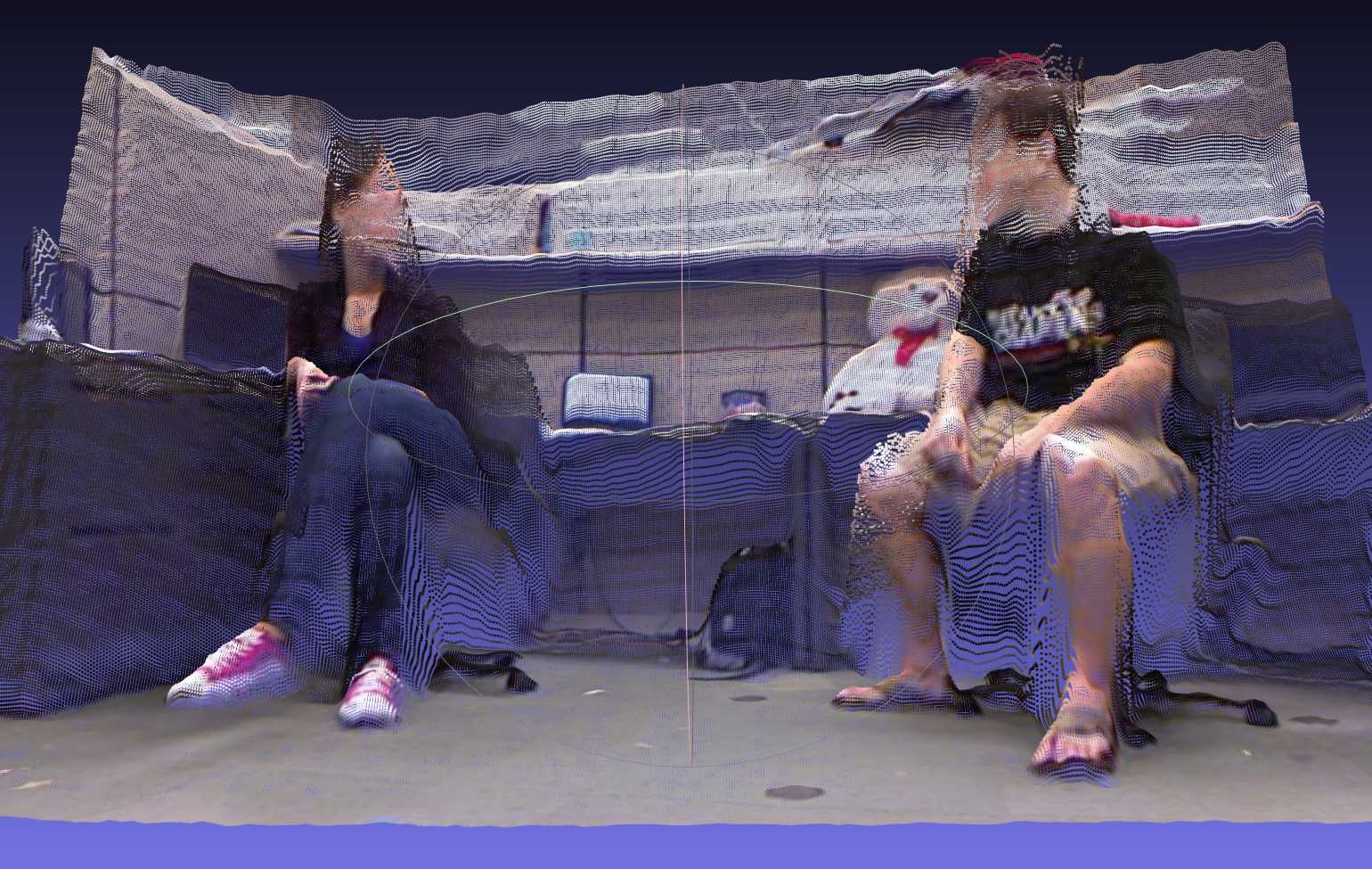}
  \caption*{Point cloud}
\end{subfigure}


\caption{Point cloud visualization on the test set of NYUv2 dataset.}
\label{fig:res-nyupcd}
\end{figure*}

To better see the 3D shape of the estimated depth map, we project the 2D pixels of the color image back to the 3D world utilizing the estimated depth map. The generated point clouds on the test set of NYUv2 dataset are displayed in Figure~\ref{fig:res-nyupcd}, where the structures of the 3D world are recovered reasonably.

\begin{figure*}[t]
\centering
\begin{subfigure}{0.8\columnwidth}
  \centering
  \scalebox{-1}[1]{\includegraphics[width=1\columnwidth, trim={0cm 0cm 0cm 0cm}, clip]{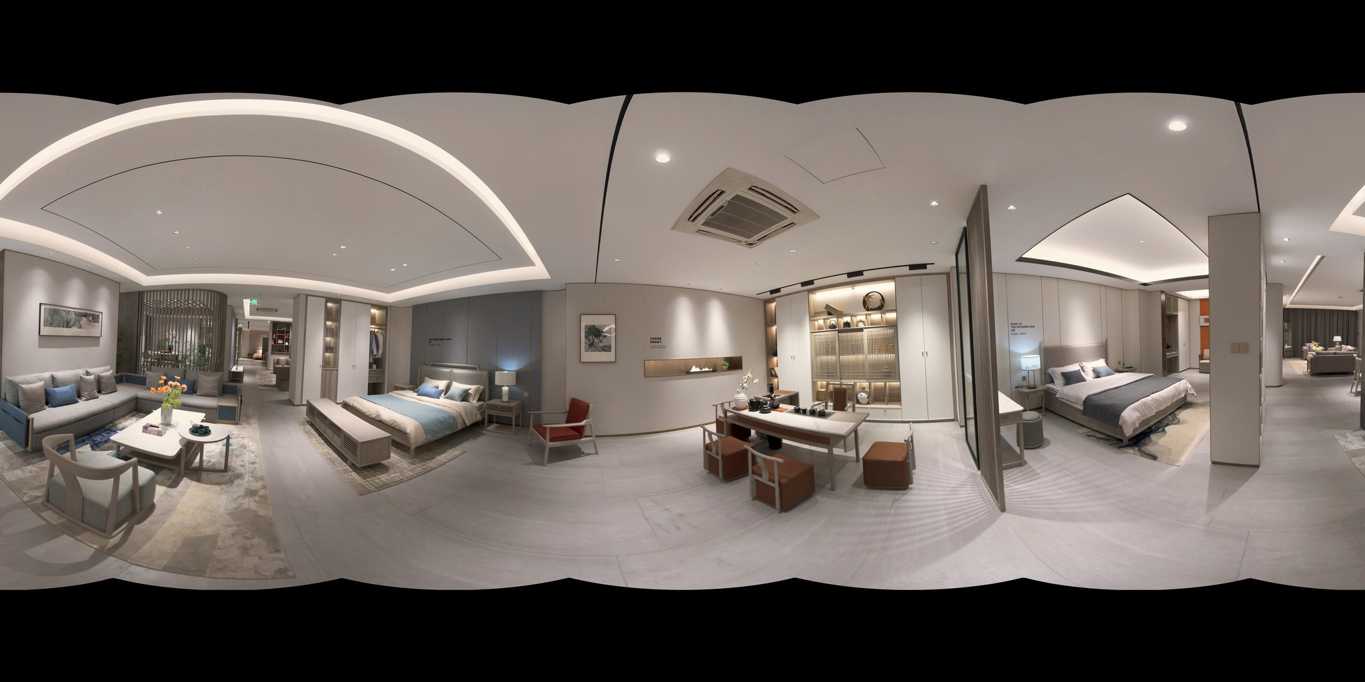}}
\end{subfigure}
\begin{subfigure}{0.8\columnwidth}
  \centering
  \scalebox{-1}[1]{\includegraphics[width=1\columnwidth, trim={0cm 0cm 0cm 0cm}, clip]{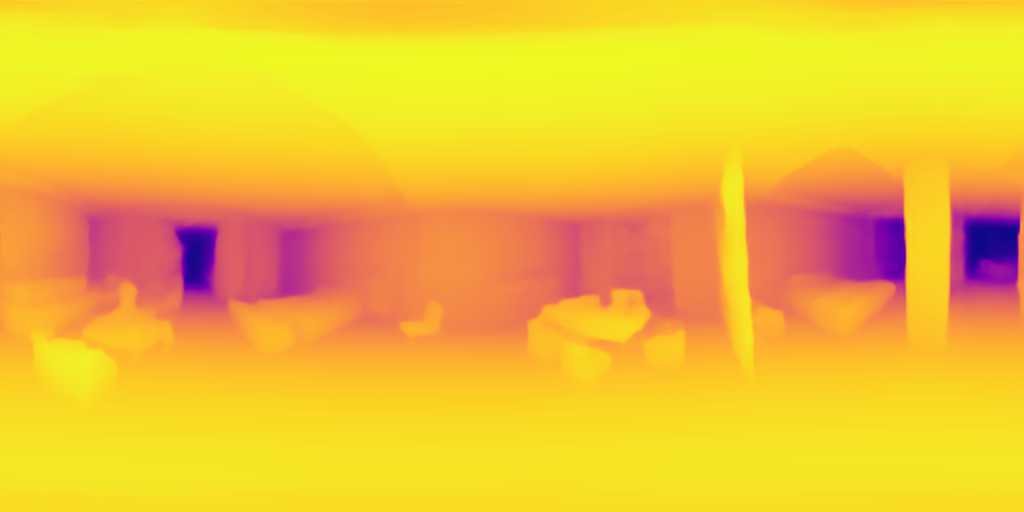}}
\end{subfigure}

\begin{subfigure}{0.9\columnwidth}
  \centering
  \includegraphics[width=1\columnwidth, trim={0cm 0cm 0cm 0cm}, clip]{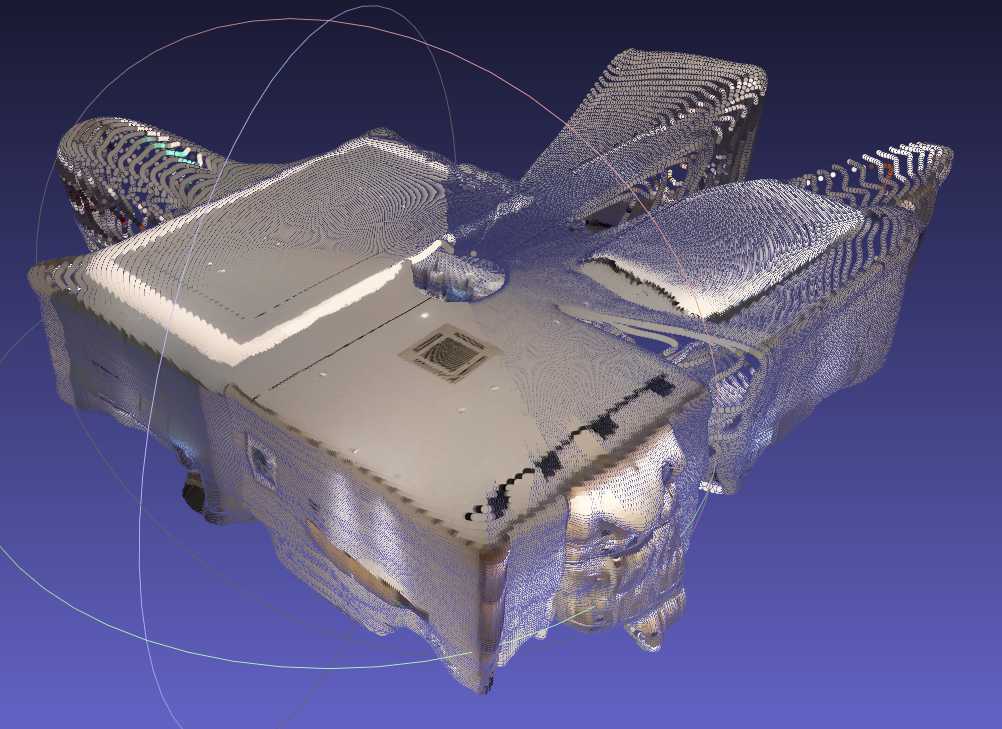}
\end{subfigure}
\begin{subfigure}{1.1\columnwidth}
  \centering
  \includegraphics[width=1\columnwidth, trim={0cm 0cm 0cm 0cm}, clip]{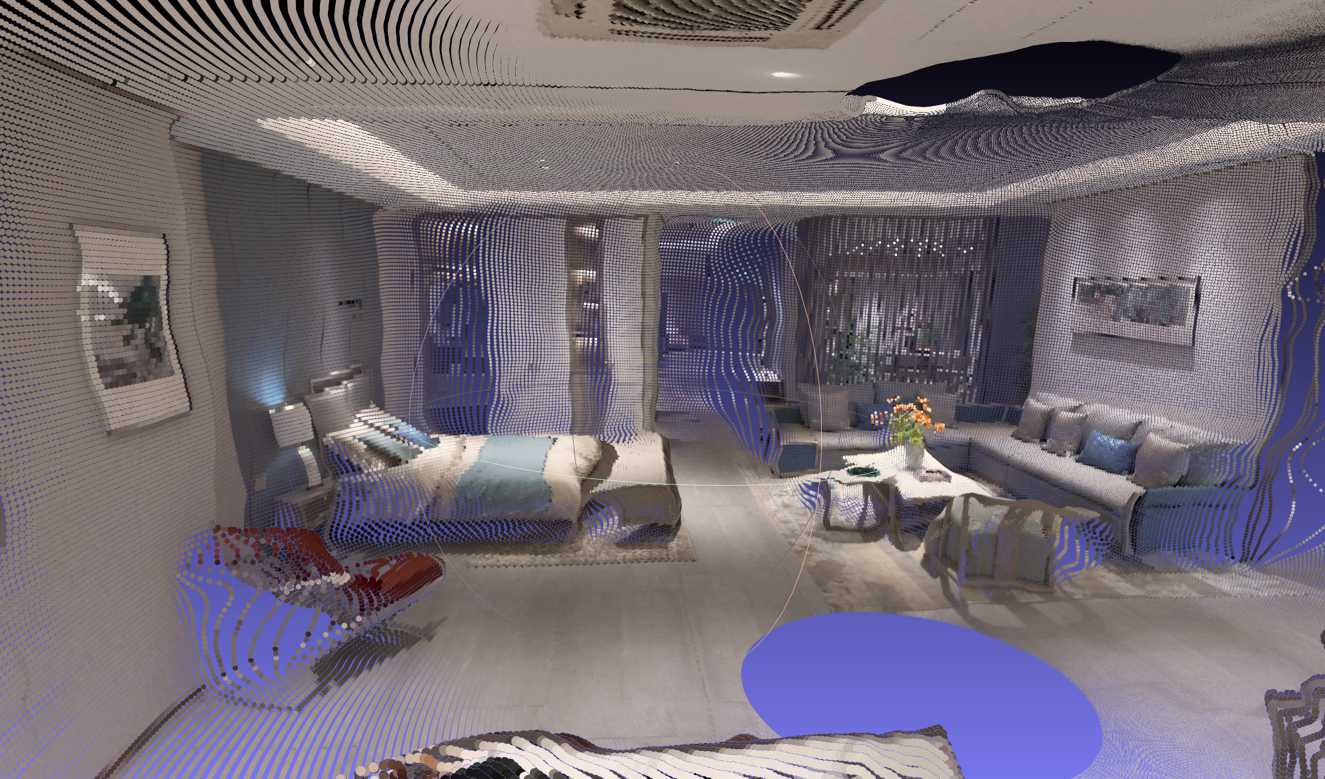}
\end{subfigure}

\begin{subfigure}{0.8\columnwidth}
  \centering
  \scalebox{-1}[1]{\includegraphics[width=1\columnwidth, trim={0cm 0cm 0cm 0cm}, clip]{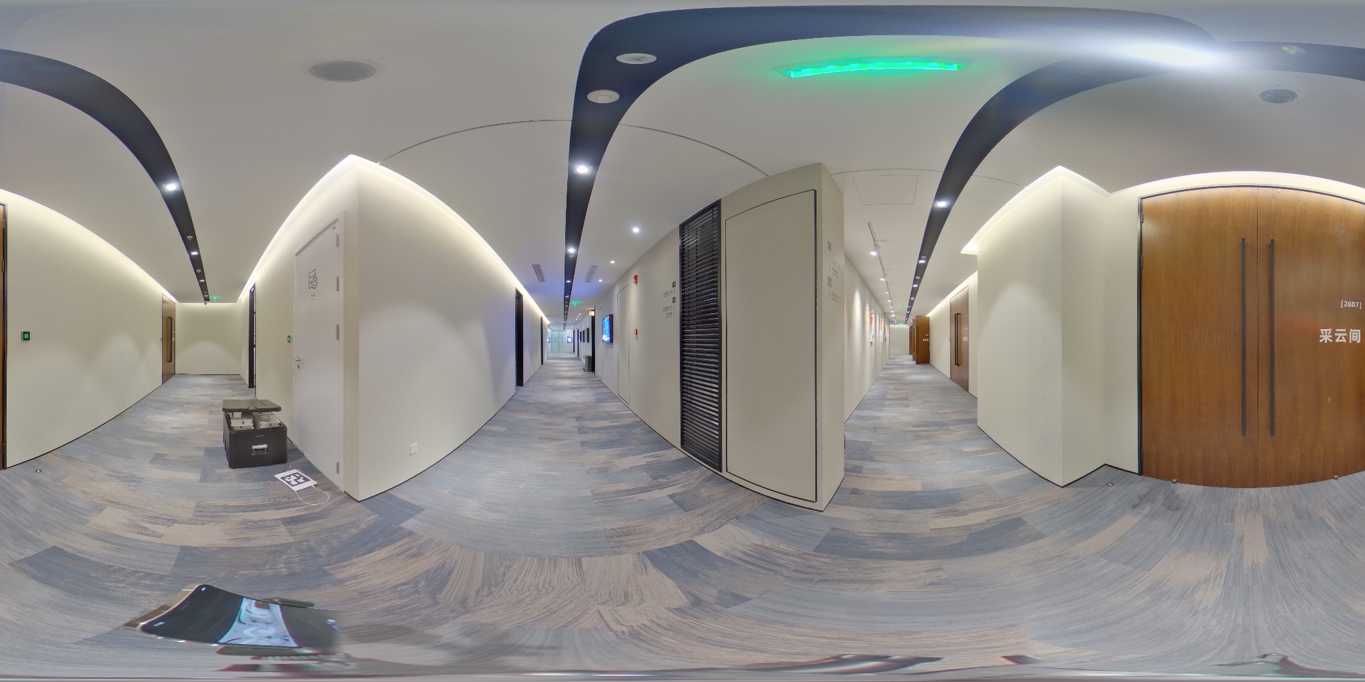}}
\end{subfigure}
\begin{subfigure}{0.8\columnwidth}
  \centering
  \scalebox{-1}[1]{\includegraphics[width=1\columnwidth, trim={0cm 0cm 0cm 0cm}, clip]{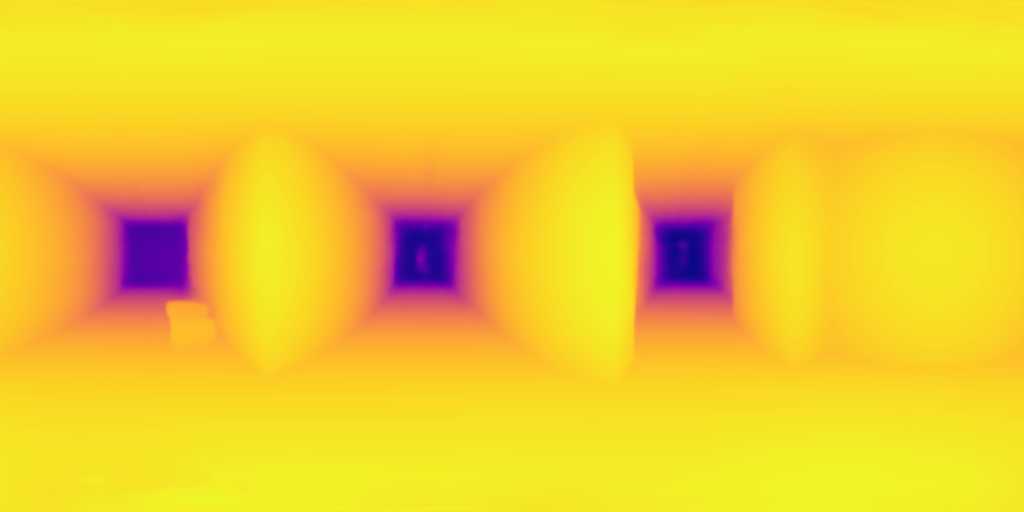}}
\end{subfigure}

\begin{subfigure}{0.85\columnwidth}
  \centering
  \includegraphics[width=1\columnwidth, trim={0cm 0cm 0cm 0cm}, clip]{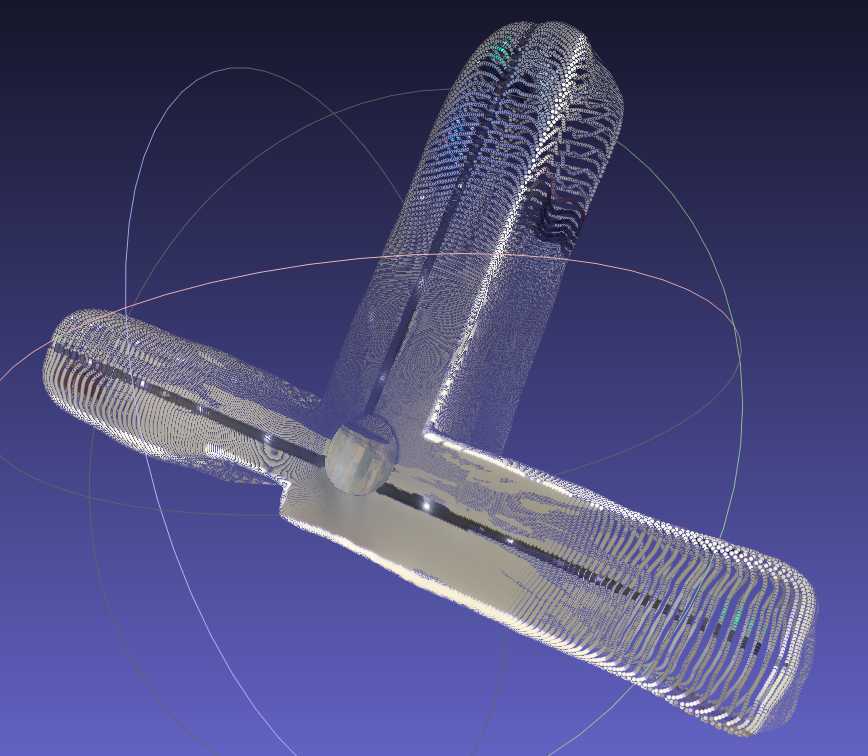}
\end{subfigure}
\begin{subfigure}{1.1\columnwidth}
  \centering
  \includegraphics[width=1\columnwidth, trim={0cm 0cm 0cm 0cm}, clip]{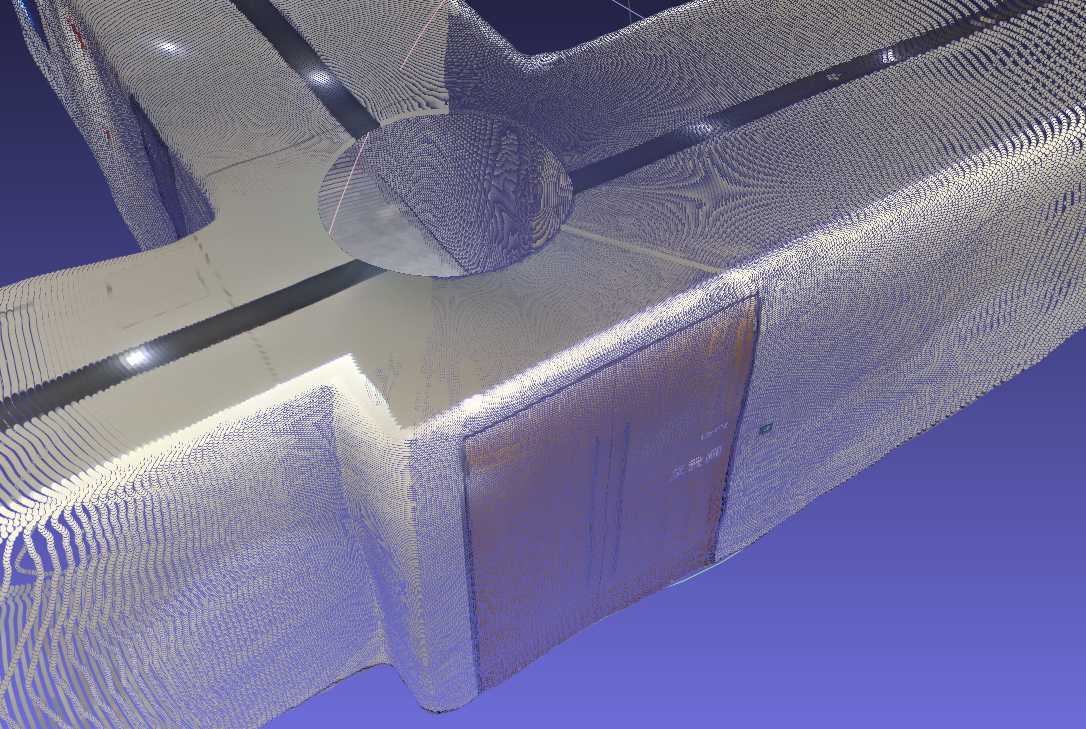}
\end{subfigure}

\caption{Point cloud visualization on unseen panorama images.}
\label{fig:res-panoramapcd}
\end{figure*}

Furthermore, to recover the complete scenarios, we collect some new indoor panorama images in the real world, and apply our model to these unseen images. The estimated depth maps and the projected 3D point clouds are displayed in Figure~\ref{fig:res-panoramapcd}, where the whole structures of the rooms are successfully reconstructed. The floors, ceilings, and the walls keep flat from the near to far. The straight lines are kept straight and the right angles are kept right. 
The decent performance on the unseen images shows the generalization ability of our model, which has the potential to be applied to real-world applications.

{\small
\bibliographystyle{ieee_fullname}
\bibliography{egbib}
}